\documentclass{article}

\PassOptionsToPackage{numbers,square,sort&compress}{natbib}
\usepackage[preprint]{neurips_2026}

\usepackage{xcolor}         
\definecolor{cvprblue}{rgb}{0.21,0.49,0.74}
\usepackage[pagebackref,breaklinks,hypertexnames,colorlinks,allcolors=cvprblue]{hyperref}

\usepackage{amsmath,amsfonts,bm}

\def\eqref#1{equation~\ref{#1}}

\def\1{\bm{1}}

\DeclareMathAlphabet{\mathsfit}{\encodingdefault}{\sfdefault}{m}{sl}
\SetMathAlphabet{\mathsfit}{bold}{\encodingdefault}{\sfdefault}{bx}{n}

\usepackage{fontawesome}
\usepackage[utf8]{inputenc} 
\usepackage[T1]{fontenc}    
\usepackage{booktabs}       
\usepackage{amsfonts}       
\usepackage{nicefrac}       
\usepackage{microtype}      
\usepackage{amsmath}
\usepackage{graphicx}
\usepackage{tabularx}
\usepackage{multirow}
\usepackage{subcaption}
\usepackage{tabu}
\usepackage{xcolor,colortbl}
\usepackage{xspace}
\usepackage{bm}
\usepackage{amssymb}
\usepackage{pifont}
\usepackage{makecell}
\usepackage{arydshln}
\usepackage{utfsym}
\usepackage{placeins}
\usepackage{cleveref}

\usepackage{wrapfig}
\usepackage{todonotes}
\setuptodonotes{inline}

\usepackage[accsupp]{axessibility}  

\usepackage{orcidlink}

\definecolor{tabfirst}{rgb}{1, 0.7, 0.7} 
\definecolor{tabsecond}{rgb}{1, 0.85, 0.7} 
\definecolor{tabthird}{rgb}{1, 1, 0.7} 

\newcommand{\benchmark}{SysCON3D\xspace}
\newcommand{\metr}{MEt3R\xspace}
\newcommand{\threesixty}{$360^\circ$}
\newcommand{\etal}{et~al.\xspace}

\input{app_toc}
\title{Can These Views Be One Scene?\\Evaluating Multiview 3D Consistency when\\3D Foundation Models Hallucinate}
\author{%
  Soumava Paul\thanks{Equal contribution} \quad Prakhar Kaushik\footnotemark[1] \quad Alan Yuille \\
  CCVL, Johns Hopkins University \\
  \texttt{\{spaul27, pkaushi1, ayuille1\}@jh.edu}\\ \\
  Project Page: \href{https://mvp18.github.io/3d-consistency-metrics}{mvp18.github.io/3d-consistency-metrics}
}

\begin{document}

\maketitle

\begin{figure}[h]
\vspace{-0.2cm}
  \centering
  \includegraphics[width=\columnwidth]{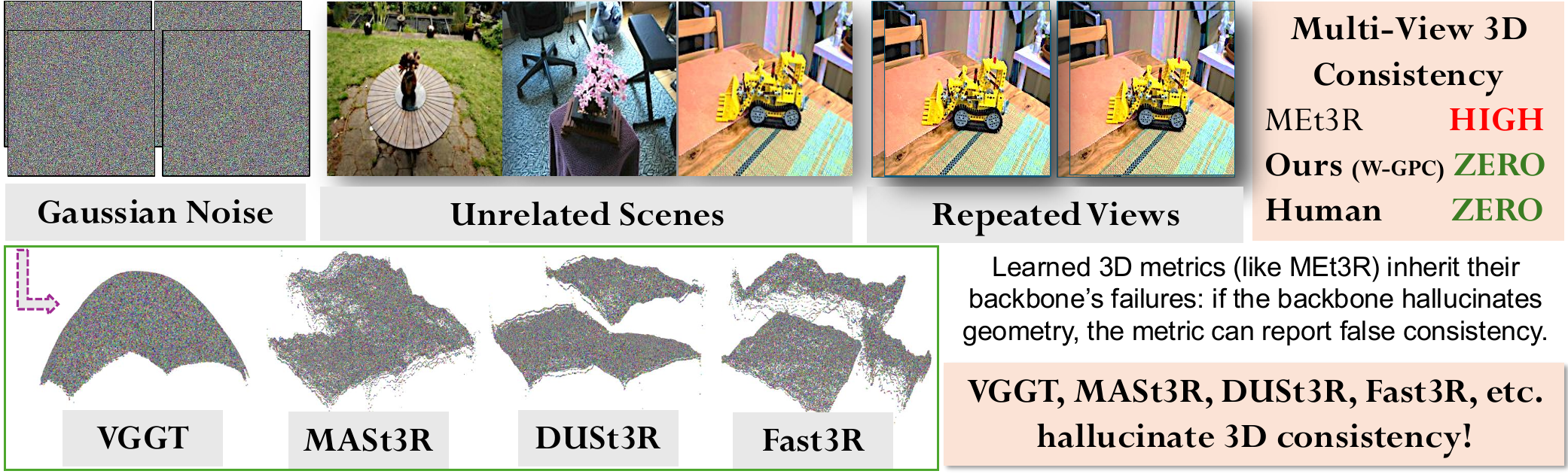}
\caption{\textbf{Can these views be one scene?} Unrelated scenes, repeated views, and Gaussian noise should be scored as 3D-inconsistent because they do not define a single static scene from multiple views. However, learned reconstruction backbones such as VGGT, MASt3R, DUSt3R, and Fast3R can still produce dense geometry on these inputs. Metrics built on these backbones, such as MEt3R, can therefore report spuriously high 3D consistency. We study
this failure mode with \benchmark, diagnose it with a backbone--residual--aggregation metric family, and introduce robust and more human-aligned COLMAP-based metrics.}
  \label{fig:teaser}
\end{figure}

\begin{abstract}
Multiview 3D evaluation assumes that the images being scored are observations of one static 3D scene. This assumption can fail in NVS and sparse-view reconstruction: inputs or generated outputs may contain artifacts, outlier frames, repeated views, or noise, yet still receive high 3D consistency scores. Existing reference-based metrics require ground truth, while ground-truth-free metrics such as MEt3R depend on learned reconstruction backbones whose failure modes are poorly characterized. We study this reliability problem by comparing neural reconstruction priors with classical geometric verification. We introduce \benchmark, a controlled robustness benchmark for multiview 3D consistency, and a parametric family that decomposes neural metrics into backbone, residual, and aggregation components. This family recovers MEt3R and yields variants up to $3\times$ more robust. Our analysis shows that VGGT, MASt3R, DUSt3R, and Fast3R can hallucinate dense geometry and cross-view support for unrelated scenes, repeated images, and random noise. We introduce COLMAP-based metrics that use matches, registration, dense support, and reconstruction failure as failure-aware consistency signals. On real NVS outputs and a structured human study, these metrics achieve up to $4\times$ higher correlation with human judgments than MEt3R.
\end{abstract}

\section{Introduction}

Most multiview 3D pipelines begin with an assumption that the input images are different observations of the same static 3D scene. In practice, this assumption is often violated. Sparse-view reconstruction may ingest internet or generated images containing artifacts, noise, repeated frames, or outlier views~\cite{han2025emergent}; Novel View Synthesis (NVS) methods may produce views that look plausible individually but are geometrically incoherent with each other. A reconstruction algorithm may still return geometry for such inputs. A 3D consistency metric has a stricter role: it must decide whether the images actually provide evidence for a coherent 3D explanation. Measuring this evidence without ground-truth geometry, reference renderings, or extensive human inspection is the problem we study.

Multiview 3D consistency exposes a useful contrast between two traditions in vision. Classical multiview geometry verifies reconstruction through explicit evidence: feature matches, epipolar constraints, inlier support, camera registration, dense stereo, and viewpoint coverage. Such pipelines can be slow and can fail under difficult imaging conditions, but their failures are visible. Learned 3D reconstruction models are fast, flexible, and often pose-free, making them attractive backbones for automatic evaluation. Their failure behavior is less transparent. When used as evaluators, they may return plausible geometry even when the input views provide no geometric support.

This question is increasingly practical. Recent NVS and sparse-view reconstruction methods operate on limited, noisy, or generated views, and their outputs may contain geometric drift, repeated content, inconsistent appearance, or out-of-scene artifacts~\cite{han2025emergent,xu2025depthsplat, yu2024viewcrafter, wu2025difix3d+}, occurring as a side-effect of the data collection process, model generalization failures, or due to adversarial or bad actors. Standard appearance metrics such as PSNR and SSIM require reference images and measure per-view fidelity. Others, such as the Chamfer distance, require ground-truth geometry. These metrics do not directly measure whether a generated view set is mutually consistent as observations of a single scene.

Recent ground-truth-free metrics address this gap by using learned 3D reconstruction models as evaluators. MEt3R~\cite{asim24met3r}, for example, reconstructs view pairs using DUSt3R~\cite{dust3r_cvpr24} or MASt3R~\cite{leroy2024grounding}, warps dense DINOv2/FeatUp features~\cite{oquab2023dinov2,fu2024featup}, and averages pairwise feature disagreement. While effective, this introduces two sources of fragility. First, pairwise mean aggregation collapses the residual distribution and can hide outlier views, cross-scene mixtures, or long-range drift. Second, the metric inherits the behavior of the reconstruction backbone. A reconstruction model is trained to produce geometry; a metric must decide whether the images support that geometry.

We study ground-truth-free 3D consistency evaluation as a robustness problem for automatic evaluators. \emph{We introduce \textbf{\benchmark}, a controlled benchmark that tests whether a metric rejects multiview inputs with known inconsistencies}: single outlier views, cross-scene mixtures, repeated images, patched noise, and full Gaussian noise. These corruptions are deliberately simple. They test a basic requirement: image sets with weaker or contradictory evidence for a shared 3D scene should receive worse consistency scores than clean multiview sets.

\emph{To diagnose learned metrics, we introduce a parametric family that decomposes each neural evaluator into a reconstruction backbone, a residual function, and an aggregation function}. This formulation recovers MEt3R as one point in the family and enables controlled ablations over where robustness is gained or lost. In particular, replacing mean aggregation with distributional aggregation yields stronger neural metrics; our best learned variant, \textbf{MASt3R-W-IMQ, is up to $3\times$ more robust than MEt3R} on \benchmark.

Our robustness analysis reveals a larger reliability failure in current learned 3D evaluators. \textbf{Modern reconstruction backbones, including VGGT~\cite{wang2025vggt}, DUSt3R~\cite{dust3r_cvpr24}, MASt3R~\cite{leroy2024grounding}, and Fast3R~\cite{Yang_2025_Fast3R}, can produce dense 3D structure and spurious cross-view support for inputs that do not correspond to any coherent 3D scene}, including unrelated scene mixtures, identical images, and random noise. The failure originates in the backbone: once unsupported geometry is produced, residual aggregation can only partially correct the score.

We therefore complement the neural metric family with COLMAP~\cite{schoenberger2016sfm}-based metrics that use classical geometric verification as a failure-aware signal. 
These metrics inherit the usual limitations of SfM/MVS, but their failure modes are measurable rather than hidden. In real NVS outputs from Mip-NeRF360 and DL3DV, and in a structured human study designed to isolate 3D consistency from visual realism and plausibility, \textbf{COLMAP-based metrics achieve the strongest alignment with human judgments, up to $4\times$ higher correlation than MEt3R.}

In summary, our main contributions are:
\begin{enumerate}
    \item \benchmark, a controlled robustness benchmark for ground-truth-free multiview 3D consistency metrics;
    \item a parametric neural metric family that decomposes learned evaluators into backbone, residual, and aggregation components, enabling systematic ablations and stronger learned metrics;
    \item \emph{a diagnostic study showing that state-of-the-art learned 3D reconstruction backbones hallucinate geometric support on inconsistent multiview inputs};
    \item COLMAP-based, failure-aware metrics that turn classical geometric verification into interpretable 3D consistency scores; and
    \item a structured human-alignment study for evaluating whether automatic metrics agree with human judgments of scene-level 3D consistency.
\end{enumerate}
\section{Related Work}
\label{sec:related}

\paragraph{Generative NVS and sparse-view reconstruction.}
Recent NVS and sparse-view reconstruction methods span diffusion-based generation and feed-forward 3D regression. Video diffusion methods such as Stable Virtual Camera~\cite{zhou2025stable}, ViewCrafter~\cite{yu2024viewcrafter}, and NVS-Solver~\cite{you2024nvs} adapt pretrained video generators with camera conditioning, while multiview diffusion methods such as MVGenMaster~\cite{cao2025mvgenmaster}, Difix3D+~\cite{wu2025difix3d+}, and GenFusion~\cite{wu2025genfusion} use image or video diffusion priors for generation and reconstruction refinement. Feed-forward methods such as DepthSplat~\cite{xu2025depthsplat}, Long-LRM~\cite{ziwen2025long}, and MVSplat360~\cite{chen2024mvsplat360} directly predict 3D Gaussian representations from sparse views. We evaluate these models at the scene level, where per-view fidelity can miss failures in cross-view geometry.

\paragraph{Learned multiview reconstruction backbones.}
DUSt3R~\cite{dust3r_cvpr24} formulates pose-free reconstruction as pairwise dense point-map regression. MASt3R~\cite{leroy2024grounding} augments this representation with dense local features for matching and registration across image pairs. Fast3R~\cite{Yang_2025_Fast3R} and VGGT~\cite{wang2025vggt} instead process all views jointly, amortizing multiview reconstruction into a single global forward pass. These models are fast and feedforward, making them attractive backbones for ground-truth-free evaluation. However, their learned priors make metric behavior difficult to attribute: a good 3d-consistency score can stem from image-supported geometry, prior-driven completion, or spurious correspondences. Our parametric family exposes this dependence by separating the reconstruction backbone from the residual and aggregation functions. Early evidence of our findings can be found in Doppelgangers++~\cite{xiangli2025doppelgangers}, which shows that MASt3R's point maps conflate visually similar but geometrically distinct surfaces, yielding incorrect 3D matches; our analysis finds this failure extends well beyond visual aliasing, to unrelated scenes, repeated views, and pure noise.

\paragraph{Evaluation metrics for multiview generation.}
When reference images or ground-truth geometry are available, NVS and reconstruction are commonly evaluated using PSNR, SSIM, LPIPS, Chamfer distance, or benchmark-specific geometric criteria, as in DTU~\cite{yao2018mvsnet} and Tanks \& Temples~\cite{knapitsch2017tanks}. These protocols are valuable for supervised benchmarks, but they do not directly measure whether a generated multiview set is mutually consistent without references. MEt3R~\cite{asim24met3r} addresses this setting with a pose-free pairwise consistency score based on learned reconstruction and warped DINOv2 features~\cite{oquab2023dinov2,fu2024featup}. SED/TSED~\cite{yu2023long} measures epipolar consistency from SIFT matches when camera poses are known, while PRISM~\cite{stern2025appreciate} compares distributions of object-centric source--target embeddings relative to a reference anchor distribution. Gen3DEval~\cite{maiti2025gen3deval} and Eval3D~\cite{cvpr2025eval3d} use foundation-model probes for object-centric evaluation across text-3D alignment, appearance, surface quality, and semantic coherence. In contrast, we study scene-level, ground-truth-free multiview consistency and explicitly test metric robustness under controlled outliers and corruptions.

\paragraph{Robustness, distributional testing, and geometric verification.}
Robustness benchmarks in vision have shown that standard evaluation can hide systematic failure under structured corruptions~\cite{hendrycks2019benchmarking}; robust statistics similarly studies estimator behavior under outliers and contamination~\cite{huber1964robust,hampel1974influence}. We bring this perspective to 3D consistency metrics by constructing \benchmark, where the corruption is multiview rather than per-image: foreign views, cross-scene mixtures, repeated inputs, and synthetic noise. For neural metrics, we use distributional aggregation to preserve information discarded by mean pairwise scoring, drawing on kernel two-sample distances such as MMD~\cite{gretton2012kernel}. For classical metrics, we build on the verification principle underlying RANSAC~\cite{fischler1981random} and structure-from-motion systems such as COLMAP~\cite{schoenberger2016sfm}: geometry is supported by inlier matches, registration, dense reconstruction, and viewpoint coverage, whose failure, can also be used as measurable signals of 3D consistency.
\section{Metric Design for Robust 3D Consistency}
\label{sec:metrics}

\paragraph{Problem setup.}
We consider a static scene observed from $K$ viewpoints,
$\mathcal{I}=\{I_k\}_{k=1}^K$, where
$I_k\in[-1,1]^{3\times H_k\times W_k}$ is an RGB image and
$\Omega_k\subset\mathbb{Z}^2$ is its pixel grid. Our goal is a
scene-level, ground-truth-free score for the 3D consistency of an image set.
For neural residual metrics, we use inconsistency scores: lower values indicate
more geometrically consistent views, while higher values indicate inconsistent
sets such as cross-scene mixtures or noise. The COLMAP metrics in
~\Cref{sec:colmap-metrics} are reported as consistency scores, where higher is
better. 

We study two metric families. The first uses learned reconstruction backbones
and is designed to make neural evaluators ablatable. The second uses classical
geometric verification, where registration, densification, and reconstruction
failure are observable signals.

\subsection{Neural metrics as backbone--residual--aggregation triplets}
\label{sec:neural-family}

All neural multiview consistency metrics we study, including
MEt3R~\cite{asim24met3r}, can be written as
\begin{equation}
  m(\mathcal{I}) \;=\; \mathcal{A}\!\Big(\;\rho\big(\,B(\mathcal{I}),\;\phi(\mathcal{I})\,\big)\;\Big).
  \label{eq:pipeline}
\end{equation}
Here $B$ is a reconstruction backbone, $\rho$ constructs a residual set from
the reconstruction and image features, and $\mathcal{A}$ aggregates residuals
into a scalar score. Each metric shown later in our robustness analysis \Cref{tab:metric_calibration_effect} corresponds to a triplet $(B,\rho,\mathcal{A})$.

\paragraph{Backbone.}
The backbone $B$ reconstructs a 3D point cloud
$\mathcal{X}=\{\mathbf{x}_n\}$ and cameras $\{\Pi_k\}_{k=1}^K$ from the input
views. A pairwise backbone processes one image pair at a time: MASt3R~\cite{leroy2024grounding}
predicts per-pixel 3D point maps and confidences, and fuses pairwise outputs
into a canonical frame after $\binom{K}{2}$ forward passes. Global backbones
such as VGGT~\cite{wang2025vggt} and Fast3R~\cite{Yang_2025_Fast3R} ingest all
$K$ views in a single forward pass, producing joint geometry and cameras with
amortized cost.

\paragraph{Residuals.}
The residual function $\rho$ produces non-negative cross-view feature
disagreements $\mathcal{E}=\{e_n\}_{n=1}^N$. A frozen DINOv2~\cite{oquab2023dinov2}
feature extractor, upsampled with FeatUp~\cite{fu2024featup}, gives dense
features $F_k=\phi(I_k)\in\mathbb{R}^{768\times H_k\times W_k}$. For a pixel
$u$ in view $i$ and a corresponding pixel $v$ in view $j$, we use cosine
dissimilarity:  
\[
d_{ij}(u,v)
=
1-\frac{\langle F_i(u),F_j(v)\rangle}
{\|F_i(u)\|\,\|F_j(v)\|+\varepsilon}.
\]
We use two residual constructions, as follows:

\paragraph{Warp-based residuals.}
For each pair $(i,j)$, a pixel $u$ in view $i$ is lifted to its reconstructed
3D point $\mathbf{x}_u\in\mathcal{X}$ and projected into view $j$ as
$\pi_j(\mathbf{x}_u)$. Residuals are collected in both warp directions:
\begin{equation}
  \mathcal{E}_{\text{warp}} = \bigcup_{\substack{i,j=1 \\ i < j}}^{K} \Big\{\, d_{ij}\!\big(u,\,\pi_j(\mathbf{x}_u)\big) : u \in \mathcal{V}_{i\to j}\,\Big\} \;\cup\; \Big\{\, d_{ji}\!\big(v,\,\pi_i(\mathbf{x}_v)\big) : v \in \mathcal{V}_{j\to i}\,\Big\},
  \label{eq:warp-residuals}
\end{equation}
where $\mathcal{V}_{i\to j}$ denotes depth-consistent pixels when warping from
view $i$ to view $j$.

\paragraph{Point-consistency residuals.}
Point-consistency (PC) residuals start from reconstructed 3D points rather than
view pairs. For a point $\mathbf{x}_n\in\mathcal{X}$ visible in
$\mathcal{V}_n=\mathrm{vis}(\mathbf{x}_n)$ with $|\mathcal{V}_n|\ge2$, we
project the point into each visible view and compare the corresponding features:
\begin{equation}
  \delta_n = \frac{1}{|\mathcal{V}_n|(|\mathcal{V}_n|{-}1)} \!\sum_{\substack{k,\ell \in \mathcal{V}_n \\ k \neq \ell}} d_{k\ell}\!\big(\pi_k(\mathbf{x}_n),\, \pi_\ell(\mathbf{x}_n)\big),
  \qquad
  \mathcal{E}_{\text{PC}} = \big\{\delta_n : |\mathcal{V}_n| \ge 2 \big\}.
  \label{eq:point-dispersion}
\end{equation}
Thus each element of $\mathcal{E}_{\text{PC}}$ summarizes how consistently one
3D point is described across its observing views. In this work, we evaluate PC
residuals with global backbones, VGGT, and Fast3R, while MEt3R-style evaluation
uses warp residuals. The PC residual set is computed in linear time per point
via a feature-sum identity (app.~\Cref{app:pc-feature-sum}).

\paragraph{Aggregation.}
Given residuals $\mathcal{E}=\{e_n\}_{n=1}^N$ with empirical distribution $P$,
the aggregation function compares $P$ to the ideal residual distribution
$\delta_0$, a Dirac mass at zero. The MEt3R-style baseline is the mean
$
m_{\text{Base}}(\mathcal{I})=\frac{1}{N}\sum_{n=1}^N e_n,
$
equivalently $W_1(P,\delta_0)$ for non-negative scalar residuals. Mean
aggregation collapses the residual distribution to one moment, which can dilute
outlier views, heavy tails, and multimodal residual patterns.

We therefore also evaluate distributional aggregations. For a kernel $k$,
\begin{equation}
\widehat{\mathrm{MMD}}^2_k(P,\delta_0)
=
\frac{1}{N(N{-}1)}\sum_{a\neq b}k(e_a,e_b)
-
\frac{2}{N}\sum_a k(e_a,0)
+
k(0,0).
\label{eq:mmd-dirac}
\end{equation}
We use either an RBF kernel
$k(e,e')=\exp\!\left(-(e-e')^2/2\sigma^2\right)$, with $\sigma$ set by the
median heuristic or fixed to $\sigma_{\text{ref}}=0.15$, or an inverse
multiquadric kernel
$k_{\text{IMQ}}(e,e')=(c^2+(e-e')^2)^{-1/2}$ with $c=1$. For the RBF kernel,
$k(0,0)=1$; for IMQ, $k(0,0)=1/c$, matching the constants used in the
corresponding estimators. We also use the energy distance
\[
\mathcal{D}_{\text{E}}(P,\delta_0)
=
2\,\mathbb{E}_{e\sim P}[|e|]
-
\mathbb{E}_{e,e'\sim P}[|e-e'|].
\]
Each neural metric is specified by a backbone, residual mode, and aggregation:
\[
B\in\{\text{MASt3R},\text{Fast3R},\text{VGGT}\},\quad
\rho\in\{\text{W},\text{PC}\},\quad
\mathcal{A}\in\{\text{Base},\text{MMD},\text{IMQ},\text{Energy}\}.
\]
For example, Fast3R-PC-Energy denotes point-consistency residuals with energy
aggregation, and MASt3R-W-IMQ denotes warp residuals with IMQ-MMD aggregation.
MEt3R~\cite{asim24met3r} corresponds to the MASt3R/W-Base point in this family. When ground-truth images are available, all three distributional aggregations
admit two-sample generalizations (app.~\Cref{app:mmd-two-sample}).

\paragraph{External baselines.}
We also report metrics outside this family. PRISM-MMD~\cite{stern2025appreciate}
extracts source--target embeddings from a Zero123-based diffusion backbone~\cite{liu2023zero}
conditioned on relative pose, then computes MMD between generated and anchor
embedding sets. SED~\cite{yu2023long} reports median symmetric epipolar
distance over SIFT matches given known fundamental matrices, and TSED thresholds
SED to count consistent pairs.

\subsection{Classical verification metrics}
\label{sec:colmap-metrics}

Learned reconstruction backbones return geometry for every input, including
inputs with weak or contradictory multiview evidence. Classical SfM/MVS exposes
a different signal: feature matches can be rejected, cameras can fail to
register, and dense reconstruction can fail to produce valid support. We convert
these observable outcomes into COLMAP-based consistency metrics.

Given a view set, we run COLMAP~\cite{schoenberger2016sfm} SfM and MVS. SfM
accepts feature matches that survive RANSAC under epipolar constraints, yielding
registered cameras and sparse 3D points. MVS then produces, for each densified
view, a photometric depth hypothesis and a geometric-consistency depth map. The
following COLMAP metrics are consistency scores, so higher is better.

\paragraph{Per-pixel agreement.}
For each view $v$ with usable dense outputs, let
$\mathcal{D}^{(v)}_g$ be the geometric-consistency depth map and
$\mathcal{D}^{(v)}_p$ the photometric depth hypothesis. Let $P_v$ be the pixel
set and define valid pixels
\[
\Omega_v=\left\{\mathbf{u}\in P_v:\ \mathcal{D}^{(v)}_{g}(\mathbf{u})>10^{-5},\
  \mathcal{D}^{(v)}_{g}(\mathbf{u}),\mathcal{D}^{(v)}_{p}(\mathbf{u})\ \text{finite}\right\}.
\]
We convert depth agreement into a bounded quality score:
\begin{equation}
  q_v(\mathbf{u})=
  \begin{cases}
    1-\mathrm{clip}\!\left(
      \frac{\left|\mathcal{D}^{(v)}_{p}(\mathbf{u})-\mathcal{D}^{(v)}_{g}(\mathbf{u})\right|}
      {\tau\,\max(\mathcal{D}^{(v)}_{g}(\mathbf{u}),10^{-6})},
    \,0,\,1\right), & \mathbf{u}\in\Omega_v,\\
    0, & \mathbf{u}\notin\Omega_v,
  \end{cases}
  \qquad \tau=0.2.
\end{equation}
The score is high when the two COLMAP depth estimates agree within tolerance
and zero for invalid pixels or large disagreements.

\paragraph{Geometric--Photometric Consistency.}
For each densified view, we separate dense support from agreement quality:
\[
\mathrm{Density}_v=\frac{|\Omega_v|}{|P_v|},
\qquad
\mathrm{Consistency}_v=\frac{1}{|\Omega_v|}\sum_{\mathbf{u}\in\Omega_v}q_v(\mathbf{u}),
\qquad
\mathrm{GPC}_v=\mathrm{Density}_v\cdot\mathrm{Consistency}_v.
\]
Let $D$ denote views with usable dense maps. The scene-level score is
$
\mathrm{GPC}=\frac{1}{|D|}\sum_{v\in D}\mathrm{GPC}_v.
$
GPC is high only when reconstruction is both dense and depth-consistent.

\paragraph{Integrated Consistency Mass.}
We also report a pixel-weighted aggregate measuring total reliable dense support:
\[
\mathrm{ICM}
=
\frac{\sum_{v\in D}\sum_{\mathbf{u}\in\Omega_v}q_v(\mathbf{u})}
{\sum_{v\in D}|P_v|}.
\]
ICM is conditional on successful densification. To penalize views that fail
registration or densification, we define the attempted-set variant,
\[
\mathrm{ICM}_{\text{all}}
=
\frac{\sum_{v\in D}\sum_{\mathbf{u}\in\Omega_v}q_v(\mathbf{u})}
{\sum_{v\in A}|P_v|},
\]
where $A$ is the set of all attempted images. Images outside $D$ contribute
zero mass but remain in the denominator.

\paragraph{Coverage-weighted GPC.}
To penalize reconstructions that explain only a narrow range of viewpoints, we
compute angular coverage $\omega\in[0^\circ,360^\circ]$ from registered camera
centers as $360^\circ$ minus the maximum azimuthal gap. We report the
coverage-weighted score (W-GPC)
\[
\mathrm{GPC}_\omega
=
\mathrm{GPC}\cdot\frac{\omega}{360^\circ}.
\]
Together, GPC, ICM, $\mathrm{ICM}_{\text{all}}$, and $\mathrm{GPC}_\omega$
measure dense agreement, reliable 3D mass, failed support, and viewpoint
coverage using classical geometric verification.
\section{\benchmark: Controlled Corruptions for 3D Consistency Metrics}
\label{sec:benchmark}

Classical SfM deliberately relies on sparse, repeatable evidence rather than explaining every pixel; unsupported images fail to match, fail epipolar verification, or fail registration.

\paragraph{Benchmark Construction}
\label{sec:benchmark-construction}

We construct three groups of multi-view image sets from Mip-NeRF360~\cite{barron2022mip} scenes at each viewpoint count $K \in \{3, 6, 9, 12, 15, 18, 21\}$. The groups comprise real 3D-consistent image sets and controlled inconsistencies of increasing severity, and each level isolates a distinct failure mode that an evaluator should detect.

\noindent\textit{(i) Consistent ($L_0$).} $K$ training views from a single scene (9 samples per $K$, one per scene). Any reliable metric should score such scenes as most 3D-consistent.

\noindent\textit{(ii) \emph{\benchmark-M}: Cross-scene mixtures.} Foreign views are introduced at increasing density (8 samples per $K$ each):

\noindent $L_1$ -- \textbf{Single-outlier}: $K{-}1$ views from one scene plus 1 foreign view (${\sim}2/K$ cross-scene pairs).

\noindent $L_2$ -- \textbf{Controlled mixture}: ${\sim}30\%$ foreign views, ${\sim}50{-}60\%$ cross-scene pairs at $K \ge 6$.

\noindent $L_3$ -- \textbf{Random mixture}: each view sampled i.i.d.\ from the 9 scenes (${\sim}89\%$ cross-scene pairs).

\noindent\textit{(iii) \emph{\benchmark-N}: Noise corruptions.}
\begin{itemize}
  \item \textbf{Patched Gaussian}: consistent views with 4 $\mathcal{N}(0.5, 0.2^2)$ noise patches (4 samples$/K$).
  \item \textbf{Gaussian}: every pixel sampled i.i.d.\ from $\mathcal{N}(0.5, 0.2^2)$, clipped to $[0,1]$ (4 samples per $K$).
\end{itemize}

In total, \benchmark contains 9 consistent and 32 inconsistent samples per $K$ across 7 view counts, all with fixed random seeds.

\paragraph{Robustness Criteria}
\label{sec:robustness-criteria}

We evaluate each metric along two complementary criteria.

\paragraph{Per-group separation (Cohen's $d$, win rate).}
For each inconsistent group $g \in \{L_1, L_2, L_3, \text{patched}, \text{Gaussian}\}$ and each view count $K$, we compute Cohen's $d = (\bar{x}_g - \bar{x}_{L_0}) / s_{\text{pooled}}$, the standardized score gap between $g$ and the consistent reference $L_0$. The \emph{win rate} is the fraction of $K$ values for which $d > 0$, i.e., the metric correctly assigns a worse score to the inconsistent group. We report per-group win rate and an overall win rate that pools across all inconsistency groups and $K$ values.

\paragraph{Severity ordering (rank agreement).}
Per-group separation rewards a metric for ranking any inconsistent input below $L_0$, but does not test whether the metric orders the inconsistency groups themselves by severity. We therefore additionally report rank-agreement statistics between each metric's induced ordering of \benchmark scene types and the expected severity ordering $L_0 < L_1 < L_2 < L_3 = \text{noise}$; full numbers are in app.~\Cref{app:rank-concordance}. This criterion separates metrics that pass per-group separation but rank scene types otherwise arbitrarily.

\subsection{Robustness Results on \benchmark}
\label{sec:syscon3d-results}

\begin{table*}[!htbp]
\centering
\caption{\textbf{Robustness analysis of 3D consistency metrics on the \benchmark Benchmark.} Cohen's $d$ (averaged over $K \in \{3,\ldots,21\}$) and \emph{win rate} (fraction of $K$ values with $d > 0$) comparing consistent ($L_0$) sets against five 3D inconsistencies. Positive $d$ indicates the metric correctly assigns higher (worse) scores to inconsistent inputs. $L_1$--$L_3$ denote increasing levels of cross-scene corruption. The repeated-views (Identical Images) degenerate diagnostic is reported separately in app.~\cref{tab:metric_calibration_effect_identical_vs_distinct}. Top-3 values per column are shaded.}
\label{tab:metric_calibration_effect}
\setlength{\tabcolsep}{3pt}
\renewcommand{\arraystretch}{1.3}
\resizebox{\columnwidth}{!}{%
\begin{tabular}{llccccccccccc}
\toprule
\multirow{3}{*}{Backbone} & \multirow{3}{*}{Variant} & \multicolumn{6}{c}{$\stackrel{\mbox{increasing cross-scene corruption}}{\longrightarrow}$} & \multicolumn{5}{c}{} \\
 & & \multicolumn{2}{c}{Mixed-1-Outlier [L$_1$]} & \multicolumn{2}{c}{Mixed-Controlled [L$_2$]} & \multicolumn{2}{c}{Mixed-Random [L$_3$]} & \multicolumn{2}{c}{Patched Gaussian} & \multicolumn{2}{c}{Gaussian Noise} & \multicolumn{1}{c}{Overall} \\
\cmidrule(lr){3-4} \cmidrule(lr){5-6} \cmidrule(lr){7-8} \cmidrule(lr){9-10} \cmidrule(lr){11-12} \cmidrule(lr){13-13}
 & & Avg $d$ & Win\% ($d$) & Avg $d$ & Win\% ($d$) & Avg $d$ & Win\% ($d$) & Avg $d$ & Win\% ($d$) & Avg $d$ & Win\% ($d$) & Win\% ($d$) \\
\midrule
MASt3R & W-Energy & +0.03 & \cellcolor{tabsecond}57\% & +1.01 & 86\% & +3.15 & \cellcolor{tabfirst}100\% & +0.94 & \cellcolor{tabsecond}100\% & -3.36 & 0\% & \cellcolor{tabsecond}69\% \\
MASt3R & W-MMD & -0.78 & 14\% & -1.10 & 14\% & +0.53 & 71\% & -0.61 & 14\% & -0.93 & 0\% & 23\% \\
MASt3R & W-IMQ & +0.12 & \cellcolor{tabthird}57\% & +1.25 & \cellcolor{tabfirst}100\% & \cellcolor{tabthird}+3.56 & \cellcolor{tabsecond}100\% & \cellcolor{tabsecond}+1.16 & \cellcolor{tabthird}100\% & -2.64 & 0\% & \cellcolor{tabfirst}71\% \\
Fast3R & W-Base & -0.04 & 29\% & +1.04 & 86\% & +2.68 & \cellcolor{tabthird}100\% & -0.66 & 14\% & -9.52 & 0\% & 46\% \\
Fast3R & W-Energy & -0.07 & 43\% & +0.66 & 86\% & +2.53 & 100\% & -0.62 & 0\% & -6.09 & 0\% & 46\% \\
Fast3R & W-MMD & -0.19 & 14\% & +0.17 & 71\% & +0.97 & 100\% & -0.90 & 0\% & -3.75 & 0\% & 37\% \\
Fast3R & W-IMQ & -0.09 & 43\% & +0.86 & 86\% & +2.30 & 100\% & -0.48 & 0\% & -5.48 & 0\% & 46\% \\
Fast3R & PC-Base & \cellcolor{tabsecond}+0.16 & 57\% & \cellcolor{tabthird}+1.41 & \cellcolor{tabsecond}100\% & +3.42 & 100\% & +0.33 & 71\% & -3.24 & 0\% & 66\% \\
Fast3R & PC-Energy & +0.08 & 43\% & \cellcolor{tabsecond}+1.52 & \cellcolor{tabthird}100\% & \cellcolor{tabsecond}+3.87 & 100\% & +0.62 & 100\% & -2.75 & 0\% & \cellcolor{tabthird}69\% \\
Fast3R & PC-MMD & -0.24 & 29\% & +0.38 & 71\% & +1.01 & 86\% & -0.63 & 57\% & -0.25 & 43\% & 57\% \\
Fast3R & PC-IMQ & \cellcolor{tabfirst}+0.18 & 43\% & \cellcolor{tabfirst}+1.67 & 100\% & \cellcolor{tabfirst}+4.46 & 100\% & +0.42 & 71\% & -2.44 & 0\% & 63\% \\
VGGT & W-Base & +0.14 & 57\% & +0.81 & 86\% & +2.36 & 100\% & -0.65 & 0\% & -9.18 & 0\% & 49\% \\
VGGT & W-Robust & +0.22 & 57\% & +1.23 & 100\% & +2.63 & 100\% & -0.67 & 0\% & -9.12 & 0\% & 51\% \\
VGGT & W-Energy & +0.12 & \cellcolor{tabfirst}71\% & +0.34 & 86\% & +1.78 & 86\% & -0.49 & 0\% & -6.43 & 0\% & 49\% \\
VGGT & W-MMD & -0.07 & 43\% & +0.08 & 43\% & +0.89 & 100\% & -0.26 & 0\% & -4.60 & 0\% & 37\% \\
VGGT & W-IMQ & +0.11 & 57\% & +0.53 & 86\% & +1.94 & 100\% & -0.60 & 0\% & -5.76 & 0\% & 49\% \\
VGGT & PC-Base & -0.17 & 43\% & +0.37 & 57\% & +1.58 & 100\% & +0.00 & 71\% & -6.89 & 0\% & 54\% \\
VGGT & PC-Energy & -0.30 & 29\% & +0.33 & 57\% & +1.41 & 100\% & +0.07 & 57\% & -5.97 & 0\% & 49\% \\
VGGT & PC-MMD & -0.46 & 14\% & -0.47 & 14\% & +0.08 & 43\% & +0.20 & 43\% & +1.31 & \cellcolor{tabfirst}100\% & 43\% \\
VGGT & PC-IMQ & -0.25 & 29\% & +0.39 & 71\% & +1.53 & 100\% & +0.02 & 57\% & -4.37 & 0\% & 51\% \\
\hdashline
\multicolumn{2}{c}{MEt3R} & -0.48 & 14\% & -1.03 & 0\% & -1.57 & 0\% & \cellcolor{tabthird}+1.13 & \cellcolor{tabfirst}100\% & -3.96 & 0\% & 23\% \\
\multicolumn{2}{c}{PRISM-MMD} & -0.08 & 43\% & -0.16 & 43\% & -0.09 & 29\% & \cellcolor{tabfirst}+1.61 & 100\% & \cellcolor{tabsecond}+4.71 & \cellcolor{tabsecond}100\% & 63\% \\
\multicolumn{2}{c}{SED} & \cellcolor{tabthird}+0.15 & 57\% & -0.16 & 43\% & -0.62 & 0\% & +0.09 & 57\% & \cellcolor{tabfirst}+161.74 & 86\% & 49\% \\
\multicolumn{2}{c}{TSED} & -0.25 & 14\% & +0.43 & 86\% & +0.75 & 71\% & -0.35 & 0\% & \cellcolor{tabthird}+2.39 & \cellcolor{tabthird}100\% & 54\% \\
\bottomrule
\end{tabular}
}
\end{table*}

\begin{figure}[!htbp]
  \centering
  \includegraphics[width=\textwidth]{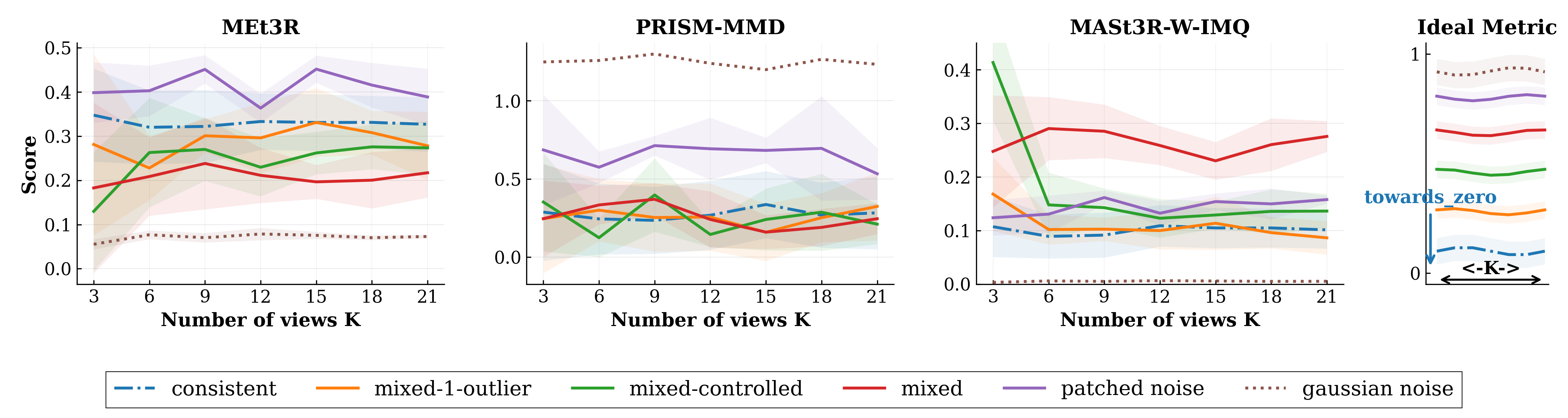}
  \caption{
    \textbf{Metric trends across SysCON3D scene types.}
    Each panel plots metric score (lower=better) vs.\ view count $K$ across SysCON3D scene types; a robust metric should separate these bands in order of increasing 3D inconsistency.
    \textbf{Panel~1:} \metr conflates $L_0$ with cross-scene mixtures and scores Gaussian noise as more consistent than $L_0$.
    \textbf{Panel~2:} PRISM-MMD -- recovers the expected SysCON3D ordering best (app.~\Cref{tab:rank_concordance}) but fails on cross-scene mixtures ($L_1$--$L_3$), correctly separating only noise.
    \textbf{Panel~3:} MASt3R W-IMQ -- distributional aggregation cleanly separates $L_0$ from cross-scene mixtures, but still scores noise as more consistent due to degenerate reconstructions (app.~\Cref{sec:recon-behavior}).
    \textbf{Panel~4:} Ideal: $L_0$ lowest, with progressively higher non-overlapping bands for $L_1$--$L_3$ and Gaussian noise.
  }
  \label{fig:calibration-bars}
\end{figure}

\paragraph{Results.}
\Cref{tab:metric_calibration_effect} reports Cohen's $d$ and win rate for each metric variant across \benchmark inconsistency groups; the repeated-views (Identical Images) degenerate diagnostic is reported separately in app.~\Cref{tab:metric_calibration_effect_identical_vs_distinct}. \Cref{fig:calibration-bars} visualizes relative score trends across $K$ for MASt3R W-Base (MEt3R) and MASt3R W-IMQ, chosen because it achieves the highest overall win rate. We highlight key patterns grouped by aggregation, residual, and backbone below.

\paragraph{Aggregation.}
Distributional aggregation yields the largest robustness improvements. For MASt3R, switching from W-Base (\metr) to W-IMQ raises the overall win rate from 23\% to 71\% with 100\% win on both $L_2$ and $L_3$; W-Energy shows similar gains (69\% overall). With Fast3R PC residuals, PC-IMQ achieves the largest effect size of any variant on $L_3$. MMD is consistently the weakest distributional aggregation (23--37\% overall), likely because its fixed RBF bandwidth loses sensitivity as residual distributions shift.

\paragraph{Residual.}
PC residuals improve robustness over warp-based residuals when paired with global backbones. On Fast3R, PC-Energy reaches 69\% overall versus 46\% for W-Energy. On VGGT, the gap is smaller (PC-Base: 54\% vs.\ W-Base: 49\%), suggesting VGGT's globally coherent geometry already provides adequate correspondences for warp-based evaluation.

\paragraph{Backbone.}
Global backbones outperform pairwise MASt3R at baseline aggregation: Fast3R and VGGT W-Base achieve 46\% and 49\% versus 23\% for MASt3R W-Base. However, the gap narrows under distributional aggregation -- MASt3R W-IMQ (71\%) surpasses both Fast3R W-IMQ (46\%) and VGGT W-IMQ (49\%), indicating that the choice of aggregation can compensate for weaker pairwise geometry. VGGT/W-Robust, our adaptation of RobustVGGT~\cite{han2025emergent} that scores each view by its late-layer attention and feature agreement with an anchor view and reruns reconstruction on the retained subset (app.~\Cref{app:vggt-robust}), helps cross-scene mixtures (100\% win on $L_2$ and $L_3$) but does not fix noise (0\% on patched/Gaussian), giving 51\% overall.

\begin{figure}[!h]
  \centering
  \includegraphics[width=\columnwidth]{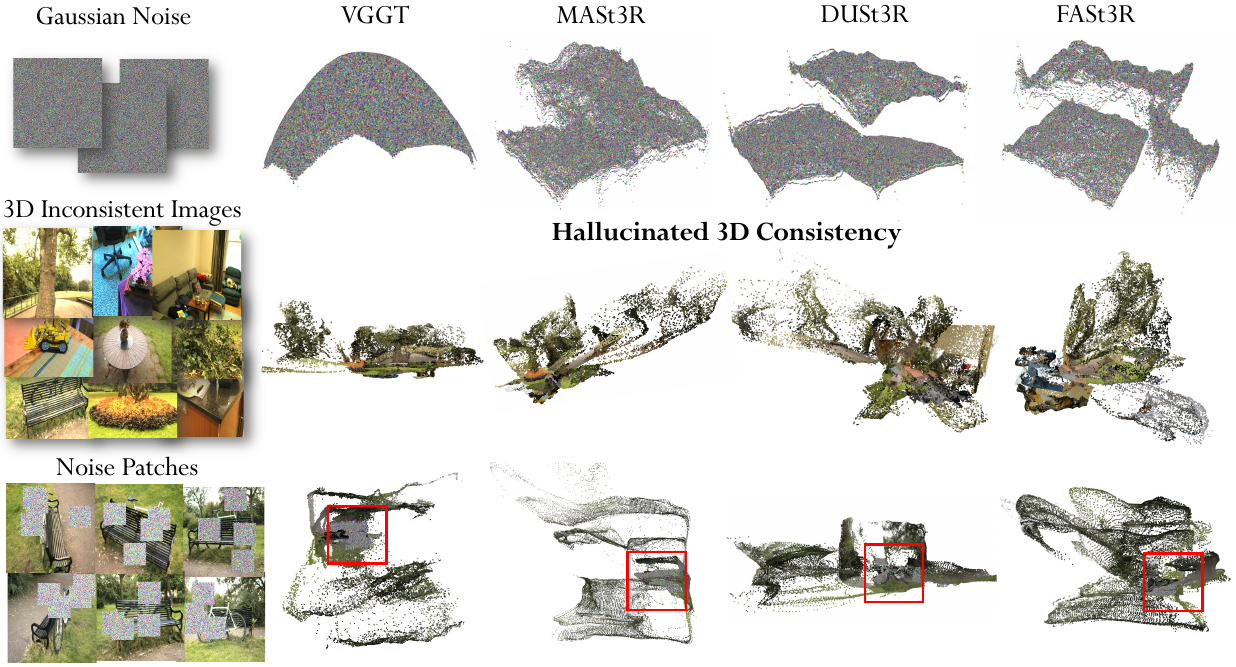}
  \caption{\textbf{Learned 3D backbones hallucinate geometric support.}
  VGGT, MASt3R, and DUSt3R produce dense point clouds on pure noise and
  cross-scene mixtures, although the inputs do not admit a coherent 3D scene. Interestingly, we notice that these models seem to produce distinct, hallucinatory patterns (e.g. a umbrella for VGGT). More in ~\Cref{app:hallu_pattern}.}
  \label{fig:vggtfailure}
\end{figure}

Two anomalous entries merit brief explanation: VGGT PC-MMD achieves $d{=}+1.31$ on Gaussian noise due to RBF kernel saturation, \& SED reports $d{=}+161.74$ as SIFT finds no keypoints on noise images (see app.~\Cref{app:robustness-details}). Among external baselines, PRISM-MMD handles noise well ($d{=}+4.71$, 100\% win) but fails on cross-scene mixtures, while TSED performs moderately (54\% overall).

\paragraph{Implications.}
No neural metric handles all inconsistency types. Distributional aggregation reliably improves $L_3$, but $L_1$/$L_2$ remain difficult, and noise cannot be resolved when residuals are uniformly near zero. These failures arise before aggregation; next, we trace them to hallucinated geometry from the underlying 3D backbones.

\section{Why Neural Metrics Fail: Hallucinated Geometric Support}
\label{sec:hallucination}
The robustness results show that distributional aggregation improves neural
metrics, but does not eliminate their hardest failures. The reason is that many
errors originate before aggregation. Metrics such as MEt3R first obtain geometry
from a learned reconstruction backbone and then score feature agreement on the
induced correspondences. If the backbone produces unsupported geometry, the
metric is already evaluating residuals on a false set of cross-view matches.

\textbf{This is a core finding of our analysis}. Learned 3D backbones, including
VGGT~\cite{wang2025vggt}, DUSt3R~\cite{dust3r_cvpr24},
MASt3R~\cite{leroy2024grounding}, and Fast3R~\cite{Yang_2025_Fast3R}, can
return nontrivial 3D structure for inputs with no valid multiview geometry:
unrelated scene mixtures, repeated images, and random noise
(\Cref{fig:vggtfailure,fig:teaser}). The overlap diagnostic makes this concrete:
in the $L_3$ setting, roughly 89\% of view pairs are cross-scene pairs and should
have zero overlap, yet DUSt3R reports zero overlap for only 26\% of pairs,
fabricating shared geometry for a large fraction of unrelated views
(\Cref{tab:overlap_gap}).

The residuals then inherit these unsupported correspondences. DINOv2/FeatUp features are strong semantic descriptors, but they are not identifiers of physical 3D points; visually similar regions from different scenes can therefore produce residuals comparable to true correspondences. On noise, the backbone and
feature extractor fail jointly: learned backbones still produce dense point clouds, while noise patches yield nearly indistinguishable features (\Cref{tab:recon_behavior}). Thus, better aggregation can reduce outlier effects, as in MASt3R-W-IMQ, but cannot fully correct hallucinated geometric support. This motivates classical-based metrics in \Cref{sec:colmap-metrics}, where registration, dense support, and reconstruction failure are explicit signals.

\section{Human Alignment on Real NVS Outputs}
\label{sec:human-alignment}

\paragraph{Evaluation setup.}


We evaluate 8 NVS methods spanning diffusion-based generation
(Stable Virtual Camera~\cite{zhou2025stable}, ViewCrafter~\cite{yu2024viewcrafter},
MVGenMaster~\cite{cao2025mvgenmaster}, NVS-Solver~\cite{you2024nvs},
Difix3D+~\cite{wu2025difix3d+}) and feed-forward regression
(DepthSplat~\cite{xu2025depthsplat}, Long-LRM~\cite{ziwen2025long},
MVSplat360~\cite{chen2024mvsplat360}). We use 33 scenes: 24 from
DL3DV-10K-Benchmark~\cite{ling2024dl3dv} and 9 from
Mip-NeRF360~\cite{barron2022mip}. For each scene, we evaluate sparse
input settings with $K\in\{3,6,9\}$ views. Each method produces a
$360^\circ$ orbital rendering with 50 frames at 10 fps. Trajectory estimation, and inference details, etc. are given in
app.~\Cref{sec:implementation-details}.

\begin{figure}[!htbp]
  \centering
  \includegraphics[width=\columnwidth,interpolate=false]{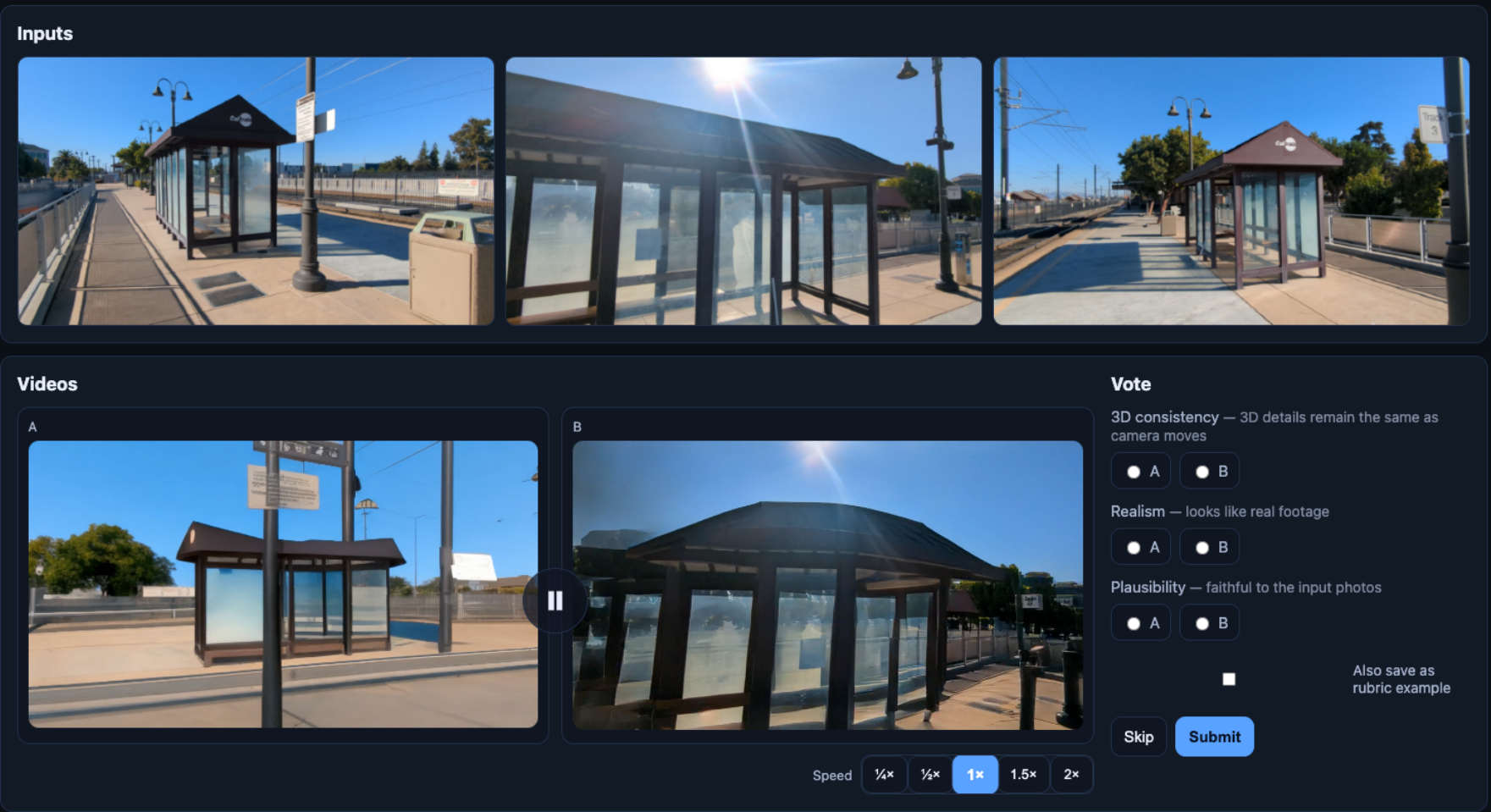}
  \caption{\textbf{Human-evaluation interface.} Participants see the $K$ input
  views (top) alongside two anonymized $360^\circ$ orbit videos (A/B) played in
  a synchronized loop with shared speed controls. Method names are never
  revealed. Each comparison requires forced A/B votes on three axes -- 3D consistency, visual realism, and plausibility (no
  per-axis ties). Method ratings are aggregated with a weighted Elo protocol,
  giving 3D consistency $2\times$ weight and the other two axes $1\times$ each.
  After each vote, the winner stays on the same side for up to three additional
  rounds against new challengers within the same scene/$K$, mainly to reduce cognitive load on the participants. The study collected 959
  pairwise comparisons from 11 participants (including 2 experts in
  sparse-view 3D reconstruction) across $K \in \{3, 6, 9\}$ input views on
  DL3DV and Mip-NeRF360 scenes.}
  \label{fig:human-study-ui}
\end{figure}

\paragraph{Structured Human Alignment Study.}
The \benchmark experiments test robustness under controlled corruptions. We now
ask whether the resulting metrics predict human judgments on real generated
outputs. Participants compare pairs of anonymized NVS videos while viewing the
source images, and vote separately on 3D consistency, visual realism, and
plausibility (\Cref{app:human-eval-details}). We convert pairwise preferences into
method rankings using an Elo-style rating procedure~\cite{glickman1999rating}.
Metric alignment is measured by comparing each metric-induced ranking with the
human ranking for 3D consistency. Full study details and the evaluation interface
(\Cref{fig:human-study-ui}) are in app.~\Cref{app:human-eval-details}.

\begin{table}[h]
  \centering
  \caption{Metric-induced rankings vs.\ human rankings of 8 NVS methods on 24 DL3DV scenes across three view-count splits ($K{=}3, 6, 9$). We compare three 3D consistency metrics based on data-driven reconstruction backbones: MEt3R, PRISM-MMD (PRISM), and MASt3R-W-IMQ (IMQ) - selected for its highest overall win rate against inconsistent scene types (\cref{tab:metric_calibration_effect}). Colors encode ranks 1 (best, \colorbox{green!35}{green}) to 8 (worst, \colorbox{red!35}{red}); rows sorted by $K{=}3$ human rank (\textbf{H}). MASt3R-W-IMQ trails PRISM-MMD marginally at $K{=}3$ but achieves the highest human alignment at $K{=}6$ and $K{=}9$, while MEt3R shows weak or negative correlation at $K{=}3$. Bottom row: Spearman rank correlation~($\rho$) measuring alignment with human ranking.}
  \label{tab:human-rank-corr-dl3dv}
  \resizebox{\columnwidth}{!}{%
    \begin{tabular}{lcccccccccccc}
      \toprule
      Method & \multicolumn{4}{c}{K=3} & \multicolumn{4}{c}{K=6} & \multicolumn{4}{c}{K=9} \\
      \cmidrule(lr){2-5} \cmidrule(lr){6-9} \cmidrule(lr){10-13}
       & H$\downarrow$ & MEt3R$\downarrow$ & IMQ$\downarrow$ & PRISM$\downarrow$ & H$\downarrow$ & MEt3R$\downarrow$ & IMQ$\downarrow$ & PRISM$\downarrow$ & H$\downarrow$ & MEt3R$\downarrow$ & IMQ$\downarrow$ & PRISM$\downarrow$ \\
      \midrule
      DepthSplat & \cellcolor{red!0!green!35}1 & \cellcolor{red!86!green!35}0.318 \textbf{\scriptsize 7} & \cellcolor{red!57!green!35}0.089 \textbf{\scriptsize 5} & \cellcolor{red!57!green!35}0.298 \textbf{\scriptsize 5} & \cellcolor{red!0!green!35}1 & \cellcolor{red!57!green!35}0.304 \textbf{\scriptsize 5} & \cellcolor{red!29!green!35}0.083 \textbf{\scriptsize 3} & \cellcolor{red!57!green!35}0.295 \textbf{\scriptsize 5} & \cellcolor{red!0!green!35}1 & \cellcolor{red!29!green!35}0.298 \textbf{\scriptsize 3} & \cellcolor{red!29!green!35}0.084 \textbf{\scriptsize 3} & \cellcolor{red!57!green!35}0.298 \textbf{\scriptsize 5} \\
      S-V-Camera & \cellcolor{red!14!green!35}2 & \cellcolor{red!0!green!35}0.259 \textbf{\scriptsize 1} & \cellcolor{red!0!green!35}0.065 \textbf{\scriptsize 1} & \cellcolor{red!0!green!35}0.168 \textbf{\scriptsize 1} & \cellcolor{red!14!green!35}2 & \cellcolor{red!0!green!35}0.266 \textbf{\scriptsize 1} & \cellcolor{red!0!green!35}0.069 \textbf{\scriptsize 1} & \cellcolor{red!14!green!35}0.152 \textbf{\scriptsize 2} & \cellcolor{red!29!green!35}3 & \cellcolor{red!0!green!35}0.270 \textbf{\scriptsize 1} & \cellcolor{red!0!green!35}0.068 \textbf{\scriptsize 1} & \cellcolor{red!0!green!35}0.129 \textbf{\scriptsize 1} \\
      Long-LRM & \cellcolor{red!29!green!35}3 & \cellcolor{red!100!green!35}0.375 \textbf{\scriptsize 8} & \cellcolor{red!100!green!35}0.111 \textbf{\scriptsize 8} & \cellcolor{red!86!green!35}0.391 \textbf{\scriptsize 7} & \cellcolor{red!29!green!35}3 & \cellcolor{red!100!green!35}0.396 \textbf{\scriptsize 8} & \cellcolor{red!100!green!35}0.131 \textbf{\scriptsize 8} & \cellcolor{red!86!green!35}0.339 \textbf{\scriptsize 7} & \cellcolor{red!14!green!35}2 & \cellcolor{red!100!green!35}0.381 \textbf{\scriptsize 8} & \cellcolor{red!86!green!35}0.125 \textbf{\scriptsize 7} & \cellcolor{red!86!green!35}0.351 \textbf{\scriptsize 7} \\
      MVGenMaster & \cellcolor{red!43!green!35}4 & \cellcolor{red!43!green!35}0.294 \textbf{\scriptsize 4} & \cellcolor{red!43!green!35}0.079 \textbf{\scriptsize 4} & \cellcolor{red!29!green!35}0.171 \textbf{\scriptsize 3} & \cellcolor{red!57!green!35}5 & \cellcolor{red!14!green!35}0.284 \textbf{\scriptsize 2} & \cellcolor{red!14!green!35}0.074 \textbf{\scriptsize 2} & \cellcolor{red!0!green!35}0.123 \textbf{\scriptsize 1} & \cellcolor{red!57!green!35}5 & \cellcolor{red!14!green!35}0.286 \textbf{\scriptsize 2} & \cellcolor{red!14!green!35}0.077 \textbf{\scriptsize 2} & \cellcolor{red!14!green!35}0.140 \textbf{\scriptsize 2} \\
      ViewCrafter & \cellcolor{red!57!green!35}5 & \cellcolor{red!14!green!35}0.287 \textbf{\scriptsize 2} & \cellcolor{red!14!green!35}0.078 \textbf{\scriptsize 2} & \cellcolor{red!43!green!35}0.187 \textbf{\scriptsize 4} & \cellcolor{red!43!green!35}4 & \cellcolor{red!29!green!35}0.296 \textbf{\scriptsize 3} & \cellcolor{red!43!green!35}0.090 \textbf{\scriptsize 4} & \cellcolor{red!29!green!35}0.200 \textbf{\scriptsize 3} & \cellcolor{red!43!green!35}4 & \cellcolor{red!57!green!35}0.318 \textbf{\scriptsize 5} & \cellcolor{red!43!green!35}0.100 \textbf{\scriptsize 4} & \cellcolor{red!29!green!35}0.204 \textbf{\scriptsize 3} \\
      Difix3D & \cellcolor{red!71!green!35}6 & \cellcolor{red!57!green!35}0.311 \textbf{\scriptsize 5} & \cellcolor{red!86!green!35}0.097 \textbf{\scriptsize 7} & \cellcolor{red!100!green!35}0.459 \textbf{\scriptsize 8} & \cellcolor{red!100!green!35}8 & \cellcolor{red!86!green!35}0.337 \textbf{\scriptsize 7} & \cellcolor{red!86!green!35}0.118 \textbf{\scriptsize 7} & \cellcolor{red!100!green!35}0.400 \textbf{\scriptsize 8} & \cellcolor{red!100!green!35}8 & \cellcolor{red!86!green!35}0.356 \textbf{\scriptsize 7} & \cellcolor{red!100!green!35}0.131 \textbf{\scriptsize 8} & \cellcolor{red!100!green!35}0.388 \textbf{\scriptsize 8} \\
      NVS-Solver & \cellcolor{red!86!green!35}7 & \cellcolor{red!29!green!35}0.290 \textbf{\scriptsize 3} & \cellcolor{red!29!green!35}0.079 \textbf{\scriptsize 3} & \cellcolor{red!14!green!35}0.169 \textbf{\scriptsize 2} & \cellcolor{red!86!green!35}7 & \cellcolor{red!43!green!35}0.302 \textbf{\scriptsize 4} & \cellcolor{red!57!green!35}0.101 \textbf{\scriptsize 5} & \cellcolor{red!43!green!35}0.224 \textbf{\scriptsize 4} & \cellcolor{red!86!green!35}7 & \cellcolor{red!43!green!35}0.307 \textbf{\scriptsize 4} & \cellcolor{red!71!green!35}0.115 \textbf{\scriptsize 6} & \cellcolor{red!43!green!35}0.228 \textbf{\scriptsize 4} \\
      MVSplat360 & \cellcolor{red!100!green!35}8 & \cellcolor{red!71!green!35}0.318 \textbf{\scriptsize 6} & \cellcolor{red!71!green!35}0.096 \textbf{\scriptsize 6} & \cellcolor{red!71!green!35}0.325 \textbf{\scriptsize 6} & \cellcolor{red!71!green!35}6 & \cellcolor{red!71!green!35}0.331 \textbf{\scriptsize 6} & \cellcolor{red!71!green!35}0.107 \textbf{\scriptsize 6} & \cellcolor{red!71!green!35}0.331 \textbf{\scriptsize 6} & \cellcolor{red!71!green!35}6 & \cellcolor{red!71!green!35}0.341 \textbf{\scriptsize 6} & \cellcolor{red!57!green!35}0.111 \textbf{\scriptsize 5} & \cellcolor{red!71!green!35}0.317 \textbf{\scriptsize 6} \\
      \midrule
      Rank Corr.\ ($\rho$)$\uparrow$ & -- & -0.10 & 0.14 & \textbf{0.19} & -- & 0.26 & \textbf{0.48} & 0.31 & -- & 0.24 & \textbf{0.48} & 0.24 \\
      \bottomrule
    \end{tabular}%
  }
\end{table}

\paragraph{Neural metric alignment.}
\Cref{tab:human-rank-corr-dl3dv} compares learned and external metrics against
human rankings on the 24 DL3DV scenes. We select MASt3R-W-IMQ because it has
the highest overall robustness win rate on \benchmark. It improves over MEt3R
at every view count, reaching Spearman $\rho=0.14$ at $K=3$ and $\rho=0.48$
at $K=6,9$. Alignment remains moderate: the automatic metrics rank Stable
Virtual Camera highly, while humans prefer DepthSplat for 3D consistency,
and several mid-ranked methods show large inversions. This matches our
robustness analysis: better aggregation improves learned metrics, but the
learned backbone can still provide unsupported geometric support.

\begin{table}[h]
  \centering
  \caption{COLMAP reconstruction quality vs.\ human rankings of 8 NVS methods on DL3DV (24 scenes) and MipNeRF360 (9 scenes) for the $K{=}9$ view split. Weighted GPC measures geometric photo-consistency weighted by angular coverage; ICM is the image-consistency metric; Ang.\ Cov.\ measures angular coverage of registered views. Colors encode ranks 1 (best, \colorbox{green!35}{green}) to 8 (worst, \colorbox{red!35}{red}); rows sorted by DL3DV human rank (\textbf{H}). Bottom row reports Spearman correlation $\rho$ between human and metric-induced method rankings.}
  \label{tab:colmap-human-rank}
  \resizebox{\columnwidth}{!}{%
    \begin{tabular}{lcccccccc}
      \toprule
      Method & \multicolumn{4}{c}{DL3DV} & \multicolumn{4}{c}{MipNeRF360} \\
      \cmidrule(lr){2-5} \cmidrule(lr){6-9}
       & H$\downarrow$ & W-GPC$\uparrow$ & ICM$\uparrow$ & Ang.\ Cov.$\uparrow$ & H$\downarrow$ & W-GPC$\uparrow$ & ICM$\uparrow$ & Ang.\ Cov.$\uparrow$ \\
      \midrule
      DepthSplat & \cellcolor{red!0!green!35}1 & \cellcolor{red!0!green!35}0.288 \textbf{\scriptsize 1} & \cellcolor{red!0!green!35}0.531 \textbf{\scriptsize 1} & \cellcolor{red!0!green!35}170.4 \textbf{\scriptsize 1} & \cellcolor{red!0!green!35}1 & \cellcolor{red!0!green!35}0.475 \textbf{\scriptsize 1} & \cellcolor{red!0!green!35}0.636 \textbf{\scriptsize 1} & \cellcolor{red!0!green!35}247.0 \textbf{\scriptsize 1} \\
      Long-LRM & \cellcolor{red!14!green!35}2 & \cellcolor{red!14!green!35}0.136 \textbf{\scriptsize 2} & \cellcolor{red!14!green!35}0.404 \textbf{\scriptsize 2} & \cellcolor{red!14!green!35}94.1 \textbf{\scriptsize 2} & \cellcolor{red!14!green!35}2 & \cellcolor{red!14!green!35}0.128 \textbf{\scriptsize 2} & \cellcolor{red!14!green!35}0.378 \textbf{\scriptsize 2} & \cellcolor{red!14!green!35}79.3 \textbf{\scriptsize 2} \\
      S-V-Camera & \cellcolor{red!29!green!35}3 & \cellcolor{red!29!green!35}0.056 \textbf{\scriptsize 3} & \cellcolor{red!29!green!35}0.351 \textbf{\scriptsize 3} & \cellcolor{red!29!green!35}53.4 \textbf{\scriptsize 3} & \cellcolor{red!43!green!35}4 & \cellcolor{red!29!green!35}0.021 \textbf{\scriptsize 3} & \cellcolor{red!29!green!35}0.363 \textbf{\scriptsize 3} & \cellcolor{red!57!green!35}21.1 \textbf{\scriptsize 5} \\
      ViewCrafter & \cellcolor{red!43!green!35}4 & \cellcolor{red!43!green!35}0.036 \textbf{\scriptsize 4} & \cellcolor{red!86!green!35}0.238 \textbf{\scriptsize 7} & \cellcolor{red!43!green!35}46.2 \textbf{\scriptsize 4} & \cellcolor{red!29!green!35}3 & \cellcolor{red!43!green!35}0.017 \textbf{\scriptsize 4} & \cellcolor{red!57!green!35}0.275 \textbf{\scriptsize 5} & \cellcolor{red!29!green!35}25.5 \textbf{\scriptsize 3} \\
      MVGenMaster & \cellcolor{red!57!green!35}5 & \cellcolor{red!57!green!35}0.035 \textbf{\scriptsize 5} & \cellcolor{red!71!green!35}0.302 \textbf{\scriptsize 6} & \cellcolor{red!57!green!35}42.2 \textbf{\scriptsize 5} & \cellcolor{red!57!green!35}5 & \cellcolor{red!57!green!35}0.015 \textbf{\scriptsize 5} & \cellcolor{red!71!green!35}0.267 \textbf{\scriptsize 6} & \cellcolor{red!43!green!35}25.2 \textbf{\scriptsize 4} \\
      MVSplat360 & \cellcolor{red!71!green!35}6 & \cellcolor{red!71!green!35}0.027 \textbf{\scriptsize 6} & \cellcolor{red!57!green!35}0.319 \textbf{\scriptsize 5} & \cellcolor{red!86!green!35}27.2 \textbf{\scriptsize 7} & \cellcolor{red!71!green!35}6 & \cellcolor{red!71!green!35}0.013 \textbf{\scriptsize 6} & \cellcolor{red!43!green!35}0.356 \textbf{\scriptsize 4} & \cellcolor{red!86!green!35}13.5 \textbf{\scriptsize 7} \\
      NVS-Solver & \cellcolor{red!86!green!35}7 & \cellcolor{red!100!green!35}0.018 \textbf{\scriptsize 8} & \cellcolor{red!43!green!35}0.350 \textbf{\scriptsize 4} & \cellcolor{red!100!green!35}18.9 \textbf{\scriptsize 8} & \cellcolor{red!100!green!35}8 & \cellcolor{red!100!green!35}0.003 \textbf{\scriptsize 8} & \cellcolor{red!100!green!35}0.142 \textbf{\scriptsize 8} & \cellcolor{red!100!green!35}13.1 \textbf{\scriptsize 8} \\
      Difix3D & \cellcolor{red!100!green!35}8 & \cellcolor{red!86!green!35}0.019 \textbf{\scriptsize 7} & \cellcolor{red!100!green!35}0.237 \textbf{\scriptsize 8} & \cellcolor{red!71!green!35}32.6 \textbf{\scriptsize 6} & \cellcolor{red!86!green!35}7 & \cellcolor{red!86!green!35}0.010 \textbf{\scriptsize 7} & \cellcolor{red!86!green!35}0.203 \textbf{\scriptsize 7} & \cellcolor{red!71!green!35}16.1 \textbf{\scriptsize 6} \\
      \midrule
      Rank Corr.\ ($\rho$)$\uparrow$ & -- & \textbf{0.98} & 0.76 & 0.93 & -- & \textbf{0.98} & 0.88 & 0.95 \\
      \bottomrule
    \end{tabular}%
  }
\end{table}

\paragraph{COLMAP metric alignment.}
\Cref{tab:colmap-human-rank} reports COLMAP-based metrics on the $K=9$ split.
W-GPC is our primary score, combining dense geometric--photometric agreement
with angular coverage; ICM complements it by measuring total reliable dense
3D mass. These classical verification metrics align more strongly with human
3D-consistency judgments than learned-backbone metrics, with W-GPC achieving
up to $4\times$ higher correlation than MEt3R. Their behavior on \benchmark
is consistent with failure-aware evaluation: noise inputs and cross-scene
mixtures receive zero or near-zero support because COLMAP cannot verify a
coherent multiview reconstruction. This strictness is useful for evaluating
geometrically consistent NVS, but can make comparisons coarse when all outputs
are poor. We therefore report registration rate and ICM alongside W-GPC, and
use MASt3R-W-IMQ as a secondary learned diagnostic when COLMAP cannot be run.
\FloatBarrier
\section{Discussion}
\label{sec:discussion}

Our results highlight a reliability issue in using learned 3D reconstruction models as automatic evaluators, or generally, blindly trusting such models' reconstructions. A ground-truth-free metric based on a reconstruction backbone is only as reliable as the geometric support produced by that backbone. When the backbone returns dense geometry for an image set that does not admit a coherent 3D explanation, the downstream metric can assign a favorable score to an invalid input. This behavior is not simply a small aggregation artifact, as it changes what the metric is measuring.

\textbf{The central empirical finding of this work is that recent learned 3D backbones, including VGGT, DUSt3R, MASt3R, and Fast3R, can hallucinate} geometric support, especially when dealing with outliers and noise. On cross-scene mixtures, repeated images, and random noise, these models often produce nontrivial 3D structure and apparent cross-view consistency. To our knowledge, \benchmark is the first controlled benchmark to isolate this failure mode for modern neural 3D evaluators. The backbone--residual--aggregation decomposition is useful precisely because it separates two effects: distributional aggregation can make neural metrics substantially more robust than MEt3R, but it cannot fully correct unsupported geometry produced by the reconstruction model.

Classical geometry provides a complementary failure model. COLMAP is slower and can fail on low-texture, specular, repetitive, or low-overlap imagery. However, its failures are explicit: matches are rejected, cameras fail to register, dense maps are absent, or viewpoint coverage is limited. For evaluation, these outcomes are informative. They expose whether the image set provides enough evidence for a rigid multiview reconstruction. Our COLMAP-based metrics convert this evidence into scene-level consistency scores and obtain the strongest alignment with human judgments in our experiments.

These results argue for caution when learned 3D backbones are used as evaluators or as sources of geometric evidence in uncurated settings. They are powerful reconstruction models, and our best neural metric shows that they can support useful evaluation when combined with better aggregation. But under noise, outliers, or distribution shift, their scores should be interpreted with safeguards such as explicit geometric verification, calibrated uncertainty, or human review. This is especially important in safety-critical applications such as forensic reconstruction, medical reconstruction, and autonomous driving, where hallucinated geometry may be mistaken for evidence. For future work, we note that although there has been a surge of large, data-driven 3D reconstruction models (such as VGGT), there is still no model that can reliably serve as a drop-in replacement for COLMAP.

{
    \small
    \bibliographystyle{ieeenat_fullname}
    \bibliography{main}
}

\clearpage
\appendix

\appendixcontents

\section{Notation and Common Terms}
\label{app:notation-terms}
\begin{center}
\small
\begin{tabular}{p{0.23\linewidth} p{0.71\linewidth}}
\toprule
\textbf{Symbol / Term} & \textbf{Meaning} \\
\midrule
\benchmark\ & Systematically Controlled Outlier and Noise 3D benchmark used to test metric robustness under controlled 3D inconsistency. \\
$L_0$ & Consistent multi-view set from a single real scene. \\
$L_1$ & Single-outlier mixture: $K{-}1$ views from one scene plus 1 foreign view. \\
$L_2$ & Controlled mixture with roughly 30\% foreign views, giving about 50--60\% cross-scene pairs for $K \ge 6$. \\
$L_3$ & Random mixture with views sampled across scenes, giving about 89\% cross-scene pairs on average. \\
SysCON3D-M & Mixture subset of \benchmark: $L_1$, $L_2$, $L_3$\\
SysCON3D-N & Noise subset of \benchmark: patched Gaussian, Gaussian noise, and uniform noise. \\
$\mathrm{FNR}_{\mathrm{cross}}$ & Cross-scene overlap hallucination rate, $P(\mathrm{overlap}\mid\mathrm{cross\mbox{-}scene})$. Ideal value is $0$. \\
Ghost mass & Fraction of rendered pixels supported by wrong-scene points. Reported for Fast3R and VGGT. \\
$B,\rho,\mathcal{A}$ & Unified metric components from~\cref{sec:neural-family}: reconstruction backbone, residual function, and aggregation function. \\
W / PC & Warp-based residuals versus point-consistency residuals. \\
Base / MMD / IMQ / Energy & Mean aggregation and the three distributional aggregations used throughout the learned metric family. \\
W-GPC & Coverage-weighted geometric-photometric consistency, our primary COLMAP-based score. Higher is better. \\
ICM & Integrated consistency mass, the total amount of reliable dense 3D support recovered by COLMAP. Higher is better. \\
Registration rate & Fraction of attempted images that COLMAP successfully registers. Higher is better. \\
\bottomrule
\end{tabular}
\end{center}
\clearpage

\section{FAQs}
\label{app:faq}


\begin{enumerate}
    \item \textbf{What does SysCON3D test?}

    SysCON3D is a controlled stress test for 3D consistency metrics. It is not intended to model the full distribution of real NVS failures. Instead, it tests a basic reliability property: image sets with weaker or contradictory evidence for a shared 3D scene should receive worse consistency scores than clean multiview sets. The corruptions are deliberately simple so that the expected ordering is known.

    \item \textbf{Are cross-scene mixtures and Gaussian noise too artificial?}

    They are diagnostic rather than naturalistic corruptions. Their purpose is to test whether a metric can reject inputs that should not support a coherent multiview reconstruction. This is analogous to using controlled corruptions to expose robustness failures before trusting a metric on real generated outputs. Further, these inputs can occur, in practice, in noisy input situations, for example data scraped from the internet, or adversarial setups.

    \item \textbf{Why include repeated views if they may come from the same scene?}

    Repeated views are not necessarily inconsistent with the scene identity, but they are degenerate as multiview evidence. A metric for multiview 3D consistency should not treat repeated or near-identical frames as strong evidence for a well-supported 3D reconstruction. We therefore use repeated views as a degenerate diagnostic rather than as a cross-scene corruption. Further, we have observed such outputs in NVS methods when they failed to generalize to the input.

    \item \textbf{What do we mean by hallucinated geometric support?}

    We use this term when a reconstruction backbone produces nontrivial 3D structure, overlap, or correspondences for inputs that do not provide valid multiview evidence, such as unrelated scenes or random noise. The issue is not merely that the reconstruction looks visually odd; it is that downstream metrics then compute residuals on unsupported correspondences.

    \item \textbf{Is the failure caused by aggregation or by the reconstruction backbone?}

    Both matter, but our results show that the hardest failures often originate before aggregation. Distributional aggregation improves robustness over the MEt3R-style mean, showing that metric design matters. However, if the backbone has already hallucinated unsupported geometry, the residuals are computed on an invalid set of correspondences, which no scalar aggregation can fully repair.

    \item \textbf{Why does MASt3R-W-IMQ improve robustness but still fail on noise?}

    MASt3R-W-IMQ uses distributional aggregation and is more sensitive to outliers and residual-distribution shape than a mean score. This improves separation for cross-scene mixtures. On Gaussian noise, however, learned backbones can still produce dense geometry, and feature residuals can become uniformly small or uninformative. In that setting, the residual distribution itself lacks the signal needed for aggregation to recover the correct ordering.

    \item \textbf{Why use DINOv2/FeatUp features if they can fail on cross-scene matches?}

    We use DINOv2/FeatUp because they are part of the MEt3R-style evaluation pipeline and provide strong dense semantic features. Our analysis shows a limitation of this choice for 3D consistency: semantic similarity is not the same as physical point identity. Visually similar regions from different scenes can yield low residuals even when the underlying correspondence is geometrically invalid.

    \item \textbf{Does COLMAP simply win because it fails to reconstruct hard cases?}

    No. In our COLMAP-based metrics, failed registration or densification is counted as zero verified 3D support, not as a missing or ignored value. This makes failure part of the consistency score. The metric therefore measures verified reconstructability under classical multiview geometry: registered cameras, dense support, depth agreement, and viewpoint coverage.

    \item \textbf{Is COLMAP a perfect 3D consistency metric?}

    No. COLMAP can fail on low-texture, specular, repetitive, blurred, or low-overlap imagery even when the scene is physically valid. We therefore interpret COLMAP-based metrics as failure-aware geometric verification metrics, not as universal ground truth. Their value is that failure modes are observable and measurable, unlike learned backbones that may always return geometry.

    \item \textbf{Why report W-GPC, ICM, and registration rate together?}

    W-GPC combines dense geometric--photometric agreement with angular coverage and is our primary COLMAP-based score. ICM measures the total amount of reliable dense 3D mass. Registration rate reports sparse geometric support. Reporting them together separates different failure modes: insufficient sparse support, poor dense agreement, low 3D mass, and narrow viewpoint coverage.

    \item \textbf{When should MASt3R-W-IMQ be used instead of COLMAP-based metrics?}

    MASt3R-W-IMQ is useful as a fast learned diagnostic, especially when COLMAP cannot be run or when inputs are known to be clean and free of outliers. It should not override a complete COLMAP failure. If COLMAP finds little geometric support but a learned metric reports high consistency, that sample should be flagged for human review. Also, if a robust 3D reconstruction backbone becomes available in future, MASt3R-W-IMQ will be a viable and sufficient metric.

    \item \textbf{Are PRISM, SED, and TSED directly comparable to our metrics?}

    They are included as external baselines with different assumptions. PRISM uses a learned object-centric embedding distribution and an anchor set. SED/TSED use explicit epipolar verification from SIFT correspondences and require known or estimated fundamental matrices. We include them to contextualize robustness, but our main comparison is among scene-level, ground-truth-free multiview consistency metrics.

    \item \textbf{Why evaluate human alignment over method-level rankings?}

    The human study asks which NVS methods produce more 3D-consistent outputs, so we compare metric-induced method rankings to human method rankings. The correlations are therefore computed over $n=8$ methods. This is a compact validation of whether metrics agree with human judgments at the level used for model comparison; it is not a per-frame perceptual-quality study.

    \item \textbf{Why separate 3D consistency, realism, and plausibility in the human study?}

    A generated video can be realistic but geometrically inconsistent, or geometrically stable but visually imperfect. We ask participants to judge these axes separately so that metric alignment is evaluated against the intended construct: scene-level 3D consistency rather than general visual appeal.

    \item \textbf{Do these results mean learned 3D backbones are poor reconstruction models?}

    No. The result concerns their use as automatic evaluators or sources of geometric evidence under corruptions. Learned backbones are powerful reconstruction models and can support useful metrics when paired with better residual aggregation. The failure arises when they return confident-looking geometry for inputs that provide little or no valid multiview support.

    \item \textbf{What is the main practical recommendation?}

    For robust evaluation of NVS outputs, we recommend reporting W-GPC with ICM and registration rate. MASt3R-W-IMQ should be reported as a complementary learned diagnostic. Large disagreement between COLMAP-based verification and learned-backbone metrics is itself informative and should trigger inspection, especially in noisy or uncurated settings.

    \item \textbf{Why is hallucinated geometry a problem?}

    Learned reconstruction models can use priors to complete geometry when the image evidence is weak. This is useful for reconstruction, but it becomes risky when the same model is used as an evaluator. If the backbone invents 3D structure or cross-view overlap, a metric built on that backbone will compute residuals on unsupported correspondences and may score an invalid view set as consistent. This can distort NVS benchmark rankings and can be dangerous in applications where geometry is treated as evidence, such as forensic or crime-scene reconstruction, medical reconstruction, autonomous driving, and robotics. In such settings, learned 3D predictions should be checked with geometric verification, calibrated uncertainty, or human review.

     \item \textbf{Why not apply RANSAC-style geometric checks to matches predicted by VGGT, DUSt3R, or MASt3R?}

Classical matching separates evidence from verification: SIFT proposes discrete matches, and RANSAC tests whether they support a geometry. Learned 3D backbones predict geometry, cameras, and correspondences together, so their matches are already tied to the model's own reconstruction. If the model hallucinates a plausible scene, a geometric check can end up validating that internal reconstruction rather than independent image evidence. COLMAP gives a separate verification signal through matches, inliers, registration, dense support, and viewpoint coverage; when these fail, the view set lacks verified multiview consistency.

\end{enumerate}

\section{Detailed Contributions}
\label{app:contributions}

\begin{enumerate}
    \item We introduce a \textbf{unified parametric framework} for ground-truth-free 3D consistency metrics, showing that existing and new neural metrics can be decomposed into three components: a \emph{reconstruction backbone}, a \emph{residual function}, and an \emph{aggregation function}. This formulation clarifies the design space of multi-view consistency metrics and enables principled variants beyond pairwise mean aggregation.

    \item We construct \textbf{SysCON3D}, a robustness benchmark for 3D consistency evaluation, which systematically injects controlled cross-scene outliers and synthetic corruptions into multi-view image sets. This benchmark enables direct testing of whether a metric can distinguish geometrically consistent scenes from increasingly inconsistent ones.

    \item Through SysCON3D, we uncover a \textbf{systematic and previously underappreciated failure mode} of modern data-driven 3D reconstruction backbones: rather than rejecting impossible inputs, they hallucinate non-trivial 3D structure and spurious cross-view consistency for cross-scene mixtures, repeated images, and random noise. Consequently, evaluation metrics built on top of these learned backbones can inherit these biases and become unreliable.

    \item To address this limitation, we develop \textbf{robust COLMAP-based 3D consistency metrics} that avoid learned priors and instead rely on classical geometric verification. These metrics provide a more interpretable, failure-aware, and robust measure of scene-level multi-view consistency.

    \item We design a \textbf{structured human preference study} for evaluating 3D consistency, with explicit protocols that distinguish 3D consistency from visual realism and plausibility. This yields a more targeted human reference for assessing how well automatic metrics align with human judgments of scene-level consistency.

    \item We perform a \textbf{comprehensive empirical evaluation} on SysCON3D, Mip-NeRF360, and DL3DV, together with our human study, and show that the proposed metrics substantially improve robustness over prior work while also aligning more closely with human judgments. In particular, neural distributional metrics improve over MEt3R, and COLMAP-based metrics achieve the strongest human alignment among the evaluated metrics in our structured analysis.
\end{enumerate}

\section{Linear-Time PC Residuals and Two-Sample MMD}
\label{app:method-details}

\paragraph{Linear-time PC residuals.}
\label{app:pc-feature-sum}
PC residuals scale as $O(K)$ in feature lookups -- one pass per view -- compared to $O(K^2)$ for warp-based residuals which iterate over all $\binom{K}{2}$ pairs. We further avoid explicitly enumerating all $\binom{|\mathcal{V}_n|}{2}$ pairs per point by accumulating the feature sum $\mathbf{s}_n = \sum_{k \in \mathcal{V}_n} \hat{f}_k$ of $\ell_2$-normalized features and computing the mean pairwise similarity as $(\|\mathbf{s}_n\|^2 - |\mathcal{V}_n|) / (|\mathcal{V}_n|(|\mathcal{V}_n|-1))$, reducing per-point cost from $\mathcal{O}(|\mathcal{V}_n|^2)$ to $\mathcal{O}(|\mathcal{V}_n|)$. The PC residual $\delta_n$ in Eq.~\ref{eq:point-dispersion} then follows by replacing similarities with cosine dissimilarities.

\paragraph{Two-sample MMD with a reference set.}
\label{app:mmd-two-sample}
The aggregation in ~\Cref{sec:neural-family} compares the empirical residual distribution $P$ against $\delta_0$, the Dirac delta at zero (perfect consistency). When ground-truth images $\mathcal{I}^{\text{gt}}$ are available, $\delta_0$ can be replaced by the empirical residual distribution $Q$ obtained by running the same backbone--residual pipeline on $\mathcal{I}^{\text{gt}}$, yielding a two-sample test. For MMD:
\begin{equation}
  \widehat{\text{MMD}}^2(P, Q)
  = \frac{1}{N(N-1)}\!\sum_{a\neq b}\! k(e_a, e_b)
  - \frac{2}{NM}\!\sum_{a,c}\! k(e_a, e'_c)
  + \frac{1}{M(M-1)}\!\sum_{c\neq d}\! k(e'_c, e'_d),
  \label{eq:mmd-two-sample}
\end{equation}
where $\{e_a\}_{a=1}^N \sim P$ and $\{e'_c\}_{c=1}^M \sim Q$ are residuals from the generated and reference sets, respectively. IMQ kernel MMD and energy distance admit analogous two-sample forms.

\section{How to Use Our Metrics to Evaluate 3D Consistency}
\label{app:how-to-use}

We recommend \textbf{W-GPC} as the primary COLMAP-based metric because it combines dense geometric-photometric agreement with viewpoint coverage. When all compared methods are evaluated on the same target views, ICM is a useful complementary measure of total reliable 3D mass and should be reported alongside W-GPC rather than replacing it. Very low registration rate typically indicates that the generated view set is not reconstructable under classical multi-view geometry, in which case COLMAP-based scores naturally approach zero. In such cases, MASt3R-W-IMQ may be reported as a secondary diagnosis, but it should not override a COLMAP failure, as learned backbones can assign spuriously favorable scores to noisy or inconsistent inputs.

Notably, MASt3R-W-IMQ can still be used for evaluation if we are certain that the input images contain no noise or outliers, or if the COLMAP pipeline cannot be run for any unusual reason. Further, if a new robust 3D reconstruction backbone becomes available that does not hallucinate 3D like VGGT \etal, MASt3R-W-IMQ becomes a more robust measure. In practice, the registration rate is the threshold we use to decide whether to fall back on neural metrics: a low registration rate usually means the input views are largely or completely 3D-inconsistent, and COLMAP scores tend to zero, in which case the neural variant can still be reported, with the caveat above.

\section{Human Evaluation Study: Implementation Details}
\label{app:human-eval-details}


This section provides implementation details for the pairwise human evaluation described in \cref{sec:human-alignment}. The study comprises 959 pairwise comparisons from 11 participants, including 2 experts in sparse-view 3D reconstruction. For each view-count setting $K \in {3, 6, 9}$, participants compare two anonymized videos from the same scene and vote on three axes: 3D consistency, visual realism, and plausibility. \emph{3D consistency} asks whether the rendered views look like they all come from a single coherent 3D scene rather than the scene morphing or drifting as the camera moves. \emph{Visual realism} asks whether each individual frame looks like a real, sharp photograph rather than a blurry, artifact-ridden, or AI-generated image. \emph{Plausibility} asks whether the synthesized frames at novel views and input observations appear to come from the same 3D scene. We aggregate these votes using a weighted Elo protocol, assigning 2 points to 3D consistency and 1 point to each of the other two axes.

\paragraph{Platform and interface.}
The study uses a static GitHub Pages frontend, a Cloudflare Worker backend, a D1 (SQLite) database, and Cloudflare R2 for input images and videos. Participants see the $K$ input views, along with two anonymized orbit videos labeled A and B (\cref{fig:human-study-ui}); method names are never exposed, and asset URLs use obfuscated keys. The frontend keeps both videos synchronized, loops them automatically, offers shared speed controls, and speeds up long clips to keep each comparison short. A manual with rubric examples is available during the session.

\paragraph{Matchup sampling and session flow.}
Sampling is coverage-biased at two levels. For a chosen view count $K \in \{3, 6, 9\}$, the backend first prefers scenes with the fewest completed games. Within that scene, it prefers methods with the fewest prior appearances, excludes rubric pairs and pairs already seen by the same participant, and randomizes the left-right assignment. After each vote, the winner enters a winner-stays block of at most three rounds: the same scene and $K$ are kept, the winner stays on the same side, and a new challenger is drawn from the least-played unseen methods for that scene. Participants can break the block with \texttt{Skip} or \texttt{New pair}; \texttt{Skip} clears the block and does not log a game. When a block ends, the frontend rotates to the next $K$ value to spread effort across 3, 6, and 9 views. Only after a participant exhausts all unique pairs at a given $K$ does the backend allow repeats.

\paragraph{Voting and Elo updates.}
Each comparison requires forced A/B choices on three axes: 3D consistency, realism, and plausibility. There are no per-axis ties. The Elo outcome is binary: the side that wins at least two of the three axes wins (and stays on for up to 3 rounds), and the other side loses, so the game itself has no draw state. However, when updating ratings for the participating methods, we assign 2$\times$ weight to 3D consistency and 1$\times$ weight to realism and plausibility, so ties are possible in the final scoring system. Ratings follow standard Elo~\cite{glickman1999rating}: given current ratings $R_A$ and $R_B$, the expected score is $E_A = 1 / (1 + 10^{(R_B - R_A)/400})$, and ratings update as $R'_A = R_A + 32 \cdot (S_A - E_A)$. All methods start with a base rating of 500. Dataset-specific leaderboards are computed by replaying the logged games after filtering to the relevant dataset.

\paragraph{Rubric examples and logging.}
Admins can run a separate rubric mode or promote an ordinary comparison to a rubric example, but only after writing reasons for all three axes. These examples populate the manual/examples panel shown to participants and are excluded from future matchup sampling to prevent calibration content from distorting coverage. All votes are timestamped, and participants can later review, edit, or delete their past submissions; leaderboards are then recomputed from the updated game log.

\paragraph{Statistics}
The study is based on 959 games played across 11 participants, involving 2 experts in sparse-view 3D reconstruction. The top-4 methods for a specific dataset and view split are usually involved in 20-30\% more games than the bottom-4 methods, as the winner of each game stays on for a maximum of 3 rounds to reduce cognitive load on participants.

\paragraph{Note.}
Human evaluation in our setting is subject to known perceptual biases. In particular, prior work suggests that observers can be disproportionately influenced by high-frequency structures such as edges and corners~\cite{langerman2025explaining}, and may also miss important inconsistencies due to inattentional blindness~\cite{simons1999gorillas}. As a result, annotators may occasionally underweight background 3D errors or over-prefer generations with sharper local detail despite poorer global 3D consistency. To reduce this source of noise, we explicitly instructed participants to prioritize \emph{3D consistency} and presented counterexamples highlighting such failure modes. Finally, we observe that human discrimination also degrades when all candidate generations are of very low quality: ranking among several poor 3D outputs becomes intrinsically error-prone, similar to the behavior of our COLMAP-based metrics under severely degraded inputs.

\section{Implementation Details: Adapting NVS Methods to Sparse-View Orbits}
\label{sec:implementation-details}

Adapting the 8 NVS methods (DepthSplat, MVSplat360, Long-LRM, Difix3D, MVGenMaster, NVS Solver, Stable Virtual Camera) to produce \threesixty orbital renderings from only $P \in \{3,6,9\}$ sparse input images required method-specific modifications, which we describe below. One component - the orbital camera trajectory - is shared across all methods and is presented first.

\subsection{Common: Orbital Trajectory from Sparse Cameras}
\label{subsec:orbit-trajectory}

All methods follow the same procedure to convert a sparse set of calibrated cameras into a smooth \threesixty orbital path. Given camera centers $\{c_i\}$ and unit forward directions $\{v_i\}$, we first estimate a shared look-at point $p^*$ by minimizing the sum of squared perpendicular distances from each camera ray:
\begin{equation}
    p^* = \arg\min_{p} \sum_{i=1}^{P} \left\| (I - v_i v_i^\top)(p - c_i) \right\|^2.
\end{equation}
Setting the gradient to zero gives the $3\times 3$ linear system
\begin{equation}
    \underbrace{\left(\sum_{i=1}^{P} I - v_i v_i^\top\right)}_{A}\; p^*
    \;=\;
    \underbrace{\sum_{i=1}^{P} (I - v_i v_i^\top)\, c_i}_{b},
\end{equation}
which we solve as $p^* = A^{-1}b$ (or via least-squares when $A$ is rank-deficient). We then estimate an up direction $u$ from the mean camera orientation, compute a median orbital radius $r$ and height $h$ from the camera positions projected onto the plane perpendicular to $u$, and generate $N$ evenly spaced orbit poses via Rodrigues rotation of an initial planar direction $\hat{d}_0$ about $u$:
\begin{equation}
    c(\theta) = p^* + r \left[\cos\theta\,\hat{d}_0 + \sin\theta\,(u \times \hat{d}_0)\right] + h\,u, \qquad \theta \in [0, 2\pi).
\end{equation}
Each orbit camera looks back at $p^*$, producing a continuous azimuthal sweep even when the inputs cover a narrow angular range.

\subsection{DepthSplat.}
DepthSplat required no architectural or algorithmic modifications.
Because the model is trained on randomly sampled context sets of varying
size, it directly accepts 3, 6, or 9 input views and renders arbitrary
target cameras - including the orbit trajectory - without retraining or
fine-tuning. The only practical consideration is checkpoint selection: a model trained on 2-6 context views is used for the 3 and 6-view settings, while one
trained on 4-10 views is used for 9 views, to match the training
distribution.

\subsection{MVSplat360.}
MVSplat360 pairs a Gaussian-splatting encoder with a video-diffusion
refiner~(SVD) that enhances the rendered frames. Two modifications were needed to apply this refiner to a long orbit sequence.

First, the SVD model accepts a fixed window of $W\!=\!28$ frames, whereas
our orbits contain $T\!=\!50$ frames.  We divide the sequence into
overlapping windows with a stride of $W - \delta$ ($\delta\!=\!6$ frames of
overlap).  Each window is refined independently.  Where two windows cover
the same frame, we keep the prediction with the higher temporal-context
score
\begin{equation}
  s = n_{\text{back}} + 0.25\, n_{\text{fwd}},
  \label{eq:frame-score}
\end{equation}
where $n_{\text{back}}$ and $n_{\text{fwd}}$ count preceding and following
frames within that window, respectively. This favours predictions that have
more causal context, matching the autoregressive nature of SVD.
To ensure consistency in overlapping regions, the diffusion noise tensors
are pre-sampled for the full $T$-frame sequence and sliced per window, so
overlapping frames receive identical noise realisations.

Second, the classifier-free guidance scale is raised from the default value
of $2.5$ to $5.0$.  With only 3--9 input views, the Gaussian-splatting
renderings may contain artifacts; a higher guidance scale steers the refiner
closer to these renderings, reducing hallucination in unsupported regions.

\subsection{Long-LRM.}
Long-LRM predicts a full set of 3D Gaussians in a single forward pass
through a transformer/Mamba-2 backbone that was pre-trained on sequences of
$M\!=\!32$ views.

The main adaptation for sparse inputs is cyclic view repetition.
Given $N$ sparse input views, we construct a length-$M$ input sequence
by tiling:
\begin{equation}
  \tilde{\mathbf{I}}_j = \mathbf{I}_{j \bmod N},
  \quad j = 0,\dots, M-1.
  \label{eq:view-repeat}
\end{equation}
The corresponding poses and intrinsics are repeated in the same order.
Because the model's self-attention aggregates information across tokens
regardless of ordering, duplicate views do not degrade prediction quality.
They instead reinforce the available observations. The predicted Gaussians are then rendered from each orbit camera in a single
rasterisation pass.

\subsection{Difix3D}

Difix3D is a post-rendering diffusion model that removes artifacts from 3D Gaussian Splatting (3DGS) outputs by conditioning on a reference photograph. Its original pipeline renders novel views along an interpolated path between the training cameras and applies the diffusion fixer per frame, selecting the nearest training view as the reference. We replace the interpolated path with the orbital trajectory described above, so the same render-then-fix pipeline now produces full \threesixty sequences from $P$ inputs without other algorithmic changes.

\subsection{MVGenMaster}

MVGenMaster is a multi-view diffusion model that conditions on reference images, their camera rays, and 3D positional encodings (3DPE) warped via metric depth maps. The original inference pipeline uses Depth Pro to obtain metric depth and supports chunked generation when the target sequence exceeds the model's context window. We adapt it to the sparse-view orbit setting with one key change.

\paragraph{Monocular depth alignment.}
The 3DPE module warps reference-view features into each target viewpoint using metric depth. In the original pipeline, Depth Pro provides metric depth directly. In our sparse-view benchmark, we instead use Depth Anything V2, which outputs disparity rather than metric depth. We align the disparity maps to world scale using the camera geometry: we set the inverse-depth scale factor as $s = r_{\text{med}} / \text{median}(1/d)$, where $r_{\text{med}}$ is the median camera-to-lookat distance. This calibration gives the 3DPE module sufficient metric accuracy to produce plausible view warps from sparse inputs.

\subsection{NVS Solver}

NVS Solver is a training-free method that steers a pretrained video diffusion model toward geometric consistency using gradient-based guidance at each denoising step. The original pipeline supports single-image and two-image conditioning via depth-based warping with mask-weighted gradient refinement. We extend it to handle $P > 2$ input views.

\paragraph{Multi-view depth fusion.}
For each target orbital pose, we forward-warp \emph{every} reference image into the target view using its depth map and known camera transformation, then fuse the reprojected pixels by averaging over valid contributions:
\begin{equation}
    \tilde{I}_t(x) = \frac{\sum_{i=1}^{P} m_{i \to t}(x)\,\Pi(I_i;\, D_i,\, T_{i \to t})(x)}{\sum_{i=1}^{P} m_{i \to t}(x) + \epsilon},
\end{equation}
where $\Pi$ denotes depth-based warping and $m_{i \to t}(x)$ indicates pixel validity. Regions with no valid reprojection are flagged as holes via a dilated binary mask $M_t$. The fused image and mask are then passed through the existing gradient-refinement mechanism, originally designed for one or two source views. This converts the sparse-view extrapolation problem into a structured inpainting task without modifying the underlying denoising procedure.

\subsection{Stable Virtual Camera (SEVA)}

SEVA conditions a video diffusion model on Pl\"ucker ray embeddings and supports multi-view conditioning, two-pass generation with anchor frames, and distance-aware classifier-free guidance out of the box. We introduce a new task mode, \texttt{img2trajvid\_m-prob}, that combines multi-view conditioning (as in SEVA's existing sparse-view task) with preset trajectory priors (as in its single-view trajectory task). Given $P$ input views, we estimate the orbital camera path from their poses and feed it as the target trajectory, while the model's existing two-pass sampling and adaptive guidance handle the long-sequence generation without modification. No changes to the model or inference algorithm were needed.

\subsection{ViewCrafter}

ViewCrafter uses a video diffusion model conditioned on point-cloud renderings and an image embedding to synthesize novel views. The original model reconstructs a point cloud from DUSt3R, renders it under target cameras as a geometric prior, and conditions the diffusion model on a single reference image via cross-attention. We add multi-reference conditioning and an orbit rendering mode for sparse-view 360° generation.

\paragraph{Multi-reference image conditioning.}
The original model encodes a reference image into an embedding, which is concatenated with the text embedding for cross-attention. We extend this to $P$ references by independently encoding each image into an embedding $e_i$ and aggregating before injection. We support averaging ($e = \frac{1}{P}\sum_i e_i$), which produces a compact summary, and concatenation ($e = [e_1;\ldots;e_P]$), which preserves per-view detail at the cost of a longer attention context. Both modes require no architectural changes and keep all observed views in context throughout generation, which is especially important for viewpoints far from any single input on the orbit.

\paragraph{Single-pass orbit generation.}
ViewCrafter renders the point cloud under the full orbital trajectory and runs a single diffusion pass over all frames, which is both more efficient and more temporally consistent than the chunked generation used by methods like SEVA. Chunking into fixed-length clips re-runs diffusion independently per chunk, introducing boundary flicker and drift where consecutive clips meet. A single pass avoids these artifacts, though it is only feasible when the trajectory fits within the model's sequence length.

\section{Reconstruction Behavior on \benchmark Benchmark}
\label{sec:recon-behavior}

To understand why learned 3D consistency metrics fail on 3D-inconsistent inputs, we inspect the reconstruction outputs produced by the backbones themselves on \benchmark. We evaluate the 4 backbones - MASt3R, DUSt3R, Fast3R, and VGGT over $K \in \{3,6,9\}$ across every scene type in \benchmark for our analysis.

\begin{table*}[t]
  \centering
  \caption{Coarse reconstruction statistics on 3D-inconsistent inputs from the controlled $K \in \{3,6,9\}$ sweep. BBox Vol, Conf, and Depth are averaged over $K \in \{3,6,9\}$; Points shows the $K{=}6$ value (scales as $\sim50\text{K} \times K$ for the DUSt3R, MASt3R and $\sim270\text{K} \times K$ for VGGT). We use this table as geometric context; the hallucination-specific overlap, residual, and ghost diagnostics are reported separately below.}
  \label{tab:recon_behavior}
  \small
  \begin{tabular}{llrrrr}
    \toprule
    Input Type & Engine & Points & BBox Vol & Conf & Depth \\
    \midrule
    \multirow{4}{*}{Mixed}
    & MASt3R & 301K & 4086 & 1.06 & 2.77 \\
    & DUSt3R & 301K & 3.7 & 1.11 & 0.47 \\
    & Fast3R & 301K & 0.5 & 1.07 & 0.20 \\
    & VGGT & 1.6M & 21 & 1.05 & 0.78 \\
    \midrule
    \multirow{4}{*}{One-Outlier}
    & MASt3R & 301K & 3975 & 2.29 & 3.20 \\
    & DUSt3R & 301K & 6.3 & 1.59 & 0.53 \\
    & Fast3R & 301K & 1.7 & 1.31 & 0.25 \\
    & VGGT & 1.6M & 109 & 1.04 & 1.20 \\
    \midrule
    \multirow{4}{*}{Gaussian Noise}
    & MASt3R & 301K & 1.3 & 1.00 & 1.54 \\
    & DUSt3R & 301K & 0.05 & 1.81 & 0.52 \\
    & Fast3R & 301K & 0.00 & 1.21 & 0.20 \\
    & VGGT & 1.6M & 0.4 & 1.00 & 0.70 \\
    \midrule
    \multirow{4}{*}{Patched}
    & MASt3R & 301K & 643 & 1.58 & 2.27 \\
    & DUSt3R & 301K & 1.0 & 1.22 & 0.52 \\
    & Fast3R & 301K & 0.4 & 1.18 & 0.23 \\
    & VGGT & 1.6M & 18 & 1.05 & 1.00 \\
    \bottomrule
  \end{tabular}
\end{table*}

\subsection{Diagnostics, definitions, and main patterns}
\label{app:recon-diagnostics}

We use four complementary diagnostics. First, \cref{tab:recon_behavior} summarizes coarse point-cloud geometry and confidence from the sweep over different scene types from \benchmark. Second, \cref{tab:overlap_gap} measures pairwise overlap hallucination via
\[
\mathrm{FNR}_{\mathrm{cross}} = P(\mathrm{overlap}\mid\mathrm{cross\mbox{-}scene}),
\]
whose ideal value is $0$. Third, \cref{tab:hallucination_residuals} splits DINOv2 cosine residuals into same-scene and cross-scene pairs after the backbone has already declared overlap. Fourth, \cref{tab:hallucination_confidence_ghost} reports confidence and, for Fast3R and VGGT, ghost mass: the fraction of rendered pixels supported by wrong-scene points. We focus on the main \benchmark scene types $L_0$--$L_3$, and Gaussian noise, while using identical images and patched Gaussian as control cases.

Across all four tools, the same picture emerges. All backbones reconstruct 3D-inconsistent inputs instead of abstaining. On cross-scene mixtures, MASt3R hallucinates the least overlap and DUSt3R the most, while Fast3R and VGGT start from comparatively high same-scene zero-overlap on $L_0$. On Gaussian noise, MASt3R, DUSt3R, and VGGT still treat essentially every pair as overlapping, so the failure begins before any feature-space aggregation. Cross-scene DINOv2 residuals are larger than same-scene residuals, but their standard deviations overlap substantially, especially for Fast3R and VGGT, which limits separability once spurious correspondences survive. Confidence is therefore only a weak failure signal: MASt3R and VGGT move toward the floor on inconsistent inputs, whereas DUSt3R remains highly confident on Gaussian noise.

\begin{figure}[t]
  \centering
  \begin{subfigure}{\linewidth}
    \centering
    \includegraphics[width=\linewidth]{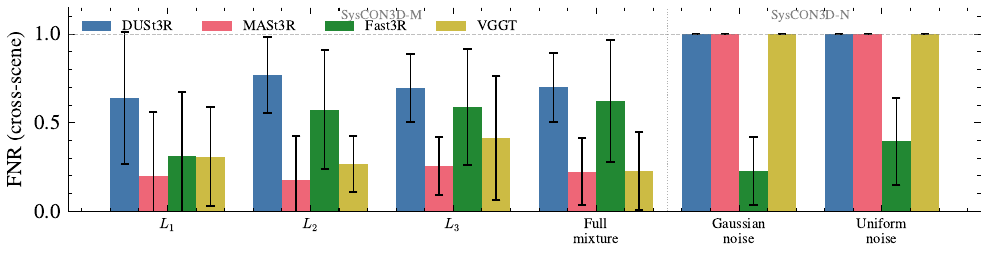}
    \caption{$\mathrm{FNR}_{\mathrm{cross}}$ on \benchmark-M and \benchmark-N.}
  \end{subfigure}
  \begin{subfigure}{\linewidth}
    \centering
    \includegraphics[width=\linewidth]{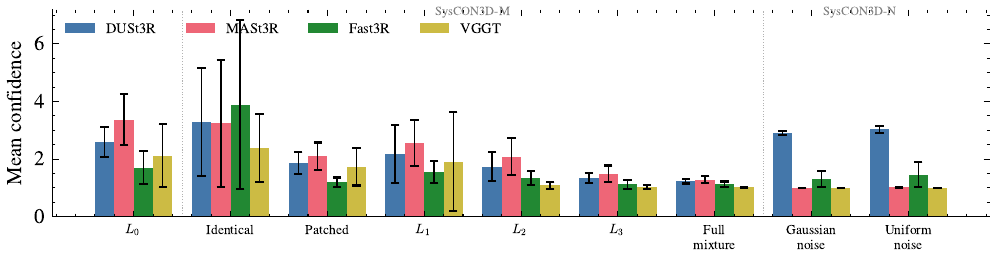}
    \caption{Mean confidence across \benchmark scene types.}
  \end{subfigure}
  \caption{\textbf{Hallucination diagnostics across the four backbones.} Left: cross-scene overlap hallucination rises sharply on $L_2$, $L_3$, and Gaussian noise. Right: confidence decreases on harder cross-scene mixtures, but it is not a reliable rejection signal on \benchmark-N, most notably for DUSt3R.}
  \label{fig:hallucination_diagnostics}
\end{figure}

\subsection{What each diagnostic reveals}
\begin{figure}[!ht]
  \centering
  \includegraphics[width=\linewidth]{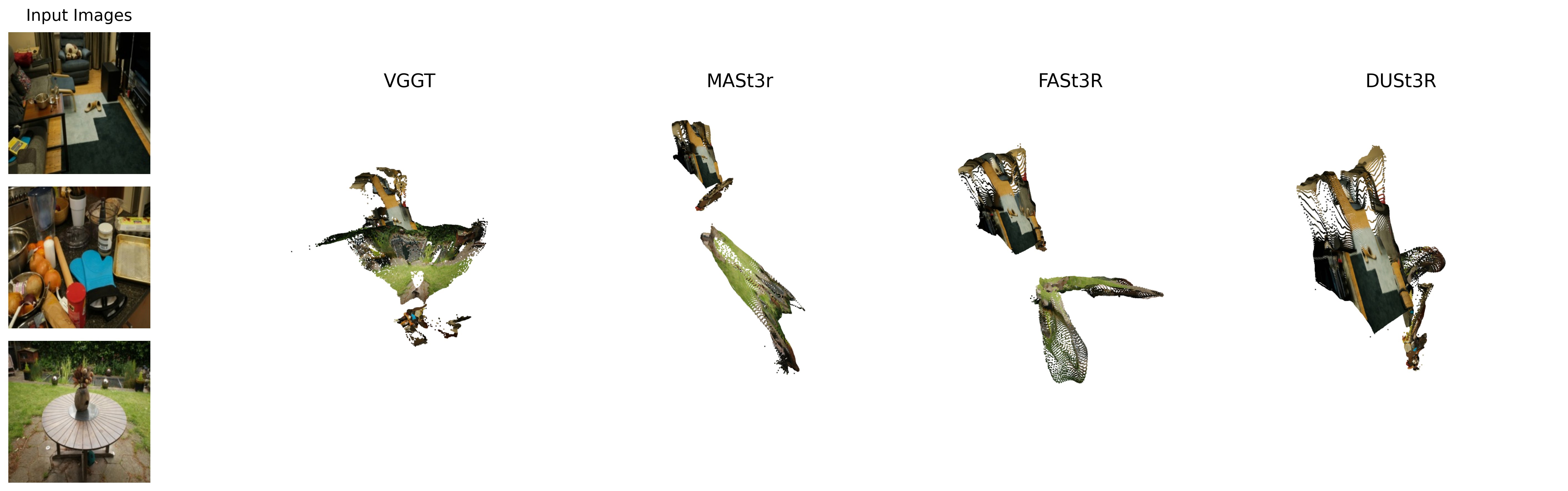}
  \caption{Analysis of 3D reconstruction backbones on SysCON3D benchmark. Input is Random mixture ($L_3$) with $K=3$.}
  \label{fig:k03_mixed_seed_1045_figure}
\end{figure}

\begin{figure}[!ht]
  \centering
  \includegraphics[width=\linewidth]{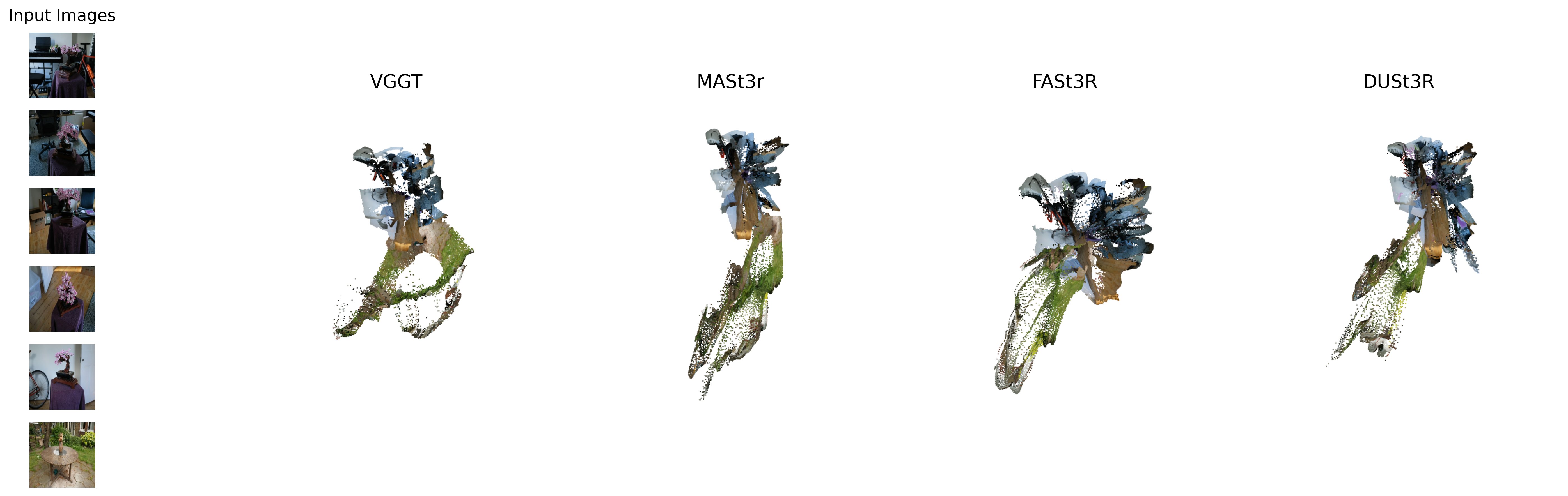}
  \caption{Analysis of 3D reconstruction backbones on SysCON3D benchmark. Input is Single-outlier ($L_1$) with $K=6$.}
  \label{fig:k06_one_outlier_base_bonsai_seed_2148_figure}
\end{figure}

\begin{figure}[!ht]
  \centering
  \includegraphics[width=\linewidth]{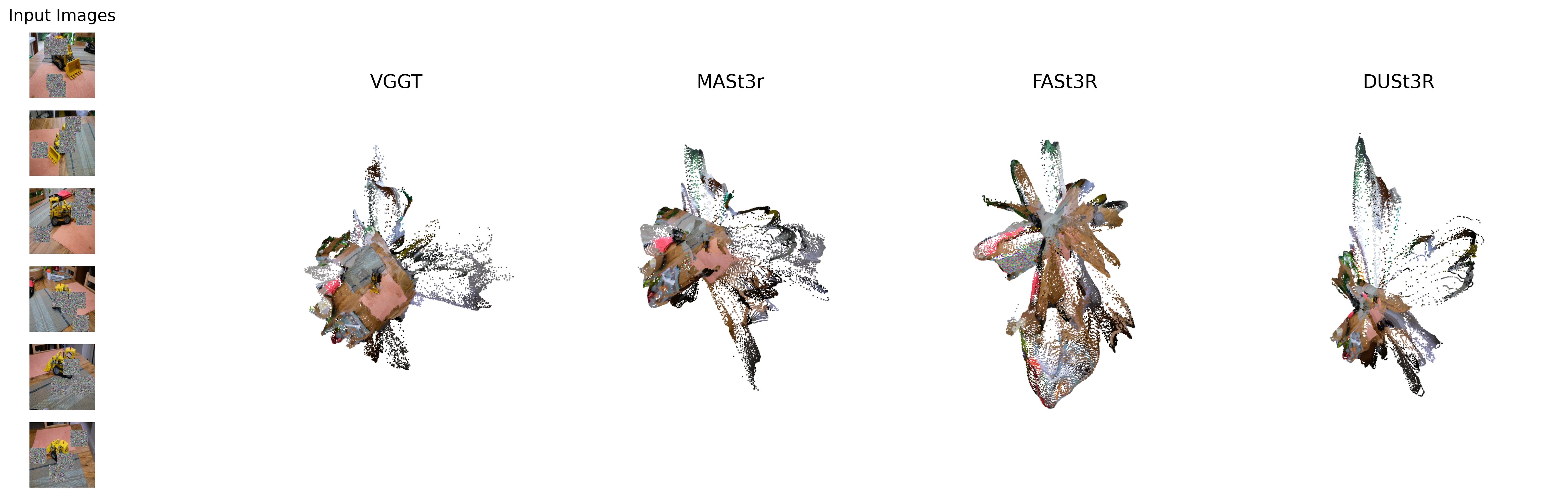}
  \caption{Analysis of 3D reconstruction backbones on SysCON3D benchmark. Input is Patched noise / Patched images with $K=6$. (Seed: 4148)}
  \label{fig:k06_patched_base_kitchen_seed_4148_figure}
\end{figure}

\begin{figure}[!ht]
  \centering
  \includegraphics[width=\linewidth]{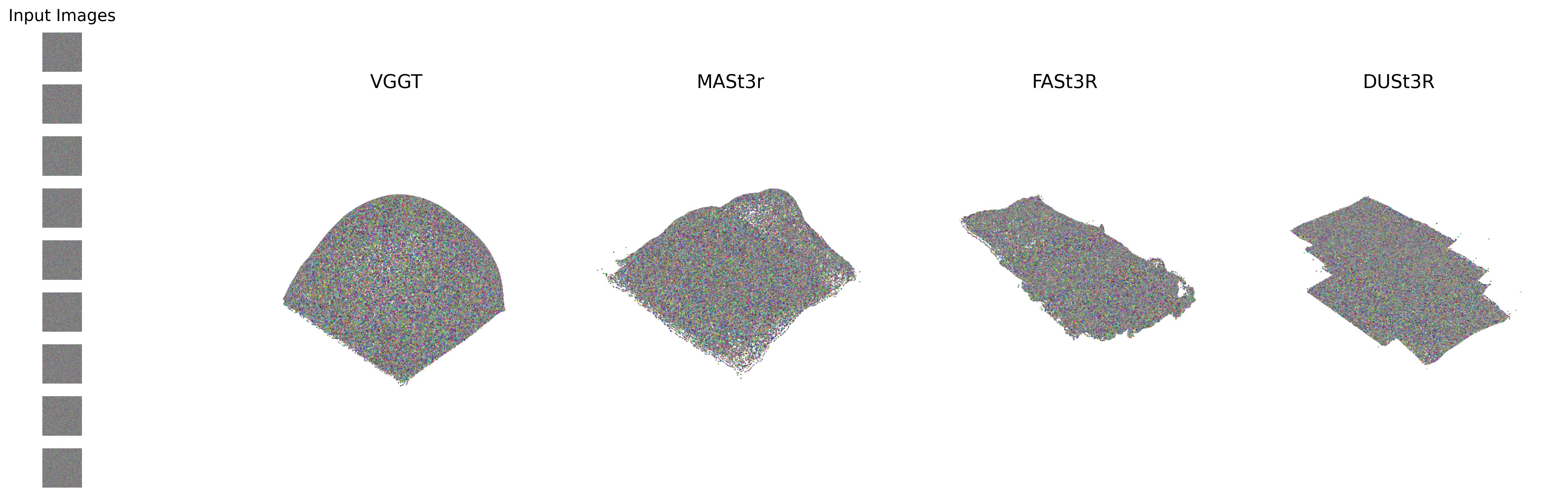}
  \caption{Analysis of 3D reconstruction backbones on SysCON3D benchmark. Input is Gaussian noise with $K=9$. (Seed: 3051)}
  \label{fig:k09_gaussian_noise_seed_3051_figure}
\end{figure}

\paragraph{Reconstruction geometry.}
\Cref{tab:recon_behavior} shows that every backbone produces dense point clouds even on inconsistent inputs. The number of points is determined mainly by image resolution, so Gaussian noise still yields 301K points for MASt3R, DUSt3R, and Fast3R, and 1.6M points for VGGT at $K{=}6$. Geometry collapses rather than disappearing: Gaussian noise shrinks the bounding-box volume to nearly zero, while mixed inputs produce large mash-up reconstructions that merge incompatible scenes into a single frame. The examples in \cref{fig:k03_mixed_seed_1045_figure,fig:k06_one_outlier_base_bonsai_seed_2148_figure,fig:k06_patched_base_kitchen_seed_4148_figure,fig:k09_gaussian_noise_seed_3051_figure} show the characteristic modes. Full-Mixed, $L_3$ and $L_2$ stitch multiple scenes together; $L_1$ keeps a dominant scene while absorbing the foreign view; patched noise corrupts local support; and Gaussian noise collapses into planes or domes.

\paragraph{Backbones hallucinate overlap on cross-scene pairs.}
\label{app:recon-overlap}
\Cref{tab:overlap_gap} and \Cref{tab:overlap_gap2} are complementary. \Cref{tab:overlap_gap} is the coarse scene-level view: it compares the expected fraction of cross-scene pairs in each level against the backbone's overall zero-overlap rate across \emph{all} pairs. If a backbone cleanly rejected foreign views while preserving same-scene matches, the two quantities would be similar. Instead, all four backbones under-detect zero-overlap on the harder mixture levels. DUSt3R is the clearest failure: on $L_3$, it reports only 26\% zero-overlap, against an expected 89\%, a $-63$pp gap. At the same level, MASt3R, Fast3R, and VGGT still under-report zero-overlap by 21pp, 48pp, and 18pp. The apparent positive deltas for Fast3R and VGGT on $L_1$ do not indicate better cross-scene rejection. They arise because these backbones already drop many same-scene pairs on $L_0$ itself, with baseline zero-overlap rates of 39\% and 21\%, respectively.

\begin{table}[t]
  \centering
  \caption{Expected cross-scene pair fraction versus the backbone's overall detected zero-overlap pair fraction, averaged across $K \in \{3,\ldots,21\}$. This is a coarse proxy because the measured zero-overlap rate mixes both same-scene and cross-scene pairs. A negative delta indicates that the backbone shows overlap where none should exist.}
  \label{tab:overlap_gap}
  \small
  \begin{tabular}{lccccc}
    \toprule
    Level & Expected & MASt3R & DUSt3R & Fast3R & VGGT \\
    \midrule
    $L_0$ (consistent) & 0\% & 4\% & 2\% & 39\% & 21\% \\
    $L_1$ (one outlier) & 25\% & 22\% (\textcolor{red}{$-3$}) & 8\% (\textcolor{red}{$-17$}) & 40\% (\textcolor{blue}{$+15$}) & 32\% (\textcolor{blue}{$+7$}) \\
    $L_2$ (controlled) & 60\% & 47\% (\textcolor{red}{$-13$}) & 15\% (\textcolor{red}{$-45$}) & 39\% (\textcolor{red}{$-22$}) & 45\% (\textcolor{red}{$-16$}) \\
    $L_3$ (random mixture) & 89\% & 68\% (\textcolor{red}{$-21$}) & 26\% (\textcolor{red}{$-63$}) & 41\% (\textcolor{red}{$-48$}) & 71\% (\textcolor{red}{$-18$}) \\
    \bottomrule
  \end{tabular}
\end{table}

\noindent
\label{app:why-corrupted-score-better}
\begin{table}[t]
\centering
\caption{\textbf{Cross-scene overlap failure rates in the hallucination sweep.} Values are mean $\pm$ std per scene type and are reported in percent. $\mathrm{FNR}_{\mathrm{cross}} = P(\mathrm{overlap}\mid\mathrm{cross\mbox{-}scene})$ is the fraction of cross-scene pairs that the backbone still treats as overlapping, so the ideal value is $0$. $\mathrm{FPR}_{L_0}$ reports the same-scene zero-overlap baseline on consistent scenes, which is already non-zero for Fast3R and VGGT on far-apart views.}
\label{tab:overlap_gap2}
\small
\begin{tabular}{lccccc}
\toprule
Backbone & $\mathrm{FPR}_{L_0}$ & $L_1$ $\mathrm{FNR}_{\mathrm{cross}}$ & $L_2$ $\mathrm{FNR}_{\mathrm{cross}}$ & $L_3$ $\mathrm{FNR}_{\mathrm{cross}}$ & Full mix $\mathrm{FNR}_{\mathrm{cross}}$ \\
\midrule
MASt3R & $0.2 \pm 0.8$ & $19.8 \pm 36.2$ & $17.9 \pm 24.6$ & $25.7 \pm 16.2$ & $22.4 \pm 18.8$ \\
DUSt3R & $0.5 \pm 1.5$ & $64.0 \pm 37.2$ & $76.9 \pm 21.4$ & $69.6 \pm 19.1$ & $69.9 \pm 19.5$ \\
Fast3R & $39.2 \pm 34.1$ & $31.2 \pm 35.9$ & $57.4 \pm 33.5$ & $59.1 \pm 32.8$ & $62.1 \pm 34.4$ \\
VGGT & $18.1 \pm 17.7$ & $30.8 \pm 27.8$ & $26.9 \pm 15.9$ & $41.3 \pm 34.9$ & $22.8 \pm 22.0$ \\
\bottomrule
\end{tabular}
\end{table}

\Cref{tab:overlap_gap2} removes that same-scene confound by conditioning only on cross-scene pairs, so $\mathrm{FNR}_{\mathrm{cross}} = P(\mathrm{overlap}\mid\mathrm{cross\mbox{-}scene})$ directly measures hallucinated correspondences. Under this cleaner view, the backbone ordering is consistent with the coarse table: MASt3R is the least permissive on cross-scene mixtures, with $\mathrm{FNR}_{\mathrm{cross}}$ of 19.8\%, 17.9\%, 25.7\%, and 22.4\% on $L_1$, $L_2$, $L_3$, while DUSt3R is the most permissive, ranging from 64.0\% on $L_1$ to 76.9\% on $L_2$. Fast3R and VGGT lie in between at $L_1$--$L_3$, but their $L_0$ baselines remain high at 39.2\% and 18.1\%, respectively, which explains why they can look less problematic in \cref{tab:overlap_gap} than they are on truly cross-scene pairs. Gaussian noise is even more extreme: MASt3R, DUSt3R, and VGGT hallucinate overlap on every pair, and Fast3R still hallucinates overlap on 22.7\% of Gaussian-noise pairs. Patched Gaussian is not a cross-scene condition and therefore does not appear in \cref{tab:overlap_gap2}; its effect is visible through the same-scene zero-overlap controls, where MASt3R and DUSt3R stay near zero but Fast3R rises to 72.4\% and VGGT to 30.4\%.

\paragraph{Residual separation after overlap is hallucinated.}
\begin{table}[t]
\centering
\caption{\textbf{Mean DINOv2 residuals on same-scene and cross-scene pairs.} Values are mean $\pm$ std of per-sample residual means. We report $L_1$, $L_2$, and $L_3$ because they each contain both same-scene and cross-scene pairs; full mixture is omitted because it has no same-scene pairs. Cross-scene residuals are consistently larger, but the gap remains modest relative to the variance, especially for Fast3R and VGGT.}
\label{tab:hallucination_residuals}
\footnotesize
\resizebox{\columnwidth}{!}{%
\begin{tabular}{lcccccc}
\toprule
Backbone & $L_1$ same & $L_1$ cross & $L_2$ same & $L_2$ cross & $L_3$ same & $L_3$ cross \\
\midrule
MASt3R & $0.345 \pm 0.055$ & $0.780 \pm 0.144$ & $0.321 \pm 0.083$ & $0.741 \pm 0.047$ & $0.339 \pm 0.136$ & $0.767 \pm 0.088$ \\
DUSt3R & $0.388 \pm 0.103$ & $0.815 \pm 0.145$ & $0.332 \pm 0.092$ & $0.796 \pm 0.093$ & $0.352 \pm 0.154$ & $0.797 \pm 0.066$ \\
Fast3R & $0.643 \pm 0.136$ & $0.928 \pm 0.099$ & $0.597 \pm 0.155$ & $0.842 \pm 0.080$ & $0.635 \pm 0.193$ & $0.869 \pm 0.062$ \\
VGGT & $0.652 \pm 0.117$ & $0.909 \pm 0.076$ & $0.621 \pm 0.112$ & $0.845 \pm 0.098$ & $0.646 \pm 0.165$ & $0.824 \pm 0.096$ \\
\bottomrule
\end{tabular}
}
\end{table}

Once a backbone decides that two views overlap, same-scene and cross-scene residuals are not cleanly separable. Cross-scene means are larger for every backbone and level, but the gaps remain modest relative to the standard deviations. MASt3R and DUSt3R show the clearest separation, with cross-scene residuals roughly 0.42--0.46 above same-scene residuals across $L_1$--$L_3$. Fast3R and VGGT are less separated because their same-scene residuals are already high, around 0.60--0.65, while cross-scene residuals stay below 0.93. This overlap is precisely why aggregation helps on $L_2$ and $L_3$ but still struggles on $L_1$: many cross-scene pairs survive with DINOv2 dissimilarities that remain inside the same broad range as genuine pairs.

\paragraph{Confidence and ghost mass.}
\begin{table}[t]
\centering
\caption{\textbf{Confidence and ghost-mass diagnostics.} Confidence is the per-sample mean confidence reported by each backbone, shown as mean $\pm$ std. We include identical-image and patched-Gaussian controls, plus representative hard cases ($L_3$, full mixture, and Gaussian noise). Ghost mass is the fraction of rendered pixels sourced from wrong-scene points for the global backbones. Confidence often moves in the right direction, but even near the floor the models still produce structured reconstructions and low residuals.}
\label{tab:hallucination_confidence_ghost}
\resizebox{\linewidth}{!}{%
\begin{tabular}{lcccccccc}
\toprule
Backbone & $L_0$ conf. & Identical conf. & Patched conf. & $L_3$ conf. & Gaussian conf. & $L_3$ ghost & Full mix ghost & Gaussian ghost \\
\midrule
MASt3R & $3.38 \pm 0.89$ & $3.25 \pm 2.20$ & $2.10 \pm 0.48$ & $1.49 \pm 0.28$ & $1.01 \pm 0.01$ & -- & -- & -- \\
DUSt3R & $2.61 \pm 0.52$ & $3.29 \pm 1.87$ & $1.87 \pm 0.37$ & $1.34 \pm 0.18$ & $2.91 \pm 0.06$ & -- & -- & -- \\
Fast3R & $1.71 \pm 0.56$ & $3.90 \pm 2.95$ & $1.20 \pm 0.16$ & $1.12 \pm 0.15$ & $1.30 \pm 0.28$ & $0.39 \pm 0.27$ & $0.51 \pm 0.26$ & $0.37 \pm 0.17$ \\
VGGT & $2.12 \pm 1.09$ & $2.39 \pm 1.17$ & $1.73 \pm 0.65$ & $1.05 \pm 0.07$ & $1.00 \pm 0.00$ & $0.41 \pm 0.17$ & $0.40 \pm 0.25$ & $0.80 \pm 0.10$ \\
\bottomrule
\end{tabular}%
}
\end{table}

\Cref{fig:hallucination_diagnostics} and \cref{tab:hallucination_confidence_ghost} show that confidence partially tracks degradation on cross-scene mixtures, but it is not a reliable rejection signal. All backbones assign high confidence to the identical-image control and lower confidence by $L_3$. However, DUSt3R remains highly confident on Gaussian noise ($2.91 \pm 0.06$), while MASt3R and VGGT sit essentially at their minimum. For the global backbones, ghost mass exposes another failure that confidence alone misses: on $L_3$, roughly 40--50\% of rendered pixels are supported by wrong-scene points, and VGGT's ghost mass rises to $0.80 \pm 0.10$ on Gaussian noise.

\subsection{Implications for robust evaluation}
\label{app:recon-implications}
Together, these diagnostics show that the learned backbones hallucinate 3D consistency rather than measuring it. For our neural metric family (backbone, aggregation, residual) in ~\Cref{sec:neural-family}, the primary failure happens before aggregation: the reconstruction module $B$ invents geometry or overlap on 3D-inconsistent inputs, and the residual module $\rho$ inherits either semantically plausible DINOv2 matches on cross-scene pairs or feature collapse on noise. No choice of aggregation $\mathcal{A}$ can fully recover the correct ordering once the residual set is already degenerate. This is why distributional aggregation improves $L_2$ and $L_3$, but still fails on single outliers, on patched noise for some global backbones, and on Gaussian noise for nearly all learned variants.

This failure mode directly motivates our shift toward classical evaluators. SIFT and COLMAP can reject matches, fail to register views, or return only sparse geometry when the input set is inconsistent. That abstention signal is precisely what the learned backbones lack on \benchmark. Our COLMAP-based metrics in ~\Cref{sec:colmap-metrics} exploit this behavior by turning match verification, registration rate, and dense depth agreement into the evaluation signal, yielding a more robust and reliable notion of 3D consistency for NVS evaluation.
\begin{figure}[!ht]
  \centering
  \includegraphics[width=\linewidth]{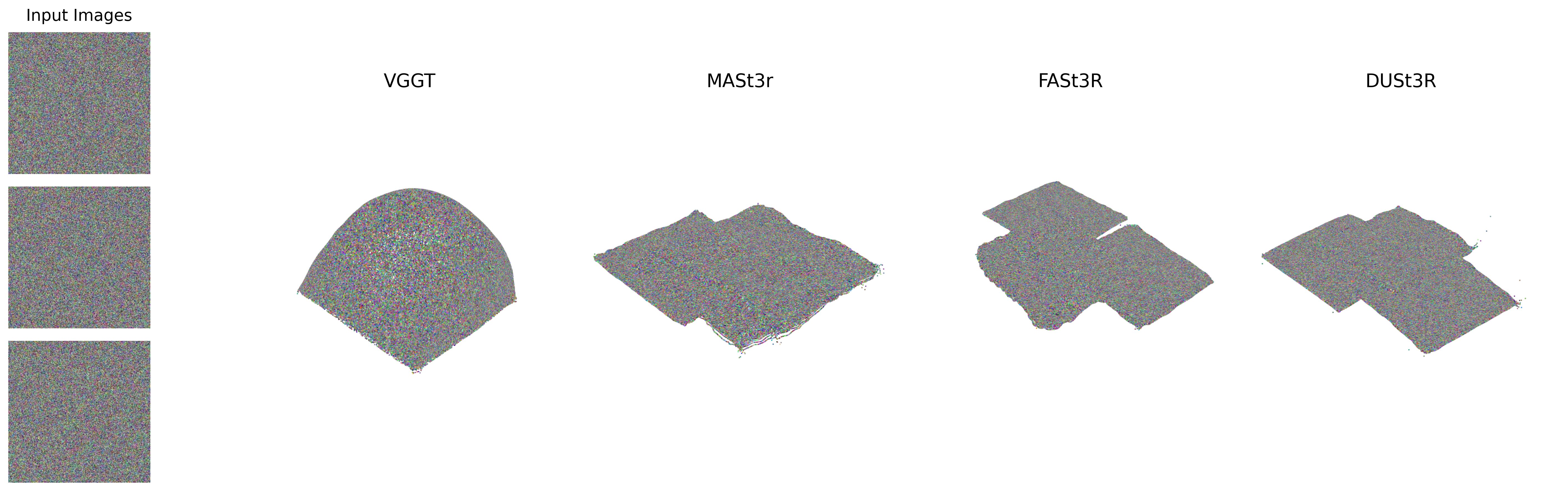}
  \caption{Analysis of 3D reconstruction backbones on SysCON3D benchmark. Input is Gaussian noise with $K=3$. (Seed: 3045)}
  \label{fig:k03_gaussian_noise_seed_3045_figure}
\end{figure}

\begin{figure}[!ht]
  \centering
  \includegraphics[width=\linewidth]{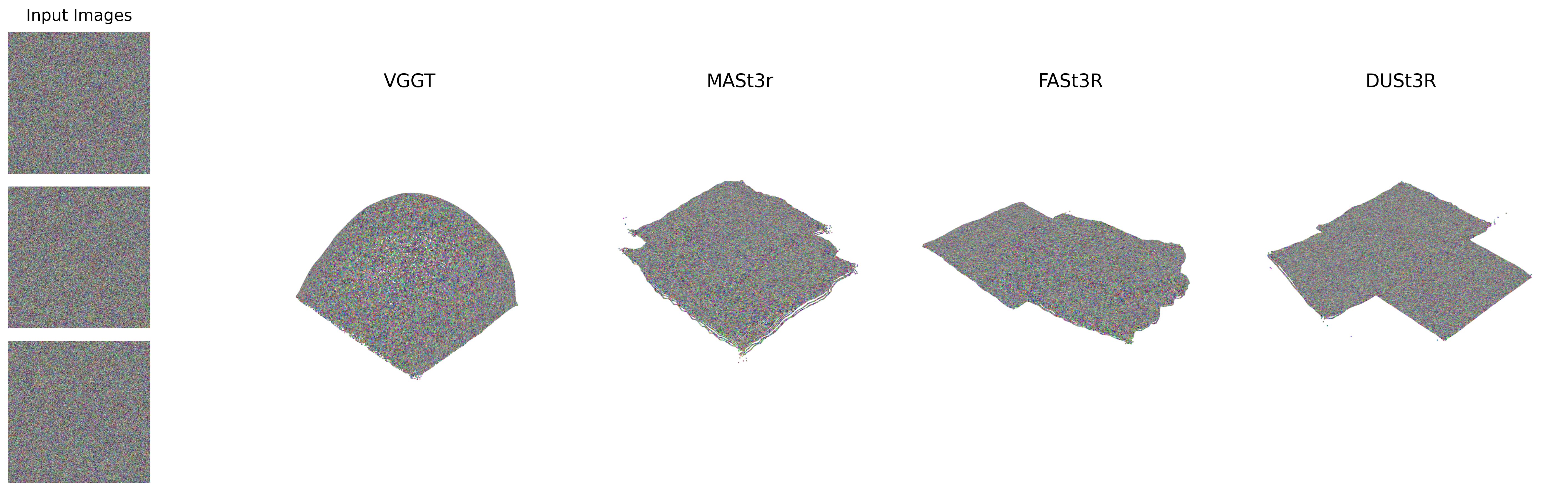}
  \caption{Analysis of 3D reconstruction backbones on SysCON3D benchmark. Input is Gaussian noise with $K=3$. (Seed: 3145)}
  \label{fig:k03_gaussian_noise_seed_3145_figure}
\end{figure}

\begin{figure}[!ht]
  \centering
  \includegraphics[width=\linewidth]{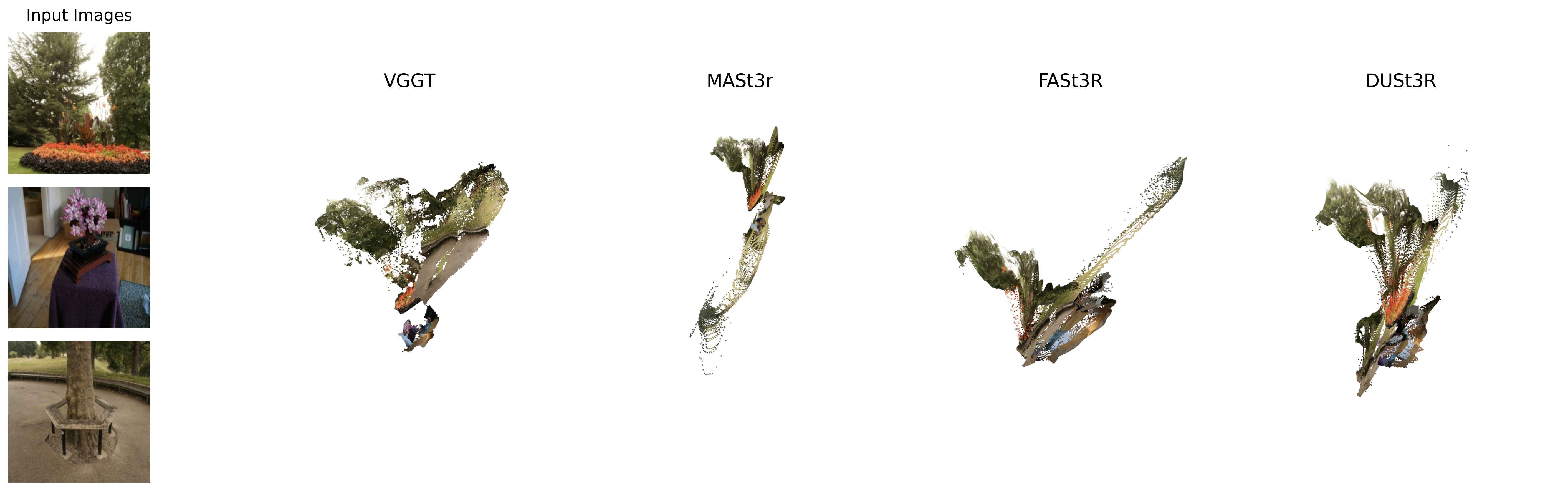}
  \caption{Analysis of 3D reconstruction backbones on SysCON3D benchmark. Input is Random mixture ($L_3$) with $K=3$.}
  \label{fig:k03_mixed_seed_1146_figure}
\end{figure}

\begin{figure}[!ht]
  \centering
  \includegraphics[width=\linewidth]{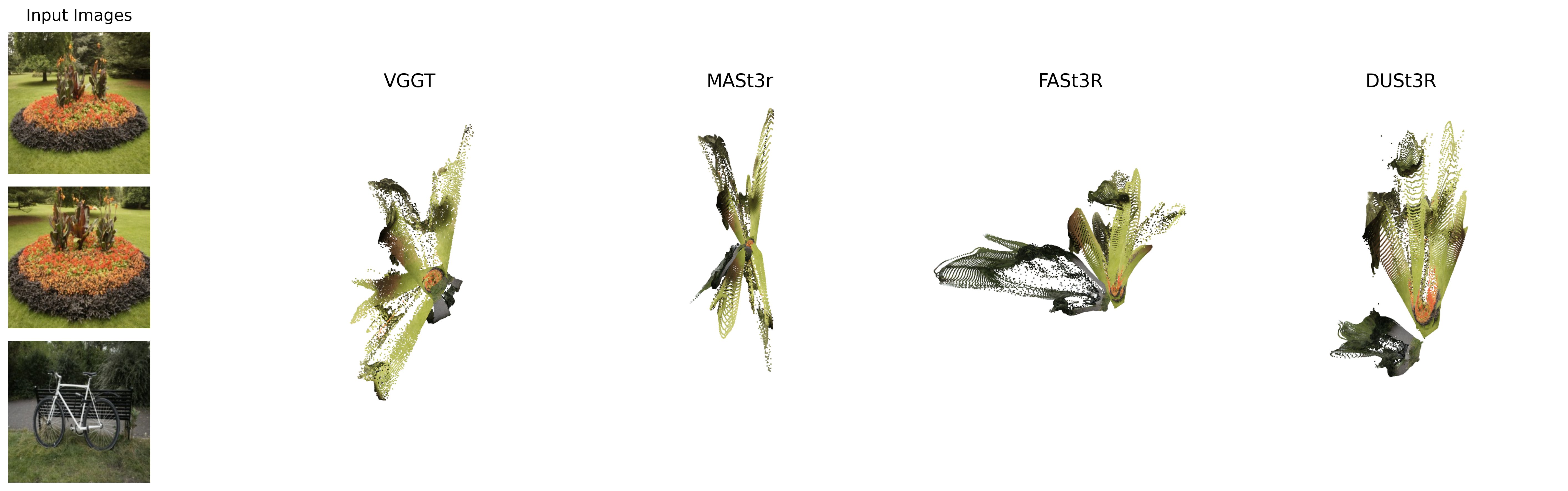}
  \caption{Analysis of 3D reconstruction backbones on SysCON3D benchmark. Input is Single-outlier ($L_1$) with $K=3$.}
  \label{fig:k03_one_outlier_base_flowers_seed_2045_figure}
\end{figure}

\begin{figure}[!ht]
  \centering
  \includegraphics[width=\linewidth]{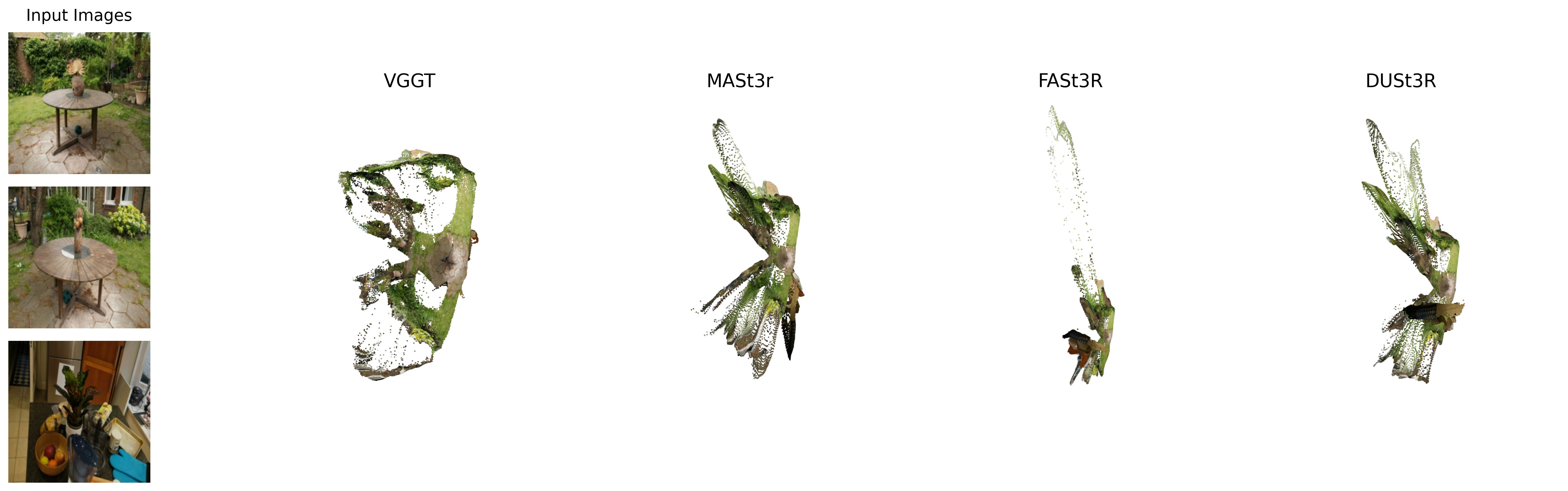}
  \caption{Analysis of 3D reconstruction backbones on SysCON3D benchmark. Input is Single-outlier ($L_1$) with $K=3$.}
  \label{fig:k03_one_outlier_base_garden_seed_2145_figure}
\end{figure}

\begin{figure}[!ht]
  \centering
  \includegraphics[width=\linewidth]{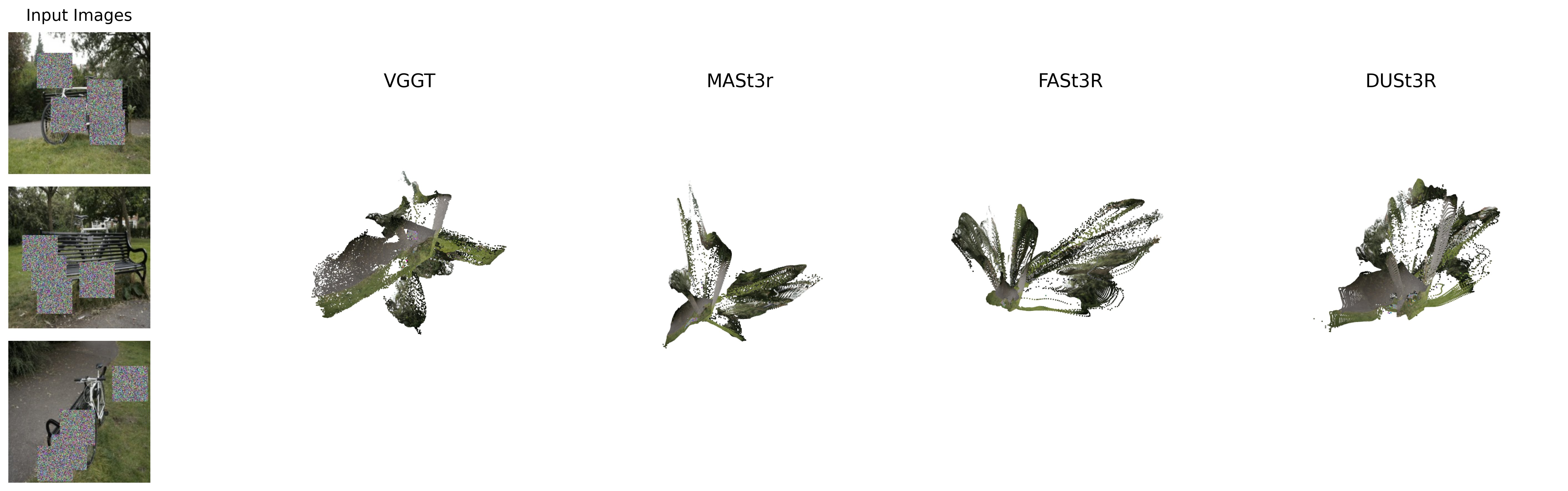}
  \caption{Analysis of 3D reconstruction backbones on SysCON3D benchmark. Input is Patched noise / Patched images with $K=3$. (Seed: 4045)}
  \label{fig:k03_patched_base_bicycle_seed_4045_figure}
\end{figure}

\begin{figure}[!ht]
  \centering
  \includegraphics[width=\linewidth]{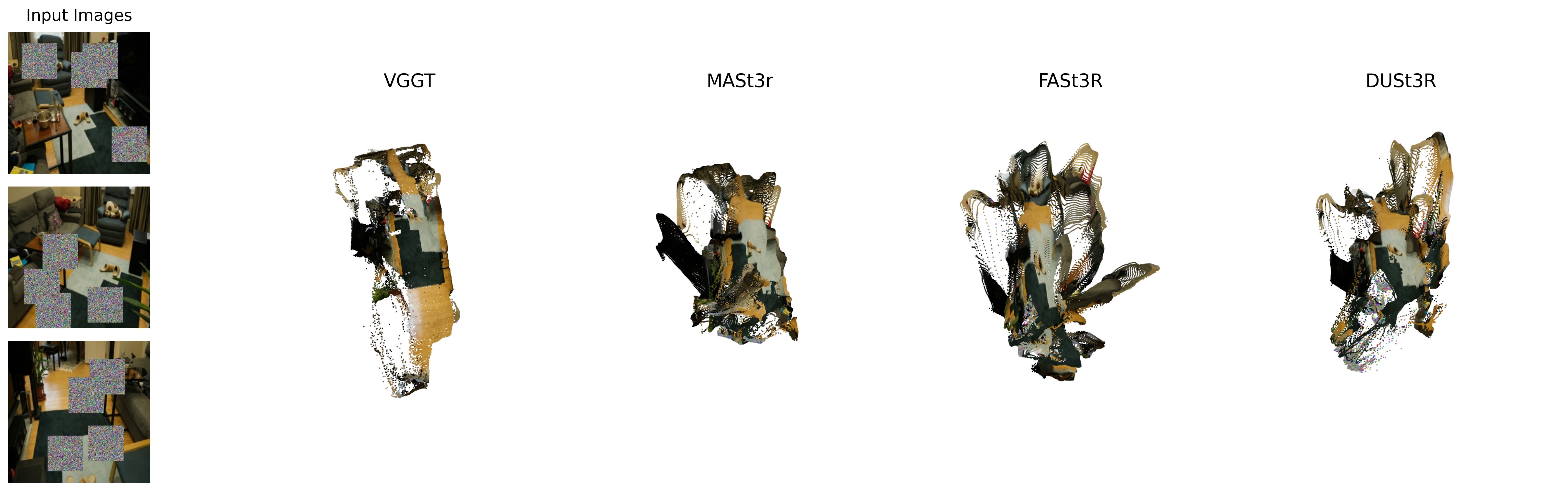}
  \caption{Analysis of 3D reconstruction backbones on SysCON3D benchmark. Input is Patched noise / Patched images with $K=3$. (Seed: 4145)}
  \label{fig:k03_patched_base_room_seed_4145_figure}
\end{figure}

\begin{figure}[!ht]
  \centering
  \includegraphics[width=\linewidth]{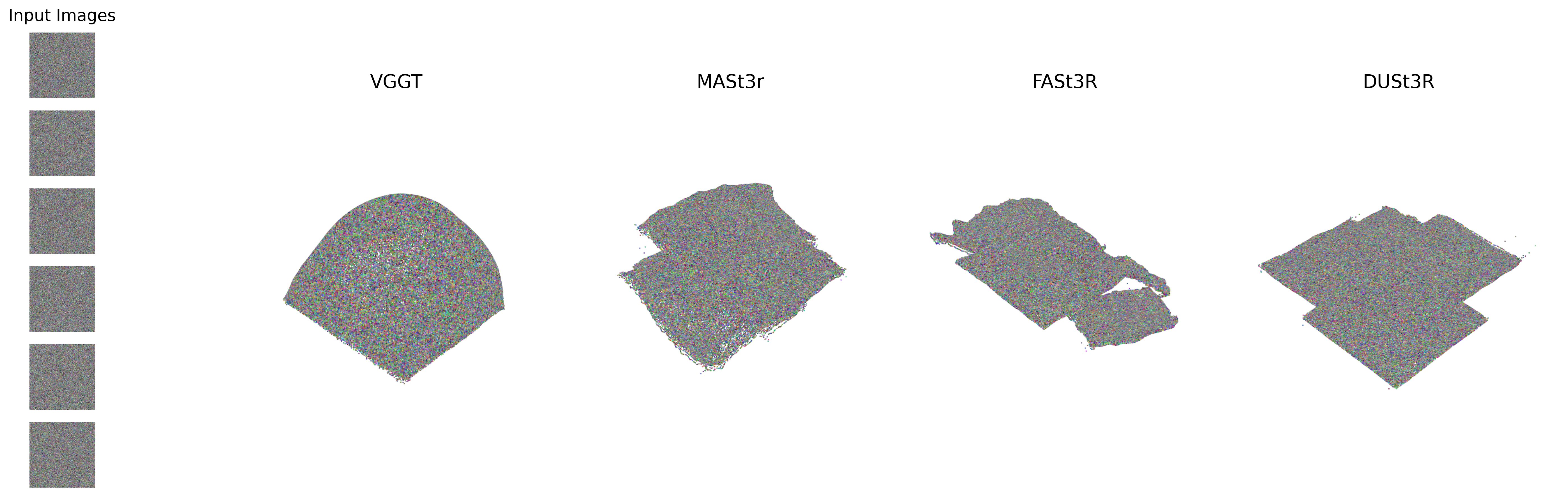}
  \caption{Analysis of 3D reconstruction backbones on SysCON3D benchmark. Input is Gaussian noise with $K=6$. (Seed: 3048)}
  \label{fig:k06_gaussian_noise_seed_3048_figure}
\end{figure}

\begin{figure}[!ht]
  \centering
  \includegraphics[width=\linewidth]{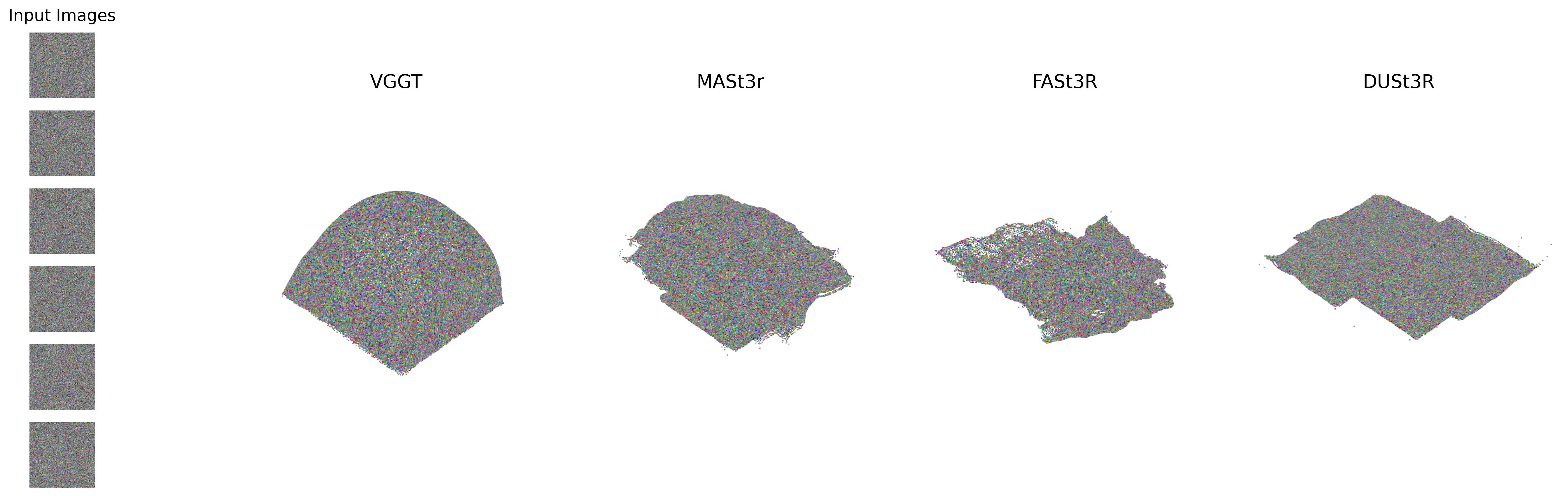}
  \caption{Analysis of 3D reconstruction backbones on SysCON3D benchmark. Input is Gaussian noise with $K=6$. (Seed: 3148)}
  \label{fig:k06_gaussian_noise_seed_3148_figure}
\end{figure}

\begin{figure}[!ht]
  \centering
  \includegraphics[width=\linewidth]{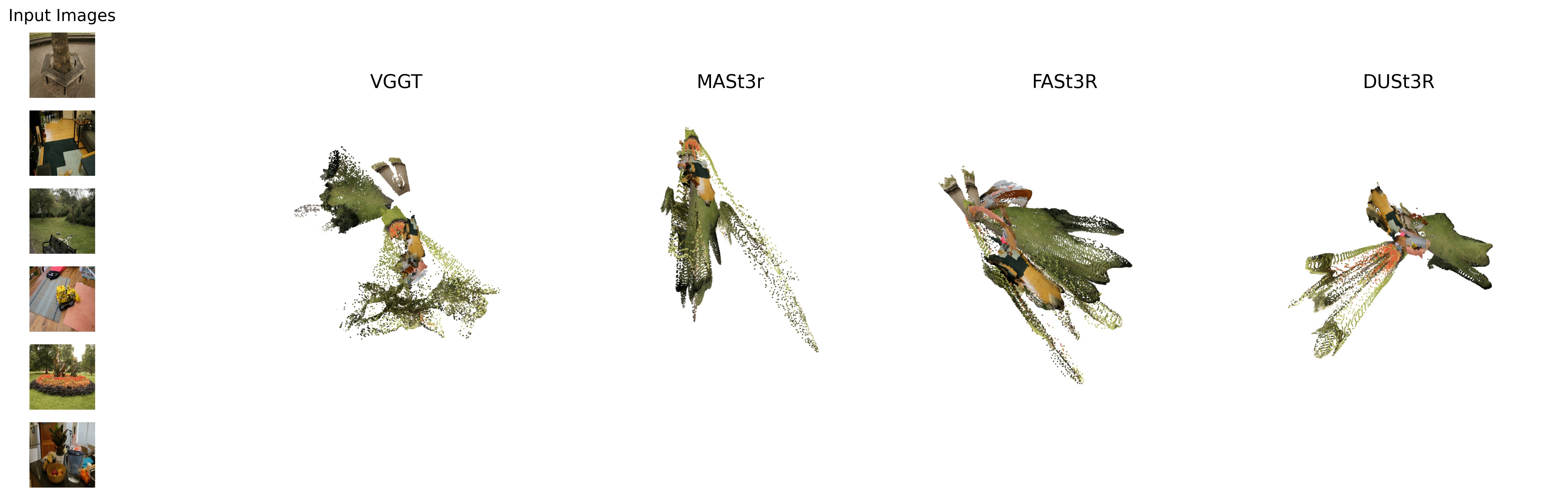}
  \caption{Analysis of 3D reconstruction backbones on SysCON3D benchmark. Input is Random mixture ($L_3$) with $K=6$.}
  \label{fig:k06_mixed_seed_1048_figure}
\end{figure}

\begin{figure}[!ht]
  \centering
  \includegraphics[width=\linewidth]{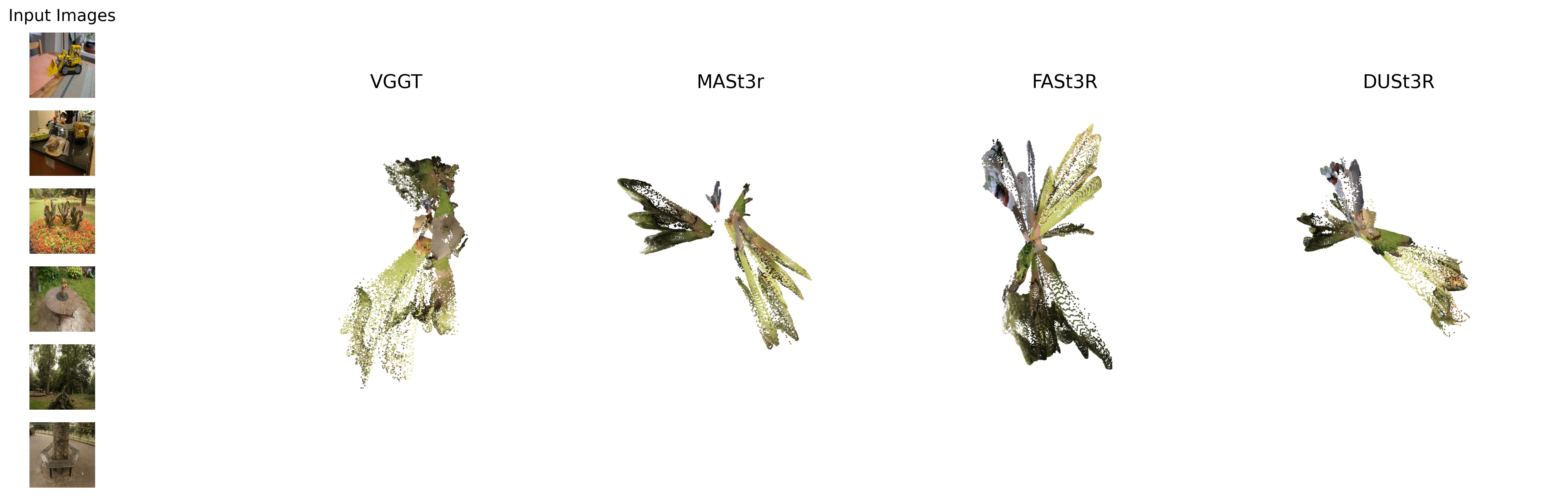}
  \caption{Analysis of 3D reconstruction backbones on SysCON3D benchmark. Input is Random mixture ($L_3$) with $K=6$.}
  \label{fig:k06_mixed_seed_1149_figure}
\end{figure}

\begin{figure}[!ht]
  \centering
  \includegraphics[width=\linewidth]{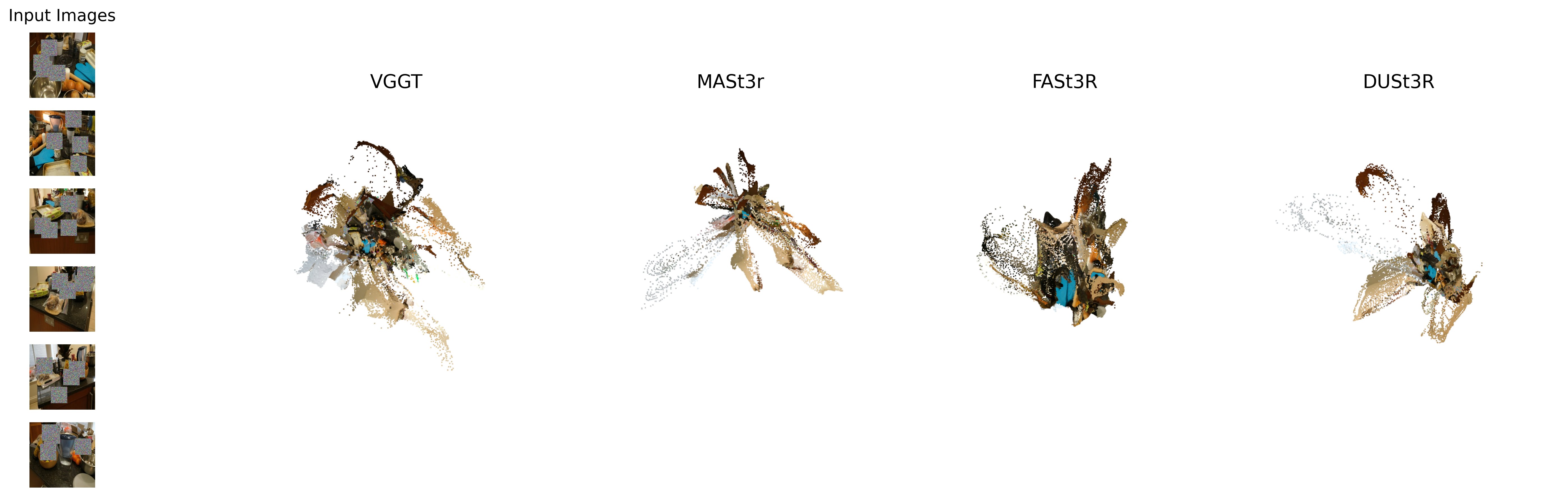}
  \caption{Analysis of 3D reconstruction backbones on SysCON3D benchmark. Input is Patched noise / Patched images with $K=6$. (Seed: 4048)}
  \label{fig:k06_patched_base_counter_seed_4048_figure}
\end{figure}

\begin{figure}[!ht]
  \centering
  \includegraphics[width=\linewidth]{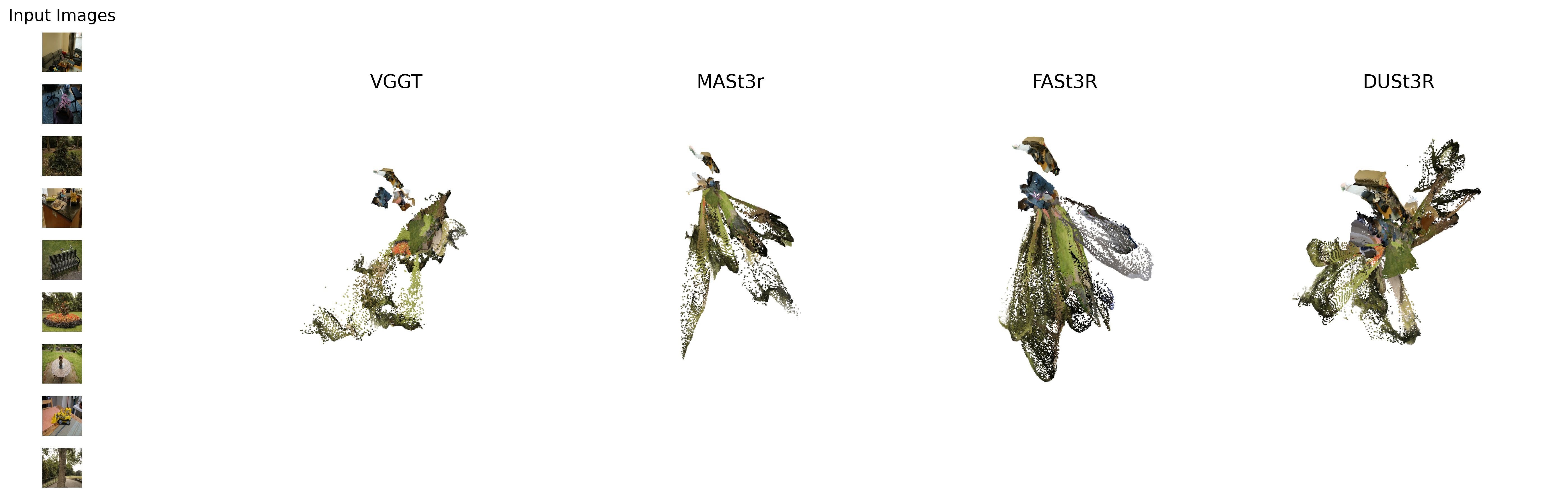}
  \caption{Analysis of 3D reconstruction backbones on SysCON3D benchmark. Input is Random mixture ($L_3$) with $K=9$.}
  \label{fig:k09_mixed_seed_1051_figure}
\end{figure}

\FloatBarrier

\subsection{Signature Hallucination Patterns}
\label{app:hallu_pattern}

Our hallucination analysis reveals that each learned 3D reconstruction model tends to produce a characteristic fallback geometry under severely inconsistent inputs. Although related artifacts can sometimes appear for real image sets with poor geometric correspondence, they are most clearly exposed by Gaussian-noise inputs. In that setting, the reconstructed geometry is remarkably stable across different noise realizations and numbers of views, indicating that it is dominated by the model prior rather than image evidence. VGGT, for instance, consistently produces a canopy-like structure, while MASt3R, DUSt3R, and Fast3R tend to reconstruct variants of a small set of approximately planar surfaces with model-specific spacing. This suggests that hallucinated geometry may exhibit backbone-specific signatures determined by architecture and training, which may be useful to study in future work.

\subsection{Hallucinated correspondences on real NVS output}
\label{app:nvs-correspondences}

\begin{figure}[!htbp]
  \centering
  \begin{subfigure}{\columnwidth}
    \centering
    \includegraphics[width=0.8\linewidth]{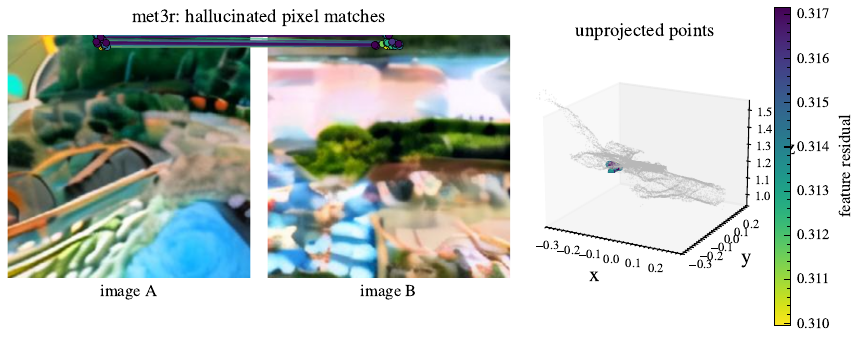}
    \caption{Scene 1, MASt3R correspondences.}
  \end{subfigure}
  \begin{subfigure}{\columnwidth}
    \centering
    \includegraphics[width=0.8\linewidth]{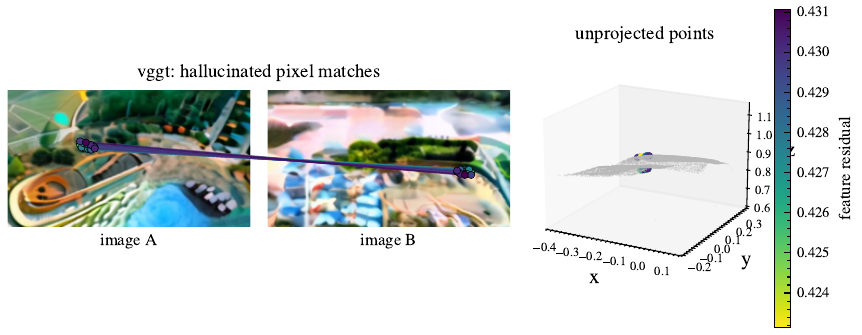}
    \caption{Scene 1, VGGT correspondences.}
  \end{subfigure}
  \begin{subfigure}{\columnwidth}
    \centering
    \includegraphics[width=0.8\linewidth]{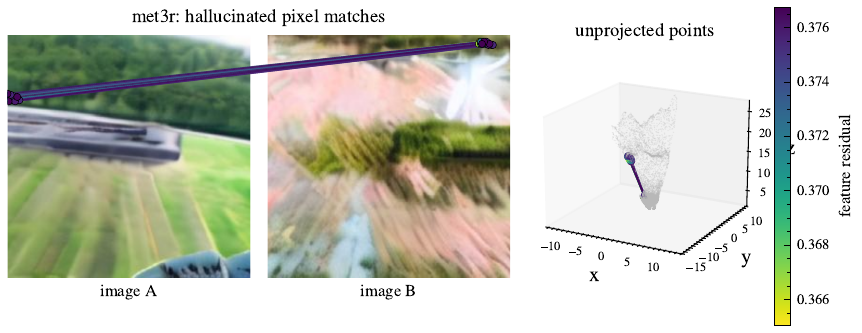}
    \caption{Scene 2, MASt3R correspondences.}
  \end{subfigure}
  \begin{subfigure}{\columnwidth}
    \centering
    \includegraphics[width=0.8\linewidth]{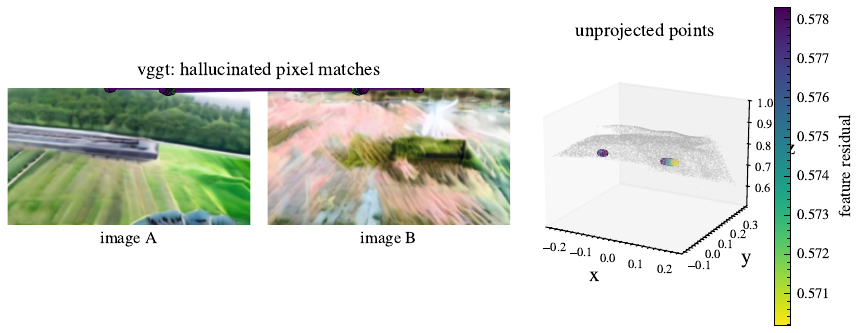}
    \caption{Scene 2, VGGT correspondences.}
  \end{subfigure}
  \caption{\textbf{Hallucinated correspondences between unrelated NVS frames.} Pixel correspondences returned by MASt3R (top of each pair) and VGGT (bottom) on two pairs of views generated by MVSplat360. The two views in each pair come from consecutive orbit cameras, and their generated content shows no perceptible geometric or semantic overlap, yet both backbones return many low-residual matches across unrelated patches. Unprojecting these spurious matches merges the two views into a shared point cloud, thereby biasing downstream consistency scores.}
  \label{fig:nvs-correspondences}
\end{figure}

The diagnostics above are measured on \benchmark, where we control the inputs directly. The same hallucination mode also appears in real NVS outputs. \cref{fig:nvs-correspondences} shows pixel correspondences returned by the two backbones on two pairs of views generated by MVSplat360~\cite{chen2024mvsplat360}, which, in our experiments, drift and hallucinate more aggressively than the other NVS methods we tested. Within each pair, we deliberately pick frames from consecutive orbit cameras whose generated content shares no perceptible geometric or semantic overlap -- a case where any backbone should return very few matches. Both MASt3R and VGGT instead link many low-residual (DINO/FeatUp feature distance small) pixel pairs across patches that clearly do not depict the same surface. When the backbone unprojects these matches, the two unrelated views are merged into a single shared point cloud, which is the same mechanism behind the metric failures in \cref{tab:metric_calibration_effect}: the residual aggregation cannot tell hallucinated matches apart from genuine ones, so the final score reports near-perfect 3D consistency on inputs that share none.

\section{Detailed Robustness Analysis}
\label{app:robustness-details}

\begin{table}[!htbp]
\centering
\caption{\textbf{Robustness on the identical-image degenerate diagnostic.} Cohen's $d$ (averaged over $K \in \{3,\ldots,21\}$) and win rate (fraction of $K$ values with $d > 0$) compare consistent sets ($L_0$) against repeated identical views. Positive $d$ means the metric correctly assigns higher (worse) scores to repeated copies than to genuine consistent sets. Most metrics fail this test: identical views yield a trivially perfect but degenerate reconstruction, so they receive lower (better) scores than genuine consistent sets (negative $d$); MMD aggregations fail most severely (e.g., VGGT PC-MMD $d{=}{-}19.72$). IMQ variants are the exception, with Fast3R W-IMQ ($d{=}{+}9.88$) and VGGT W-IMQ ($d{=}{+}11.56$) reliably separating the two. MEt3R fails ($d{<}0$, 0\% win rate); among external baselines, PRISM-MMD detects the diagnostic ($d{=}{+}2.92$). The top-3 values per column are shaded.}
\label{tab:metric_calibration_effect_identical_vs_distinct}
\begin{tabular}{llcc}
\toprule
\multirow{2}{*}{Backbone} & \multirow{2}{*}{Variant} & \multicolumn{2}{c}{Identical Images} \\
\cmidrule(lr){3-4}
 & & Avg $d$ & Win\% ($d$) \\
\midrule
MASt3R & W-Energy & -4.19 & 0\% \\
MASt3R & W-MMD & -8.89 & 0\% \\
MASt3R & W-IMQ & -2.78 & 0\% \\
Fast3R & W-Base & -0.53 & 14\% \\
Fast3R & W-Energy & -0.49 & 14\% \\
Fast3R & W-MMD & -0.75 & 29\% \\
Fast3R & W-IMQ & \cellcolor{tabsecond}+9.88 & \cellcolor{tabfirst}100\% \\
Fast3R & PC-Base & -4.28 & 0\% \\
Fast3R & PC-Energy & -3.72 & 0\% \\
Fast3R & PC-MMD & -7.92 & 0\% \\
Fast3R & PC-IMQ & -2.64 & 0\% \\
VGGT & W-Base & -0.85 & 14\% \\
VGGT & W-Robust & -0.86 & 14\% \\
VGGT & W-Energy & -1.10 & 0\% \\
VGGT & W-MMD & -1.84 & 0\% \\
VGGT & W-IMQ & \cellcolor{tabfirst}+11.56 & \cellcolor{tabsecond}100\% \\
VGGT & PC-Base & -7.25 & 0\% \\
VGGT & PC-Energy & -6.31 & 0\% \\
VGGT & PC-MMD & -19.72 & 0\% \\
VGGT & PC-IMQ & -4.38 & 0\% \\
\hdashline
\multicolumn{2}{l}{MEt3R} & -5.06 & 0\% \\
\multicolumn{2}{l}{PRISM-MMD} & \cellcolor{tabthird}+2.92 & \cellcolor{tabthird}100\% \\
\multicolumn{2}{l}{SED} & -1.17 & 0\% \\
\multicolumn{2}{l}{TSED} & -1.39 & 0\% \\
\bottomrule
\end{tabular}
\end{table}

This subsection introduces a metric based on the RobustVGGT method~\cite{han2025emergent}, and then explains some anomalous entries in \cref{tab:metric_calibration_effect,tab:metric_calibration_effect_identical_vs_distinct}.

\paragraph{RobustVGGT metric.}
\label{app:vggt-robust}
RobustVGGT~\cite{han2025emergent} is a test-time reconstruction method for mixed-scene VGGT inputs: it probes late global-attention features, scores each view by how well it agrees with an anchor view, rejects likely distractors, and reruns reconstruction on the retained subset. We adapt this into a scalar metric, \emph{VGGT/W-Robust}, as follows. We hook layer 23 of the VGGT aggregator and compute one score per view from two late signals: the mean attention mass received from the anchor-view patches, and the late-feature cosine similarity to the anchor. We min-max normalize the two terms, average them with equal weights, and keep views whose combined score exceeds 0.4 while always retaining the anchor. We then rerun the standard weighted VGGT metric on the retained subset and add an explicit penalty proportional to the rejected-view fraction, so the metric cannot improve by simply discarding evidence and ignoring the rejected views.

\paragraph{RobustVGGT behavior.}
The VGGT/W-Robust row in \cref{tab:metric_calibration_effect} shows that the added rejection signal helps on cross-scene mixtures (100\% win on $L_2$ and $L_3$) but fails on patched Gaussian and Gaussian noise (0\% win each), for an overall win rate of only 51\%. The same pattern holds in \cref{tab:metric_calibration_effect_identical_vs_distinct}: VGGT/W-Robust still ranks repeated identical views below true consistent sets ($d{=}{-}0.86$, 14\% win). Late-layer outlier rejection, therefore, helps when the failure is driven by real-world distractor views, but not when the backbone collapses on degenerate inputs.

\paragraph{VGGT PC-MMD on Gaussian noise.}
This paragraph explains the Gaussian-noise outlier score in \cref{tab:metric_calibration_effect}. VGGT PC-MMD is the only data-driven variant that achieves a positive Cohen's $d$ ($+1.31$, 100\% win rate) on Gaussian noise. This is an artifact of RBF kernel saturation rather than genuine sensitivity to noise. On noise inputs with zero visual overlap, PC residuals are uniformly high ($\sim\!1.0$), and the RBF kernel $k(e,e') = \exp(-(e{-}e')^2/2\sigma^2)$ with $\sigma{=}0.15$ saturates near its maximum for nearby high residuals, pushing MMD to $\sim\!1.55$. On real consistent scenes, PC residuals have intermediate values that keep MMD near $\sim\!1.45$, so noise accidentally exceeds consistent scores. This variant performs poorly on cross-scene mixtures (43\% overall), limiting its practical utility.

\paragraph{SIFT-SED on Gaussian noise.}
This paragraph explains why SIFT-SED produces an extreme Gaussian-noise effect size. SIFT-SED reports a degenerate Cohen's $d$ of $+161.74$ on Gaussian noise. SIFT detects no keypoints on structureless noise images, causing SED to return a fixed fallback value of 17.5 with zero variance. For $K \ge 12$, consistent-scene SED is near zero ($\sim\!0.1 \pm 0.07$), producing a tiny pooled standard deviation that inflates $d$ to $\sim\!300$ at individual $K$ values. Averaging across all 7 values of $K$ yields the reported 161.74. This is a measurement artifact: SED's keypoint-level evaluation is undefined on inputs without detectable features.

\paragraph{Single-outlier detection.}
This paragraph explains why $L_1$ remains difficult even after adding explicit view rejection. Detecting a single foreign view among $K{-}1$ consistent ones remains difficult for all metrics. The best win rate is still 71\% (VGGT W-Energy), while VGGT / W-Robust improves only to 57\%, and most variants achieve $|d| < 0.2$. This is expected: with only $\sim\!\frac{2}{K}$ cross-scene view pairs, the signal from one outlier is diluted by the consistent majority, particularly at larger $K$.

\paragraph{Per-module interaction effects.}
This paragraph places W-Robust relative to the rest of the data-driven family. The interaction between backbone and aggregation is still non-trivial. While global backbones outperform pairwise MASt3R at baseline weighted aggregation (46--49\% vs.\ 23\% overall), the strongest data-driven metric remains MASt3R W-IMQ at 71\%. VGGT/W-Robust's gains come almost entirely from $L_2$, $L_3$, and full-mixture cases, so it does not close the gap to MASt3R W-IMQ or to the better-calibrated external baselines.

\paragraph{Ranking with explicit outlier rejection.}
This paragraph compares W-Robust against the stronger ordinal baselines. In \cref{tab:rank_concordance}, VGGT / W-Robust reaches Kendall's $\tau{=}0.24$ and PPC${=}0.56$. This is slightly better than most VGGT variants, but it remains well below PRISM-MMD ($\tau{=}0.45$, PPC${=}0.72$) and TSED ($\tau{=}0.70$, PPC${=}0.76$). Even with explicit outlier rejection, the VGGT-based metric still does not maintain the ideal severity ordering as reliably as PRISM-MMD or TSED.

\section{Rank Concordance Analysis}
\label{app:rank-concordance}

\begin{table}[!ht]
\centering
\caption{\textbf{Rank concordance with the ideal scene-type ordering.} External baselines TSED and PRISM-MMD achieve the highest concordance, while MEt3R nearly perfectly inverts the intended ordering. Kendall's $\tau$ compares each metric's point-estimate ranking of the calibration scene types against the ideal ordering ($L_0 < L_1 < L_2 < L_3 \leq \text{Gaussian noise}$). Probabilistic Pairwise Concordance (PPC) accounts for score uncertainty: for each pair $(i,j)$ that should be separated, PPC $=\Phi\bigl(|\mu_i - \mu_j| / \sqrt{\sigma_i^2 + \sigma_j^2}\bigr)$, averaged over all pairs and all $K$ values. TSED ($\tau{=}0.70$, PPC${=}0.76$) and PRISM-MMD ($\tau{=}0.45$, PPC${=}0.72$) lead by a wide margin; among backbone-based variants, PC-MMD aggregations perform best but remain below $\tau{=}0.30$. MEt3R scores $\tau{=}{-}0.89$ (PPC${=}0.20$), nearly inverting the severity ordering - consistent with its inability to separate cross-scene corruption levels in \cref{tab:metric_calibration_effect}. Top-3 values per column are shaded.}
\label{tab:rank_concordance}
\begin{tabular}{llcc}
\toprule
Backbone & Variant & Kendall's $\tau$ & PPC \\
\midrule
MASt3R & W-Energy & 0.22 & 0.57 \\
MASt3R & W-MMD & 0.07 & 0.57 \\
MASt3R & W-IMQ & 0.26 & 0.59 \\
Fast3R & W-Base & 0.10 & 0.53 \\
Fast3R & W-Energy & 0.12 & 0.53 \\
Fast3R & W-MMD & 0.05 & 0.50 \\
Fast3R & W-IMQ & 0.20 & 0.57 \\
Fast3R & PC-Base & 0.20 & 0.56 \\
Fast3R & PC-Energy & 0.18 & 0.56 \\
Fast3R & PC-MMD & 0.28 & 0.62 \\
Fast3R & PC-IMQ & 0.26 & 0.60 \\
VGGT & W-Base & 0.10 & 0.50 \\
VGGT & W-Energy & 0.07 & 0.49 \\
VGGT & W-MMD & 0.12 & 0.50 \\
VGGT & W-IMQ & 0.20 & 0.56 \\
VGGT & PC-Base & 0.16 & 0.51 \\
VGGT & PC-Energy & 0.10 & 0.49 \\
VGGT & PC-MMD & \cellcolor{tabthird}0.30 & \cellcolor{tabthird}0.64 \\
VGGT & PC-IMQ & 0.16 & 0.54 \\
VGGT & W-Robust & 0.24 & 0.56 \\
\hdashline
\multicolumn{2}{l}{MEt3R} & -0.89 & 0.20 \\
\multicolumn{2}{l}{PRISM-MMD} & \cellcolor{tabsecond}0.45 & \cellcolor{tabsecond}0.72 \\
\multicolumn{2}{l}{SED} & -0.03 & 0.52 \\
\multicolumn{2}{l}{TSED} & \cellcolor{tabfirst}0.70 & \cellcolor{tabfirst}0.76 \\
\bottomrule
\end{tabular}
\end{table}

Win rate (\cref{sec:syscon3d-results}) only tests whether each inconsistent group scores worse than consistent inputs. Here we ask a stricter question: does a metric recover the full severity ladder $L_0 < L_1 < L_2 < L_3 < \text{Gaussian noise}$ across the six calibration scene types? We therefore report two complementary rank-based measures.

\paragraph{Kendall's $\tau$.}
For each $K \in \{3, 6, 9, 12, 15, 18, 21\}$, we rank the 6 scene types by their mean metric score and compare that ranking against the ideal ordering above, where lower scores mean better 3D consistency. We average Kendall's $\tau$ across all $K$ values. We use Kendall's $\tau$ rather than Spearman's $\rho$, which is reported for the human-alignment tables in the main paper and appendix, because the target here is a fixed ordinal ladder, not a correlation between two empirical rankings of methods. Every pairwise inversion between scene types is meaningful, and Kendall's $\tau$ directly measures concordant versus discordant pairs. A perfect ordering gives $\tau = 1$, a random ordering is near $0$, and a complete reversal gives $\tau = -1$.

\paragraph{Probabilistic Pairwise Concordance (PPC).}
Kendall's $\tau$ uses only point estimates. PPC accounts for score uncertainty by modeling each scene type with empirical mean $\mu$ and standard deviation $\sigma$. For each pair $(i, j)$ where type $i$ should score lower than type $j$, the probability of correct separation is:
\begin{equation}
  \text{PPC}_{ij} = \Phi\!\left(\frac{\mu_j - \mu_i}{\sqrt{\sigma_i^2 + \sigma_j^2}}\right),
\end{equation}
where $\Phi$ is the standard normal CDF. PPC is averaged over all $\binom{6}{2} = 15$ pairs and all $K$ values. A PPC of $0.5$ indicates chance-level separation, while values near $1.0$ indicate confident correct ordering.

\paragraph{Results.}
\Cref{tab:rank_concordance} evaluates a stricter property than the win-rate analysis in Table 2: rank concordance with the ideal metric ranking introduced in ~\cref{fig:calibration-bars}. TSED and PRISM-MMD clearly lead here; PC-MMD variants are the strongest backbone-based models; and MEt3R still nearly reverses the intended severity ladder. TSED achieves the best overall concordance ($\tau{=}0.70$, PPC${=}0.76$), followed by PRISM-MMD ($\tau{=}0.45$, PPC${=}0.72$). Among the backbone-based metrics, VGGT PC-MMD ($\tau{=}0.30$, PPC${=}0.64$) and Fast3R PC-MMD ($\tau{=}0.28$, PPC${=}0.62$) perform best, while most other variants remain only weakly positive with $\tau$ between $0.05$ and $0.26$. MASt3R W-IMQ, despite the best overall win rate in \cref{tab:metric_calibration_effect}, reaches only moderate concordance ($\tau{=}0.26$, PPC${=}0.59$), showing that separating consistent from inconsistent inputs is easier than recovering the full severity ordering. At the other extreme, MEt3R scores $\tau{=}{-}0.89$ and PPC${=}0.20$, which is close to a complete inversion of the desired ranking. SED is also near chance ($\tau{=}{-}0.03$, PPC${=}0.52$), whereas TSED converts the same geometric signal into a much more reliable ordinal ranking. Overall, the non-learned baselines remain much better aligned with the calibration ladder than the data-driven backbone-based metrics, reinforcing the motivation for COLMAP-based metrics.

\section{Evaluation of Parametric Metric Family on MipNeRF360}
\label{app:mipnerf360-eval}

In \cref{tab:human-rank-corr-mipnerf360}, we evaluate our best performing neural metric from the parametric family, MASt3R-W-IMQ, on NVS renderings of the 8 regression and diffusion-based reconstruction methods on 9 scenes of the MipNeRF360 dataset across 3, 6, and 9 views.

\begin{table}[!htbp]
  \centering
  \caption{\textbf{Metric rankings vs.\ human rankings on MipNeRF360.} IMQ shows the strongest alignment with human judgments, especially at higher view counts, while PRISM is anti-correlated across all splits. We compare MEt3R, PRISM-MMD (PRISM), and MASt3R-W-IMQ (IMQ) rankings of 8 NVS methods on 9 MipNeRF360 scenes across three view-count splits ($K{=}3, 6, 9$). IMQ is selected for its highest overall win rate against inconsistent scene types (\cref{tab:metric_calibration_effect}). Colors encode ranks 1 (best, \colorbox{green!35}{green}) to 8 (worst, \colorbox{red!35}{red}); rows sorted by $K{=}3$ human rank (\textbf{H}). Bottom row: Spearman rank correlation~($\rho$). IMQ achieves the highest $\rho$ at every split, reaching $0.52$ at $K{=}9$. PRISM shows negative correlation across all splits ($\rho \in [-0.40, -0.19]$), in contrast to its moderate positive correlation on DL3DV (\cref{tab:human-rank-corr-dl3dv}). Notably, DepthSplat - ranked 1st by humans - is placed 7th--8th by MEt3R and IMQ at $K{=}3$, but converges to rank 2--3 at higher $K$, suggesting metric-human agreement improves with more input views on MipNeRF360's wider-baseline scenes.}
  \label{tab:human-rank-corr-mipnerf360}
  \resizebox{\columnwidth}{!}{%
    \begin{tabular}{lcccccccccccc}
      \toprule
      Method & \multicolumn{4}{c}{K=3} & \multicolumn{4}{c}{K=6} & \multicolumn{4}{c}{K=9} \\
      \cmidrule(lr){2-5} \cmidrule(lr){6-9} \cmidrule(lr){10-13}
       & H & MEt3R & IMQ & PRISM-MMD & H & MEt3R & IMQ & PRISM-MMD & H & MEt3R & IMQ & PRISM-MMD \\
      \midrule
      DepthSplat & \cellcolor{red!0!green!35}1 & \cellcolor{red!86!green!35}0.360 \textbf{\scriptsize 7} & \cellcolor{red!100!green!35}0.116 \textbf{\scriptsize 8} & \cellcolor{red!71!green!35}0.643 \textbf{\scriptsize 6} & \cellcolor{red!0!green!35}1 & \cellcolor{red!29!green!35}0.329 \textbf{\scriptsize 3} & \cellcolor{red!29!green!35}0.094 \textbf{\scriptsize 3} & \cellcolor{red!71!green!35}0.719 \textbf{\scriptsize 6} & \cellcolor{red!0!green!35}1 & \cellcolor{red!14!green!35}0.325 \textbf{\scriptsize 2} & \cellcolor{red!14!green!35}0.095 \textbf{\scriptsize 2} & \cellcolor{red!86!green!35}0.723 \textbf{\scriptsize 7} \\
      S-V-Camera & \cellcolor{red!14!green!35}2 & \cellcolor{red!0!green!35}0.267 \textbf{\scriptsize 1} & \cellcolor{red!0!green!35}0.069 \textbf{\scriptsize 1} & \cellcolor{red!14!green!35}0.362 \textbf{\scriptsize 2} & \cellcolor{red!43!green!35}4 & \cellcolor{red!0!green!35}0.261 \textbf{\scriptsize 1} & \cellcolor{red!0!green!35}0.071 \textbf{\scriptsize 1} & \cellcolor{red!14!green!35}0.382 \textbf{\scriptsize 2} & \cellcolor{red!43!green!35}4 & \cellcolor{red!0!green!35}0.291 \textbf{\scriptsize 1} & \cellcolor{red!0!green!35}0.076 \textbf{\scriptsize 1} & \cellcolor{red!0!green!35}0.362 \textbf{\scriptsize 1} \\
      ViewCrafter & \cellcolor{red!29!green!35}3 & \cellcolor{red!14!green!35}0.304 \textbf{\scriptsize 2} & \cellcolor{red!14!green!35}0.083 \textbf{\scriptsize 2} & \cellcolor{red!43!green!35}0.500 \textbf{\scriptsize 4} & \cellcolor{red!29!green!35}3 & \cellcolor{red!14!green!35}0.315 \textbf{\scriptsize 2} & \cellcolor{red!14!green!35}0.091 \textbf{\scriptsize 2} & \cellcolor{red!57!green!35}0.601 \textbf{\scriptsize 5} & \cellcolor{red!29!green!35}3 & \cellcolor{red!29!green!35}0.331 \textbf{\scriptsize 3} & \cellcolor{red!29!green!35}0.101 \textbf{\scriptsize 3} & \cellcolor{red!57!green!35}0.590 \textbf{\scriptsize 5} \\
      Long-LRM & \cellcolor{red!43!green!35}4 & \cellcolor{red!100!green!35}0.381 \textbf{\scriptsize 8} & \cellcolor{red!86!green!35}0.114 \textbf{\scriptsize 7} & \cellcolor{red!100!green!35}0.728 \textbf{\scriptsize 8} & \cellcolor{red!14!green!35}2 & \cellcolor{red!100!green!35}0.444 \textbf{\scriptsize 8} & \cellcolor{red!86!green!35}0.150 \textbf{\scriptsize 7} & \cellcolor{red!86!green!35}0.739 \textbf{\scriptsize 7} & \cellcolor{red!14!green!35}2 & \cellcolor{red!100!green!35}0.426 \textbf{\scriptsize 8} & \cellcolor{red!86!green!35}0.149 \textbf{\scriptsize 7} & \cellcolor{red!71!green!35}0.702 \textbf{\scriptsize 6} \\
      MVSplat360 & \cellcolor{red!57!green!35}5 & \cellcolor{red!71!green!35}0.337 \textbf{\scriptsize 6} & \cellcolor{red!43!green!35}0.103 \textbf{\scriptsize 4} & \cellcolor{red!86!green!35}0.706 \textbf{\scriptsize 7} & \cellcolor{red!71!green!35}6 & \cellcolor{red!57!green!35}0.349 \textbf{\scriptsize 5} & \cellcolor{red!43!green!35}0.108 \textbf{\scriptsize 4} & \cellcolor{red!100!green!35}0.748 \textbf{\scriptsize 8} & \cellcolor{red!71!green!35}6 & \cellcolor{red!71!green!35}0.389 \textbf{\scriptsize 6} & \cellcolor{red!71!green!35}0.144 \textbf{\scriptsize 6} & \cellcolor{red!100!green!35}0.736 \textbf{\scriptsize 8} \\
      MVGenMaster & \cellcolor{red!71!green!35}6 & \cellcolor{red!29!green!35}0.329 \textbf{\scriptsize 3} & \cellcolor{red!29!green!35}0.097 \textbf{\scriptsize 3} & \cellcolor{red!29!green!35}0.375 \textbf{\scriptsize 3} & \cellcolor{red!57!green!35}5 & \cellcolor{red!43!green!35}0.342 \textbf{\scriptsize 4} & \cellcolor{red!57!green!35}0.114 \textbf{\scriptsize 5} & \cellcolor{red!29!green!35}0.392 \textbf{\scriptsize 3} & \cellcolor{red!57!green!35}5 & \cellcolor{red!43!green!35}0.342 \textbf{\scriptsize 4} & \cellcolor{red!43!green!35}0.109 \textbf{\scriptsize 4} & \cellcolor{red!14!green!35}0.368 \textbf{\scriptsize 2} \\
      NVS-Solver & \cellcolor{red!86!green!35}7 & \cellcolor{red!57!green!35}0.330 \textbf{\scriptsize 5} & \cellcolor{red!57!green!35}0.107 \textbf{\scriptsize 5} & \cellcolor{red!0!green!35}0.254 \textbf{\scriptsize 1} & \cellcolor{red!86!green!35}7 & \cellcolor{red!86!green!35}0.388 \textbf{\scriptsize 7} & \cellcolor{red!100!green!35}0.158 \textbf{\scriptsize 8} & \cellcolor{red!0!green!35}0.262 \textbf{\scriptsize 1} & \cellcolor{red!100!green!35}8 & \cellcolor{red!86!green!35}0.407 \textbf{\scriptsize 7} & \cellcolor{red!100!green!35}0.174 \textbf{\scriptsize 8} & \cellcolor{red!29!green!35}0.454 \textbf{\scriptsize 3} \\
      Difix3D & \cellcolor{red!100!green!35}8 & \cellcolor{red!43!green!35}0.329 \textbf{\scriptsize 4} & \cellcolor{red!71!green!35}0.113 \textbf{\scriptsize 6} & \cellcolor{red!57!green!35}0.505 \textbf{\scriptsize 5} & \cellcolor{red!100!green!35}8 & \cellcolor{red!71!green!35}0.374 \textbf{\scriptsize 6} & \cellcolor{red!71!green!35}0.139 \textbf{\scriptsize 6} & \cellcolor{red!43!green!35}0.430 \textbf{\scriptsize 4} & \cellcolor{red!86!green!35}7 & \cellcolor{red!57!green!35}0.369 \textbf{\scriptsize 5} & \cellcolor{red!57!green!35}0.142 \textbf{\scriptsize 5} & \cellcolor{red!43!green!35}0.499 \textbf{\scriptsize 4} \\
      \midrule
      Rank Corr.\ ($\rho$) & -- & 0.00 & \textbf{0.07} & -0.19 & -- & 0.33 & \textbf{0.43} & -0.40 & -- & 0.38 & \textbf{0.52} & -0.33 \\
      \bottomrule
    \end{tabular}%
  }
\end{table}

\section{COLMAP-based 3D consistency metrics: extended notation and coverage}
\label{app:colmap-metrics-apx}

This section supplements the COLMAP-based 3D consistency metrics introduced in Sec.~\ref{sec:colmap-metrics} of the main paper. Definitions of the per-pixel quality $q_v(\mathbf{u})$, $\mathrm{Density}_v$, $\mathrm{Consistency}_v$, $\mathrm{GPC}$, $\mathrm{ICM}$, $\mathrm{ICM}_{\text{all}}$, and the coverage-weighted score W-GPC follow exactly the formulas given there; we restate them here only when needed for the additional notation and the angular-coverage construction below. 

\subsection{Motivation: why classical geometric verification}
\label{app:colmap-motivation}
The robustness failures (\cref{sec:syscon3d-results}, \cref{tab:metric_calibration_effect}) and limited human alignment (\cref{tab:human-rank-corr-dl3dv}) share a common cause: data-driven reconstruction models always produce a reconstruction, even when no coherent 3D interpretation exists. Notable exceptions are the non-learned baselines in \cref{tab:metric_calibration_effect}: SIFT-SED achieves $d{=}+161.74$ on Gaussian noise because SIFT simply finds no keypoints on structureless inputs, and TSED achieves 54\% overall win rate with its strongest results on $L_2$ (86\%) and Gaussian noise (100\%). This suggests that classical geometric verification, which can explicitly \emph{fail} on inconsistent inputs, provides a more principled foundation. COLMAP~\cite{schoenberger2016sfm} offers this property: robust feature matching and geometric verification (RANSAC, epipolar constraints) reject spurious correspondences, and COLMAP refuses to register images when insufficient inlier matches are found -- penalizing inputs that lack verifiable geometric structure rather than hallucinating one.

\subsection{Notation and COLMAP outputs}
Let $A$ denote the set of all \emph{attempted} images in a COLMAP run (obtained from \texttt{database.db}). Let $R\subseteq A$ be the subset of images registered by sparse SfM (read from \texttt{sparse/0/images.bin}). Let $D\subseteq R$ be the subset of registered images for which COLMAP produces readable dense maps \texttt{$\cdot$.geometric.bin} and \texttt{$\cdot$.photometric.bin}. We additionally report the sparse registration rate
\begin{equation}
\mathrm{registration\_rate}=\frac{|R|}{|A|}.
\end{equation}
For each $v\in D$, COLMAP provides a geometric-consistency depth map $\mathcal{D}^{(v)}_{g}$ (\texttt{*.geometric.bin}) and a photometric depth hypothesis $\mathcal{D}^{(v)}_{p}$ (\texttt{*.photometric.bin}). Let $P_v$ be the pixel set of view $v$ and $|P_v|=W_vH_v$. For the survivor-only metrics (GPC/ICM) we use the dense-map resolution. For the attempted-set normalization in $\mathrm{ICM}_{\text{all}}$, we use $(W_v,H_v)$ from the COLMAP database, which keeps the denominator consistent across methods with different output resolutions. Recall that under $\mathrm{ICM}_{\text{all}}$, images in $A\setminus R$ (SfM failures) and $R\setminus D$ (dense failures) contribute $0$ to the numerator but still contribute their pixels to the denominator, so $\mathrm{ICM}_{\text{all}}$ simultaneously captures (i) reconstructability and (ii) geometric fidelity on the explainable subset.

\subsection{Angular coverage}
Coverage measures how broadly the registered cameras span azimuth around the reconstructed object. Let $\{\mathbf{x}_j\}$ be sparse 3D points; we define the object center as the coordinate-wise median:
\begin{equation}
\mathbf{o}=\mathrm{median}\{\mathbf{x}_j\}.
\end{equation}
For each registered image $i\in R$, with quaternion $\mathbf{q}_i$ and translation $\mathbf{t}_i$, we form $R_i=\mathrm{quat2mat}(\mathbf{q}_i)$ and compute the camera center:
\begin{equation}
\mathbf{c}_i=-R_i^\top\mathbf{t}_i.
\end{equation}
We compute azimuth angles $\theta_i\in[0,360)$ around $\mathbf{o}$ either in (i) the top-2 PCA plane of camera centers (fallback to world $XZ$ if degenerate), or (ii) directly in world $XZ$. After sorting $\theta_{(1)}\le\cdots\le\theta_{(m)}$, define circular gaps:
\begin{equation}
g_k=\theta_{(k+1)}-\theta_{(k)}\ (k=1,\dots,m-1),\qquad
g_m=(360-\theta_{(m)})+\theta_{(1)}.
\end{equation}
Coverage is:
\begin{equation}
\mathrm{coverage}=360^\circ-\max_{1\le k\le m} g_k.
\end{equation}
Intuitively, coverage is high when registered views wrap around the object and low when only a narrow arc of viewpoints is geometrically supported. The W-GPC of Sec.~\ref{sec:colmap-metrics} is then $\mathrm{gpc}\cdot\mathrm{coverage}/360^\circ$.

\paragraph{Compute cost.}
Running our COLMAP-based metrics takes on average $\sim$5 minutes on our evaluated datasets on an RTX 3090.

If SfM registration or MVS densification fails for an attempted view, we treat that view as contributing zero verified 3D support rather than assigning a missing score; this convention makes reconstruction failure part of the consistency metric.

\subsection{COLMAP failures}
\begin{figure}[!h]
    \centering
    \includegraphics[width=\linewidth]{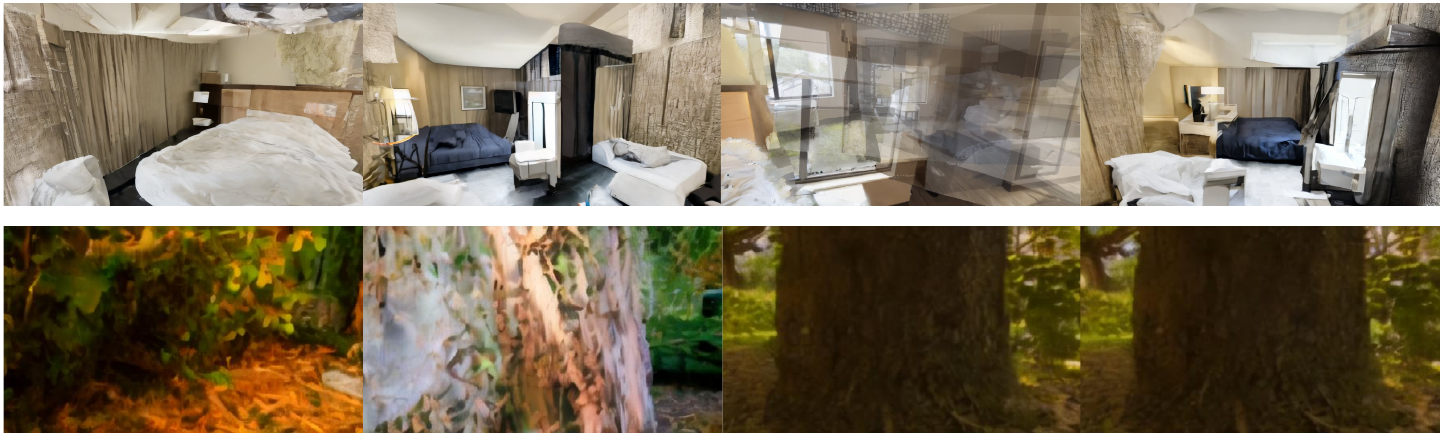}
    \caption{Examples of COLMAP failures from our evaluation on DL3DV and MipNeRF360 generations. We observed that COLMAP failures corresponded to poor quality generations with little observable shared 3D consistency, as seen in the images.(Top) NVS Solver on DL3DV (Bottom) MVSplat360 on MipNeRF360 (stump)}
    \label{fig:colmap-failures}
\end{figure}

Approximately $1\%$ of our NVS experiments on DL3DV and Mip-NeRF360 resulted in COLMAP failures. Although COLMAP can fail under standard SfM failure modes such as low texture, limited overlap, repetitive structure, or large illumination/viewpoint changes, we found that the failures in our setting were primarily caused by low-quality generated views, as illustrated in \autoref{fig:colmap-failures}. To rule out ordinary SIFT matching failures, we additionally repeated the affected evaluations using stronger correspondence pipelines, including SIFT+LightGlue and ALIKED+LightGlue. These alternatives substantially relax the dependence on brute-force SIFT matching, but did not produce statistically significant changes relative to the standard SIFT-based results. This suggests that the observed failures reflect degradation in the generated imagery rather than limitations of the COLMAP matching backend.

\subsection{COLMAP Metric Results}
\label{app:colmap-results}

This section collects extended results for the COLMAP-based metrics defined in Sec.~\ref{sec:colmap-metrics}. \cref{tab:colmap-human-rank-dl3dv-all,tab:colmap-human-rank-mip-all,tab:colmap-human-rank-dl3dv-all-k3,tab:colmap-human-rank-dl3dv-all-k6,tab:colmap-human-rank-mip-all-k3} show that COLMAP metrics (W-GPC) are more human-aligned in ranking 3D consistency of NVS methods than the data-driven baselines.

\begin{table}
  \centering
  \caption{COLMAP reconstruction quality vs.\ human rankings of 8 NVS methods on DL3DV (24 scenes), full-metric version for the $K{=}9$ view split. Colors encode ranks 1 (best, \colorbox{green!35}{green}) to 8 (worst, \colorbox{red!35}{red}); rows sorted by human rank (\textbf{H}). Bottom row: Spearman rank correlation ($\rho$) between human and metric-induced rankings.}
  \label{tab:colmap-human-rank-dl3dv-all}
  \resizebox{\columnwidth}{!}{%
    \begin{tabular}{lcccccccccc}
      \toprule
      Method & H$\downarrow$ & W-GPC$\uparrow$ & ICM$\uparrow$ & Ang.\ Cov.$\uparrow$ & GPC$\uparrow$ & Avg.\ Den.$\uparrow$ & Avg.\ Cons.$\uparrow$ & GPC All$\uparrow$ & ICM All$\uparrow$ & Reg.\ Rate$\uparrow$ \\
      \midrule
      DepthSplat & \cellcolor{red!0!green!35}1 & \cellcolor{red!0!green!35}0.288 {\scriptsize$\pm$ 0.25} \textbf{\scriptsize 1} & \cellcolor{red!0!green!35}0.531 {\scriptsize$\pm$ 0.16} \textbf{\scriptsize 1} & \cellcolor{red!0!green!35}170.4 {\scriptsize$\pm$ 126.89} \textbf{\scriptsize 1} & \cellcolor{red!0!green!35}0.531 {\scriptsize$\pm$ 0.16} \textbf{\scriptsize 1} & \cellcolor{red!0!green!35}0.622 {\scriptsize$\pm$ 0.16} \textbf{\scriptsize 1} & \cellcolor{red!14!green!35}0.833 {\scriptsize$\pm$ 0.09} \textbf{\scriptsize 2} & \cellcolor{red!0!green!35}0.294 {\scriptsize$\pm$ 0.25} \textbf{\scriptsize 1} & \cellcolor{red!0!green!35}0.294 {\scriptsize$\pm$ 0.25} \textbf{\scriptsize 1} & \cellcolor{red!0!green!35}0.496 {\scriptsize$\pm$ 0.35} \textbf{\scriptsize 1} \\
      Long-LRM & \cellcolor{red!14!green!35}2 & \cellcolor{red!14!green!35}0.136 {\scriptsize$\pm$ 0.20} \textbf{\scriptsize 2} & \cellcolor{red!14!green!35}0.404 {\scriptsize$\pm$ 0.17} \textbf{\scriptsize 2} & \cellcolor{red!14!green!35}94.1 {\scriptsize$\pm$ 115.33} \textbf{\scriptsize 2} & \cellcolor{red!14!green!35}0.415 {\scriptsize$\pm$ 0.16} \textbf{\scriptsize 2} & \cellcolor{red!29!green!35}0.496 {\scriptsize$\pm$ 0.16} \textbf{\scriptsize 3} & \cellcolor{red!29!green!35}0.826 {\scriptsize$\pm$ 0.10} \textbf{\scriptsize 3} & \cellcolor{red!14!green!35}0.138 {\scriptsize$\pm$ 0.20} \textbf{\scriptsize 2} & \cellcolor{red!14!green!35}0.136 {\scriptsize$\pm$ 0.20} \textbf{\scriptsize 2} & \cellcolor{red!14!green!35}0.294 {\scriptsize$\pm$ 0.36} \textbf{\scriptsize 2} \\
      S-V-Camera & \cellcolor{red!29!green!35}3 & \cellcolor{red!29!green!35}0.056 {\scriptsize$\pm$ 0.12} \textbf{\scriptsize 3} & \cellcolor{red!29!green!35}0.351 {\scriptsize$\pm$ 0.16} \textbf{\scriptsize 3} & \cellcolor{red!29!green!35}53.4 {\scriptsize$\pm$ 95.19} \textbf{\scriptsize 3} & \cellcolor{red!57!green!35}0.353 {\scriptsize$\pm$ 0.16} \textbf{\scriptsize 5} & \cellcolor{red!57!green!35}0.421 {\scriptsize$\pm$ 0.18} \textbf{\scriptsize 5} & \cellcolor{red!0!green!35}0.844 {\scriptsize$\pm$ 0.07} \textbf{\scriptsize 1} & \cellcolor{red!29!green!35}0.052 {\scriptsize$\pm$ 0.14} \textbf{\scriptsize 3} & \cellcolor{red!29!green!35}0.052 {\scriptsize$\pm$ 0.14} \textbf{\scriptsize 3} & \cellcolor{red!43!green!35}0.126 {\scriptsize$\pm$ 0.27} \textbf{\scriptsize 4} \\
      ViewCrafter & \cellcolor{red!43!green!35}4 & \cellcolor{red!43!green!35}0.036 {\scriptsize$\pm$ 0.06} \textbf{\scriptsize 4} & \cellcolor{red!86!green!35}0.238 {\scriptsize$\pm$ 0.12} \textbf{\scriptsize 7} & \cellcolor{red!43!green!35}46.2 {\scriptsize$\pm$ 58.19} \textbf{\scriptsize 4} & \cellcolor{red!100!green!35}0.255 {\scriptsize$\pm$ 0.11} \textbf{\scriptsize 8} & \cellcolor{red!100!green!35}0.328 {\scriptsize$\pm$ 0.14} \textbf{\scriptsize 8} & \cellcolor{red!71!green!35}0.793 {\scriptsize$\pm$ 0.10} \textbf{\scriptsize 6} & \cellcolor{red!79!green!35}0.014 {\scriptsize$\pm$ 0.02} \textbf{\scriptsize 6.5} & \cellcolor{red!86!green!35}0.011 {\scriptsize$\pm$ 0.01} \textbf{\scriptsize 7} & \cellcolor{red!71!green!35}0.113 {\scriptsize$\pm$ 0.18} \textbf{\scriptsize 6} \\
      MVGenMaster & \cellcolor{red!57!green!35}5 & \cellcolor{red!57!green!35}0.035 {\scriptsize$\pm$ 0.06} \textbf{\scriptsize 5} & \cellcolor{red!71!green!35}0.302 {\scriptsize$\pm$ 0.18} \textbf{\scriptsize 6} & \cellcolor{red!57!green!35}42.2 {\scriptsize$\pm$ 51.68} \textbf{\scriptsize 5} & \cellcolor{red!71!green!35}0.313 {\scriptsize$\pm$ 0.17} \textbf{\scriptsize 6} & \cellcolor{red!71!green!35}0.415 {\scriptsize$\pm$ 0.22} \textbf{\scriptsize 6} & \cellcolor{red!86!green!35}0.772 {\scriptsize$\pm$ 0.11} \textbf{\scriptsize 7} & \cellcolor{red!43!green!35}0.039 {\scriptsize$\pm$ 0.08} \textbf{\scriptsize 4} & \cellcolor{red!43!green!35}0.046 {\scriptsize$\pm$ 0.09} \textbf{\scriptsize 4} & \cellcolor{red!29!green!35}0.143 {\scriptsize$\pm$ 0.21} \textbf{\scriptsize 3} \\
      MVSplat360 & \cellcolor{red!71!green!35}6 & \cellcolor{red!71!green!35}0.027 {\scriptsize$\pm$ 0.06} \textbf{\scriptsize 6} & \cellcolor{red!57!green!35}0.319 {\scriptsize$\pm$ 0.19} \textbf{\scriptsize 5} & \cellcolor{red!86!green!35}27.2 {\scriptsize$\pm$ 31.09} \textbf{\scriptsize 7} & \cellcolor{red!43!green!35}0.370 {\scriptsize$\pm$ 0.19} \textbf{\scriptsize 4} & \cellcolor{red!43!green!35}0.466 {\scriptsize$\pm$ 0.22} \textbf{\scriptsize 4} & \cellcolor{red!43!green!35}0.805 {\scriptsize$\pm$ 0.12} \textbf{\scriptsize 4} & \cellcolor{red!79!green!35}0.014 {\scriptsize$\pm$ 0.02} \textbf{\scriptsize 6.5} & \cellcolor{red!71!green!35}0.019 {\scriptsize$\pm$ 0.02} \textbf{\scriptsize 6} & \cellcolor{red!86!green!35}0.060 {\scriptsize$\pm$ 0.05} \textbf{\scriptsize 7} \\
      NVS-Solver & \cellcolor{red!86!green!35}7 & \cellcolor{red!100!green!35}0.018 {\scriptsize$\pm$ 0.04} \textbf{\scriptsize 8} & \cellcolor{red!43!green!35}0.350 {\scriptsize$\pm$ 0.25} \textbf{\scriptsize 4} & \cellcolor{red!100!green!35}18.9 {\scriptsize$\pm$ 36.75} \textbf{\scriptsize 8} & \cellcolor{red!29!green!35}0.406 {\scriptsize$\pm$ 0.23} \textbf{\scriptsize 3} & \cellcolor{red!14!green!35}0.502 {\scriptsize$\pm$ 0.23} \textbf{\scriptsize 2} & \cellcolor{red!57!green!35}0.795 {\scriptsize$\pm$ 0.13} \textbf{\scriptsize 5} & \cellcolor{red!57!green!35}0.026 {\scriptsize$\pm$ 0.06} \textbf{\scriptsize 5} & \cellcolor{red!57!green!35}0.028 {\scriptsize$\pm$ 0.07} \textbf{\scriptsize 5} & \cellcolor{red!57!green!35}0.116 {\scriptsize$\pm$ 0.17} \textbf{\scriptsize 5} \\
      Difix3D & \cellcolor{red!100!green!35}8 & \cellcolor{red!86!green!35}0.019 {\scriptsize$\pm$ 0.01} \textbf{\scriptsize 7} & \cellcolor{red!100!green!35}0.237 {\scriptsize$\pm$ 0.13} \textbf{\scriptsize 8} & \cellcolor{red!71!green!35}32.6 {\scriptsize$\pm$ 25.98} \textbf{\scriptsize 6} & \cellcolor{red!86!green!35}0.261 {\scriptsize$\pm$ 0.13} \textbf{\scriptsize 7} & \cellcolor{red!86!green!35}0.348 {\scriptsize$\pm$ 0.15} \textbf{\scriptsize 7} & \cellcolor{red!100!green!35}0.748 {\scriptsize$\pm$ 0.13} \textbf{\scriptsize 8} & \cellcolor{red!100!green!35}0.007 {\scriptsize$\pm$ 0.01} \textbf{\scriptsize 8} & \cellcolor{red!100!green!35}0.006 {\scriptsize$\pm$ 0.01} \textbf{\scriptsize 8} & \cellcolor{red!100!green!35}0.055 {\scriptsize$\pm$ 0.03} \textbf{\scriptsize 8} \\
      \midrule
      Rank Corr.\ ($\rho$)$\uparrow$ & -- & \textbf{0.976} & 0.762 & 0.929 & 0.500 & 0.381 & 0.738 & 0.862 & 0.833 & 0.833 \\
      \bottomrule
    \end{tabular}%
  }
\end{table}
\begin{table}
  \centering
  \caption{COLMAP reconstruction quality vs.\ human rankings of 8 NVS methods on MipNeRF360 (9 scenes), full-metric version for the $K{=}9$ view split. Colors encode ranks 1 (best, \colorbox{green!35}{green}) to 8 (worst, \colorbox{red!35}{red}); rows sorted by human rank (\textbf{H}). Bottom row: Spearman rank correlation ($\rho$) between human and metric-induced rankings.}
  \label{tab:colmap-human-rank-mip-all}
  \resizebox{\columnwidth}{!}{%
    \begin{tabular}{lcccccccccc}
      \toprule
      Method & H$\downarrow$ & W-GPC$\uparrow$ & ICM$\uparrow$ & Ang.\ Cov.$\uparrow$ & GPC$\uparrow$ & Avg.\ Den.$\uparrow$ & Avg.\ Cons.$\uparrow$ & GPC All$\uparrow$ & ICM All$\uparrow$ & Reg.\ Rate$\uparrow$ \\
      \midrule
      DepthSplat & \cellcolor{red!0!green!35}1 & \cellcolor{red!0!green!35}0.475 {\scriptsize$\pm$ 0.31} \textbf{\scriptsize 1} & \cellcolor{red!0!green!35}0.636 {\scriptsize$\pm$ 0.17} \textbf{\scriptsize 1} & \cellcolor{red!0!green!35}247.0 {\scriptsize$\pm$ 145.55} \textbf{\scriptsize 1} & \cellcolor{red!0!green!35}0.636 {\scriptsize$\pm$ 0.17} \textbf{\scriptsize 1} & \cellcolor{red!0!green!35}0.747 {\scriptsize$\pm$ 0.17} \textbf{\scriptsize 1} & \cellcolor{red!14!green!35}0.835 {\scriptsize$\pm$ 0.08} \textbf{\scriptsize 2} & \cellcolor{red!0!green!35}0.427 {\scriptsize$\pm$ 0.33} \textbf{\scriptsize 1} & \cellcolor{red!0!green!35}0.427 {\scriptsize$\pm$ 0.33} \textbf{\scriptsize 1} & \cellcolor{red!0!green!35}0.621 {\scriptsize$\pm$ 0.43} \textbf{\scriptsize 1} \\
      Long-LRM & \cellcolor{red!14!green!35}2 & \cellcolor{red!14!green!35}0.128 {\scriptsize$\pm$ 0.26} \textbf{\scriptsize 2} & \cellcolor{red!14!green!35}0.378 {\scriptsize$\pm$ 0.25} \textbf{\scriptsize 2} & \cellcolor{red!14!green!35}79.3 {\scriptsize$\pm$ 110.39} \textbf{\scriptsize 2} & \cellcolor{red!29!green!35}0.381 {\scriptsize$\pm$ 0.25} \textbf{\scriptsize 3} & \cellcolor{red!29!green!35}0.443 {\scriptsize$\pm$ 0.27} \textbf{\scriptsize 3} & \cellcolor{red!29!green!35}0.829 {\scriptsize$\pm$ 0.09} \textbf{\scriptsize 3} & \cellcolor{red!14!green!35}0.135 {\scriptsize$\pm$ 0.26} \textbf{\scriptsize 2} & \cellcolor{red!14!green!35}0.136 {\scriptsize$\pm$ 0.27} \textbf{\scriptsize 2} & \cellcolor{red!14!green!35}0.250 {\scriptsize$\pm$ 0.33} \textbf{\scriptsize 2} \\
      ViewCrafter & \cellcolor{red!29!green!35}3 & \cellcolor{red!43!green!35}0.017 {\scriptsize$\pm$ 0.02} \textbf{\scriptsize 4} & \cellcolor{red!57!green!35}0.275 {\scriptsize$\pm$ 0.18} \textbf{\scriptsize 5} & \cellcolor{red!29!green!35}25.5 {\scriptsize$\pm$ 18.88} \textbf{\scriptsize 3} & \cellcolor{red!71!green!35}0.274 {\scriptsize$\pm$ 0.18} \textbf{\scriptsize 6} & \cellcolor{red!71!green!35}0.362 {\scriptsize$\pm$ 0.23} \textbf{\scriptsize 6} & \cellcolor{red!57!green!35}0.784 {\scriptsize$\pm$ 0.12} \textbf{\scriptsize 5} & \cellcolor{red!71!green!35}0.009 {\scriptsize$\pm$ 0.01} \textbf{\scriptsize 6} & \cellcolor{red!100!green!35}0.007 {\scriptsize$\pm$ 0.01} \textbf{\scriptsize 8} & \cellcolor{red!29!green!35}0.083 {\scriptsize$\pm$ 0.11} \textbf{\scriptsize 3} \\
      S-V-Camera & \cellcolor{red!43!green!35}4 & \cellcolor{red!29!green!35}0.021 {\scriptsize$\pm$ 0.04} \textbf{\scriptsize 3} & \cellcolor{red!29!green!35}0.363 {\scriptsize$\pm$ 0.15} \textbf{\scriptsize 3} & \cellcolor{red!57!green!35}21.1 {\scriptsize$\pm$ 41.89} \textbf{\scriptsize 5} & \cellcolor{red!43!green!35}0.361 {\scriptsize$\pm$ 0.15} \textbf{\scriptsize 4} & \cellcolor{red!43!green!35}0.437 {\scriptsize$\pm$ 0.18} \textbf{\scriptsize 4} & \cellcolor{red!0!green!35}0.839 {\scriptsize$\pm$ 0.12} \textbf{\scriptsize 1} & \cellcolor{red!29!green!35}0.021 {\scriptsize$\pm$ 0.02} \textbf{\scriptsize 3} & \cellcolor{red!29!green!35}0.021 {\scriptsize$\pm$ 0.02} \textbf{\scriptsize 3} & \cellcolor{red!43!green!35}0.077 {\scriptsize$\pm$ 0.09} \textbf{\scriptsize 4} \\
      MVGenMaster & \cellcolor{red!57!green!35}5 & \cellcolor{red!57!green!35}0.015 {\scriptsize$\pm$ 0.02} \textbf{\scriptsize 5} & \cellcolor{red!71!green!35}0.267 {\scriptsize$\pm$ 0.25} \textbf{\scriptsize 6} & \cellcolor{red!43!green!35}25.2 {\scriptsize$\pm$ 28.04} \textbf{\scriptsize 4} & \cellcolor{red!57!green!35}0.283 {\scriptsize$\pm$ 0.24} \textbf{\scriptsize 5} & \cellcolor{red!57!green!35}0.410 {\scriptsize$\pm$ 0.23} \textbf{\scriptsize 5} & \cellcolor{red!86!green!35}0.637 {\scriptsize$\pm$ 0.19} \textbf{\scriptsize 7} & \cellcolor{red!71!green!35}0.009 {\scriptsize$\pm$ 0.02} \textbf{\scriptsize 6} & \cellcolor{red!71!green!35}0.013 {\scriptsize$\pm$ 0.02} \textbf{\scriptsize 6} & \cellcolor{red!86!green!35}0.056 {\scriptsize$\pm$ 0.06} \textbf{\scriptsize 7} \\
      MVSplat360 & \cellcolor{red!71!green!35}6 & \cellcolor{red!71!green!35}0.013 {\scriptsize$\pm$ 0.01} \textbf{\scriptsize 6} & \cellcolor{red!43!green!35}0.356 {\scriptsize$\pm$ 0.24} \textbf{\scriptsize 4} & \cellcolor{red!86!green!35}13.5 {\scriptsize$\pm$ 13.40} \textbf{\scriptsize 7} & \cellcolor{red!14!green!35}0.420 {\scriptsize$\pm$ 0.22} \textbf{\scriptsize 2} & \cellcolor{red!14!green!35}0.527 {\scriptsize$\pm$ 0.23} \textbf{\scriptsize 2} & \cellcolor{red!43!green!35}0.790 {\scriptsize$\pm$ 0.14} \textbf{\scriptsize 4} & \cellcolor{red!43!green!35}0.012 {\scriptsize$\pm$ 0.01} \textbf{\scriptsize 4} & \cellcolor{red!50!green!35}0.018 {\scriptsize$\pm$ 0.03} \textbf{\scriptsize 4} & \cellcolor{red!100!green!35}0.044 {\scriptsize$\pm$ 0.02} \textbf{\scriptsize 8} \\
      Difix3D & \cellcolor{red!86!green!35}7 & \cellcolor{red!86!green!35}0.010 {\scriptsize$\pm$ 0.01} \textbf{\scriptsize 7} & \cellcolor{red!86!green!35}0.203 {\scriptsize$\pm$ 0.19} \textbf{\scriptsize 7} & \cellcolor{red!71!green!35}16.1 {\scriptsize$\pm$ 12.03} \textbf{\scriptsize 6} & \cellcolor{red!86!green!35}0.245 {\scriptsize$\pm$ 0.18} \textbf{\scriptsize 7} & \cellcolor{red!86!green!35}0.337 {\scriptsize$\pm$ 0.18} \textbf{\scriptsize 7} & \cellcolor{red!71!green!35}0.692 {\scriptsize$\pm$ 0.17} \textbf{\scriptsize 6} & \cellcolor{red!71!green!35}0.009 {\scriptsize$\pm$ 0.02} \textbf{\scriptsize 6} & \cellcolor{red!50!green!35}0.018 {\scriptsize$\pm$ 0.04} \textbf{\scriptsize 5} & \cellcolor{red!71!green!35}0.061 {\scriptsize$\pm$ 0.04} \textbf{\scriptsize 6} \\
      NVS-Solver & \cellcolor{red!100!green!35}8 & \cellcolor{red!100!green!35}0.003 {\scriptsize$\pm$ 0.00} \textbf{\scriptsize 8} & \cellcolor{red!100!green!35}0.142 {\scriptsize$\pm$ 0.09} \textbf{\scriptsize 8} & \cellcolor{red!100!green!35}13.1 {\scriptsize$\pm$ 5.34} \textbf{\scriptsize 8} & \cellcolor{red!100!green!35}0.177 {\scriptsize$\pm$ 0.07} \textbf{\scriptsize 8} & \cellcolor{red!100!green!35}0.299 {\scriptsize$\pm$ 0.14} \textbf{\scriptsize 8} & \cellcolor{red!100!green!35}0.625 {\scriptsize$\pm$ 0.14} \textbf{\scriptsize 8} & \cellcolor{red!100!green!35}0.007 {\scriptsize$\pm$ 0.01} \textbf{\scriptsize 8} & \cellcolor{red!86!green!35}0.008 {\scriptsize$\pm$ 0.01} \textbf{\scriptsize 7} & \cellcolor{red!57!green!35}0.070 {\scriptsize$\pm$ 0.02} \textbf{\scriptsize 5} \\
      \midrule
      Rank Corr.\ ($\rho$)$\uparrow$ & -- & \textbf{0.976} & 0.881 & 0.952 & 0.690 & 0.690 & 0.714 & 0.805 & 0.563 & 0.786 \\
      \bottomrule
    \end{tabular}%
  }
\end{table}

\begin{table}
  \centering
  \caption{COLMAP reconstruction quality vs.\ human rankings of 8 NVS methods on DL3DV, full-metric version for the $K{=}6$ view split. Values are mean $\pm$ std over scenes. Colors encode ranks 1 (best, \colorbox{green!35}{green}) to 8 (worst, \colorbox{red!35}{red}); rows sorted by human rank (\textbf{H}). Bottom row: Spearman rank correlation ($\rho$) between human and metric-induced rankings.}
  \label{tab:colmap-human-rank-dl3dv-all-k6}
  \resizebox{\columnwidth}{!}{%
    \begin{tabular}{lcccccccccc}
      \toprule
      Method & H$\downarrow$ & W-GPC$\uparrow$ & ICM$\uparrow$ & Ang.\ Cov.$\uparrow$ & GPC$\uparrow$ & Avg.\ Den.$\uparrow$ & Avg.\ Cons.$\uparrow$ & GPC All$\uparrow$ & ICM All$\uparrow$ & Reg.\ Rate$\uparrow$ \\
      \midrule
      DepthSplat & \cellcolor{red!0!green!35}1 & \cellcolor{red!0!green!35}0.251 $\pm$ {\tiny 0.30} \textbf{\scriptsize 1} & \cellcolor{red!0!green!35}0.582 $\pm$ {\tiny 0.22} \textbf{\scriptsize 1} & \cellcolor{red!0!green!35}128.985 $\pm$ {\tiny 140.26} \textbf{\scriptsize 1} & \cellcolor{red!0!green!35}0.585 $\pm$ {\tiny 0.22} \textbf{\scriptsize 1} & \cellcolor{red!14!green!35}0.671 $\pm$ {\tiny 0.23} \textbf{\scriptsize 2} & \cellcolor{red!29!green!35}0.847 $\pm$ {\tiny 0.10} \textbf{\scriptsize 3} & \cellcolor{red!0!green!35}0.238 $\pm$ {\tiny 0.29} \textbf{\scriptsize 1} & \cellcolor{red!0!green!35}0.238 $\pm$ {\tiny 0.29} \textbf{\scriptsize 1} & \cellcolor{red!0!green!35}0.381 $\pm$ {\tiny 0.40} \textbf{\scriptsize 1} \\
      S-V-Camera & \cellcolor{red!14!green!35}2 & \cellcolor{red!29!green!35}0.060 $\pm$ {\tiny 0.13} \textbf{\scriptsize 3} & \cellcolor{red!43!green!35}0.386 $\pm$ {\tiny 0.19} \textbf{\scriptsize 4} & \cellcolor{red!29!green!35}54.748 $\pm$ {\tiny 96.31} \textbf{\scriptsize 3} & \cellcolor{red!43!green!35}0.398 $\pm$ {\tiny 0.18} \textbf{\scriptsize 4} & \cellcolor{red!43!green!35}0.464 $\pm$ {\tiny 0.20} \textbf{\scriptsize 4} & \cellcolor{red!14!green!35}0.855 $\pm$ {\tiny 0.07} \textbf{\scriptsize 2} & \cellcolor{red!29!green!35}0.053 $\pm$ {\tiny 0.14} \textbf{\scriptsize 3} & \cellcolor{red!29!green!35}0.053 $\pm$ {\tiny 0.14} \textbf{\scriptsize 3} & \cellcolor{red!29!green!35}0.141 $\pm$ {\tiny 0.28} \textbf{\scriptsize 3} \\
      Long-LRM & \cellcolor{red!29!green!35}3 & \cellcolor{red!14!green!35}0.113 $\pm$ {\tiny 0.18} \textbf{\scriptsize 2} & \cellcolor{red!57!green!35}0.334 $\pm$ {\tiny 0.14} \textbf{\scriptsize 5} & \cellcolor{red!14!green!35}89.564 $\pm$ {\tiny 120.46} \textbf{\scriptsize 2} & \cellcolor{red!57!green!35}0.337 $\pm$ {\tiny 0.14} \textbf{\scriptsize 5} & \cellcolor{red!57!green!35}0.445 $\pm$ {\tiny 0.16} \textbf{\scriptsize 5} & \cellcolor{red!71!green!35}0.763 $\pm$ {\tiny 0.14} \textbf{\scriptsize 6} & \cellcolor{red!14!green!35}0.114 $\pm$ {\tiny 0.18} \textbf{\scriptsize 2} & \cellcolor{red!14!green!35}0.113 $\pm$ {\tiny 0.18} \textbf{\scriptsize 2} & \cellcolor{red!14!green!35}0.281 $\pm$ {\tiny 0.35} \textbf{\scriptsize 2} \\
      ViewCrafter & \cellcolor{red!43!green!35}4 & \cellcolor{red!57!green!35}0.035 $\pm$ {\tiny 0.07} \textbf{\scriptsize 5} & \cellcolor{red!100!green!35}0.256 $\pm$ {\tiny 0.17} \textbf{\scriptsize 8} & \cellcolor{red!57!green!35}42.834 $\pm$ {\tiny 58.66} \textbf{\scriptsize 5} & \cellcolor{red!100!green!35}0.267 $\pm$ {\tiny 0.15} \textbf{\scriptsize 8} & \cellcolor{red!100!green!35}0.372 $\pm$ {\tiny 0.18} \textbf{\scriptsize 8} & \cellcolor{red!100!green!35}0.721 $\pm$ {\tiny 0.12} \textbf{\scriptsize 8} & \cellcolor{red!57!green!35}0.019 $\pm$ {\tiny 0.05} \textbf{\scriptsize 5} & \cellcolor{red!71!green!35}0.019 $\pm$ {\tiny 0.05} \textbf{\scriptsize 6} & \cellcolor{red!71!green!35}0.083 $\pm$ {\tiny 0.14} \textbf{\scriptsize 6} \\
      MVGenMaster & \cellcolor{red!57!green!35}5 & \cellcolor{red!43!green!35}0.040 $\pm$ {\tiny 0.05} \textbf{\scriptsize 4} & \cellcolor{red!86!green!35}0.283 $\pm$ {\tiny 0.15} \textbf{\scriptsize 7} & \cellcolor{red!43!green!35}45.951 $\pm$ {\tiny 45.28} \textbf{\scriptsize 4} & \cellcolor{red!71!green!35}0.316 $\pm$ {\tiny 0.13} \textbf{\scriptsize 6} & \cellcolor{red!71!green!35}0.414 $\pm$ {\tiny 0.16} \textbf{\scriptsize 6} & \cellcolor{red!57!green!35}0.775 $\pm$ {\tiny 0.08} \textbf{\scriptsize 5} & \cellcolor{red!43!green!35}0.030 $\pm$ {\tiny 0.07} \textbf{\scriptsize 4} & \cellcolor{red!43!green!35}0.033 $\pm$ {\tiny 0.07} \textbf{\scriptsize 4} & \cellcolor{red!43!green!35}0.114 $\pm$ {\tiny 0.17} \textbf{\scriptsize 4} \\
      MVSplat360 & \cellcolor{red!71!green!35}6 & \cellcolor{red!100!green!35}0.010 $\pm$ {\tiny 0.01} \textbf{\scriptsize 8} & \cellcolor{red!14!green!35}0.570 $\pm$ {\tiny 0.22} \textbf{\scriptsize 2} & \cellcolor{red!100!green!35}7.154 $\pm$ {\tiny 6.17} \textbf{\scriptsize 8} & \cellcolor{red!14!green!35}0.582 $\pm$ {\tiny 0.20} \textbf{\scriptsize 2} & \cellcolor{red!0!green!35}0.694 $\pm$ {\tiny 0.23} \textbf{\scriptsize 1} & \cellcolor{red!43!green!35}0.840 $\pm$ {\tiny 0.06} \textbf{\scriptsize 4} & \cellcolor{red!86!green!35}0.011 $\pm$ {\tiny 0.01} \textbf{\scriptsize 7} & \cellcolor{red!86!green!35}0.011 $\pm$ {\tiny 0.01} \textbf{\scriptsize 7} & \cellcolor{red!100!green!35}0.041 $\pm$ {\tiny 0.01} \textbf{\scriptsize 8} \\
      NVS-Solver & \cellcolor{red!86!green!35}7 & \cellcolor{red!86!green!35}0.013 $\pm$ {\tiny 0.02} \textbf{\scriptsize 7} & \cellcolor{red!29!green!35}0.409 $\pm$ {\tiny 0.26} \textbf{\scriptsize 3} & \cellcolor{red!86!green!35}14.371 $\pm$ {\tiny 23.18} \textbf{\scriptsize 7} & \cellcolor{red!29!green!35}0.437 $\pm$ {\tiny 0.25} \textbf{\scriptsize 3} & \cellcolor{red!29!green!35}0.497 $\pm$ {\tiny 0.26} \textbf{\scriptsize 3} & \cellcolor{red!0!green!35}0.875 $\pm$ {\tiny 0.11} \textbf{\scriptsize 1} & \cellcolor{red!71!green!35}0.018 $\pm$ {\tiny 0.01} \textbf{\scriptsize 6} & \cellcolor{red!57!green!35}0.021 $\pm$ {\tiny 0.02} \textbf{\scriptsize 5} & \cellcolor{red!57!green!35}0.089 $\pm$ {\tiny 0.03} \textbf{\scriptsize 5} \\
      Difix3D & \cellcolor{red!100!green!35}8 & \cellcolor{red!71!green!35}0.020 $\pm$ {\tiny 0.02} \textbf{\scriptsize 6} & \cellcolor{red!71!green!35}0.287 $\pm$ {\tiny 0.14} \textbf{\scriptsize 6} & \cellcolor{red!71!green!35}30.194 $\pm$ {\tiny 29.85} \textbf{\scriptsize 6} & \cellcolor{red!86!green!35}0.304 $\pm$ {\tiny 0.13} \textbf{\scriptsize 7} & \cellcolor{red!86!green!35}0.410 $\pm$ {\tiny 0.15} \textbf{\scriptsize 7} & \cellcolor{red!86!green!35}0.743 $\pm$ {\tiny 0.11} \textbf{\scriptsize 7} & \cellcolor{red!100!green!35}0.008 $\pm$ {\tiny 0.01} \textbf{\scriptsize 8} & \cellcolor{red!100!green!35}0.009 $\pm$ {\tiny 0.01} \textbf{\scriptsize 8} & \cellcolor{red!86!green!35}0.059 $\pm$ {\tiny 0.05} \textbf{\scriptsize 7} \\
      \midrule
      Rank Corr.\ ($\rho$)$\uparrow$ & -- & 0.857 & 0.238 & 0.857 & 0.310 & 0.190 & 0.167 & \textbf{0.929} & 0.857 & 0.810 \\
      \bottomrule
    \end{tabular}%
  }
\end{table}

\begin{table}
  \centering
  \caption{COLMAP reconstruction quality vs.\ human rankings of 8 NVS methods on DL3DV, full-metric version for the $K{=}3$ view split. Values are mean $\pm$ std over scenes. Colors encode ranks 1 (best, \colorbox{green!35}{green}) to 8 (worst, \colorbox{red!35}{red}); rows sorted by human rank (\textbf{H}). Bottom row: Spearman rank correlation ($\rho$) between human and metric-induced rankings.}
  \label{tab:colmap-human-rank-dl3dv-all-k3}
  \resizebox{\columnwidth}{!}{%
    \begin{tabular}{lcccccccccc}
      \toprule
      Method & H$\downarrow$ & W-GPC$\uparrow$ & ICM$\uparrow$ & Ang.\ Cov.$\uparrow$ & GPC$\uparrow$ & Avg.\ Den.$\uparrow$ & Avg.\ Cons.$\uparrow$ & GPC All$\uparrow$ & ICM All$\uparrow$ & Reg.\ Rate$\uparrow$ \\
      \midrule
      DepthSplat & \cellcolor{red!0!green!35}1 & \cellcolor{red!0!green!35}0.190 $\pm$ {\tiny 0.27} \textbf{\scriptsize 1} & \cellcolor{red!0!green!35}0.552 $\pm$ {\tiny 0.22} \textbf{\scriptsize 1} & \cellcolor{red!0!green!35}106.946 $\pm$ {\tiny 133.21} \textbf{\scriptsize 1} & \cellcolor{red!0!green!35}0.580 $\pm$ {\tiny 0.18} \textbf{\scriptsize 1} & \cellcolor{red!0!green!35}0.673 $\pm$ {\tiny 0.17} \textbf{\scriptsize 1} & \cellcolor{red!7!green!35}0.848 $\pm$ {\tiny 0.12} \textbf{\scriptsize 1.5} & \cellcolor{red!0!green!35}0.170 $\pm$ {\tiny 0.27} \textbf{\scriptsize 1} & \cellcolor{red!0!green!35}0.170 $\pm$ {\tiny 0.27} \textbf{\scriptsize 1} & \cellcolor{red!0!green!35}0.286 $\pm$ {\tiny 0.38} \textbf{\scriptsize 1} \\
      S-V-Camera & \cellcolor{red!14!green!35}2 & \cellcolor{red!57!green!35}0.023 $\pm$ {\tiny 0.02} \textbf{\scriptsize 5} & \cellcolor{red!43!green!35}0.381 $\pm$ {\tiny 0.19} \textbf{\scriptsize 4} & \cellcolor{red!57!green!35}32.645 $\pm$ {\tiny 39.35} \textbf{\scriptsize 5} & \cellcolor{red!43!green!35}0.387 $\pm$ {\tiny 0.18} \textbf{\scriptsize 4} & \cellcolor{red!57!green!35}0.450 $\pm$ {\tiny 0.21} \textbf{\scriptsize 5} & \cellcolor{red!7!green!35}0.848 $\pm$ {\tiny 0.08} \textbf{\scriptsize 1.5} & \cellcolor{red!43!green!35}0.022 $\pm$ {\tiny 0.03} \textbf{\scriptsize 4} & \cellcolor{red!50!green!35}0.022 $\pm$ {\tiny 0.03} \textbf{\scriptsize 4} & \cellcolor{red!57!green!35}0.077 $\pm$ {\tiny 0.10} \textbf{\scriptsize 5} \\
      Long-LRM & \cellcolor{red!29!green!35}3 & \cellcolor{red!29!green!35}0.028 $\pm$ {\tiny 0.03} \textbf{\scriptsize 3} & \cellcolor{red!93!green!35}0.283 $\pm$ {\tiny 0.17} \textbf{\scriptsize 7} & \cellcolor{red!29!green!35}34.183 $\pm$ {\tiny 31.68} \textbf{\scriptsize 3} & \cellcolor{red!100!green!35}0.292 $\pm$ {\tiny 0.16} \textbf{\scriptsize 8} & \cellcolor{red!100!green!35}0.368 $\pm$ {\tiny 0.18} \textbf{\scriptsize 8} & \cellcolor{red!43!green!35}0.792 $\pm$ {\tiny 0.13} \textbf{\scriptsize 4} & \cellcolor{red!57!green!35}0.021 $\pm$ {\tiny 0.03} \textbf{\scriptsize 5} & \cellcolor{red!50!green!35}0.022 $\pm$ {\tiny 0.04} \textbf{\scriptsize 5} & \cellcolor{red!43!green!35}0.101 $\pm$ {\tiny 0.11} \textbf{\scriptsize 4} \\
      MVGenMaster & \cellcolor{red!43!green!35}4 & \cellcolor{red!71!green!35}0.018 $\pm$ {\tiny 0.03} \textbf{\scriptsize 6} & \cellcolor{red!57!green!35}0.320 $\pm$ {\tiny 0.11} \textbf{\scriptsize 5} & \cellcolor{red!71!green!35}21.760 $\pm$ {\tiny 34.98} \textbf{\scriptsize 6} & \cellcolor{red!57!green!35}0.334 $\pm$ {\tiny 0.10} \textbf{\scriptsize 5} & \cellcolor{red!43!green!35}0.459 $\pm$ {\tiny 0.14} \textbf{\scriptsize 4} & \cellcolor{red!100!green!35}0.738 $\pm$ {\tiny 0.07} \textbf{\scriptsize 8} & \cellcolor{red!86!green!35}0.014 $\pm$ {\tiny 0.02} \textbf{\scriptsize 7} & \cellcolor{red!86!green!35}0.015 $\pm$ {\tiny 0.02} \textbf{\scriptsize 7} & \cellcolor{red!71!green!35}0.063 $\pm$ {\tiny 0.07} \textbf{\scriptsize 6} \\
      ViewCrafter & \cellcolor{red!57!green!35}5 & \cellcolor{red!14!green!35}0.038 $\pm$ {\tiny 0.06} \textbf{\scriptsize 2} & \cellcolor{red!71!green!35}0.291 $\pm$ {\tiny 0.14} \textbf{\scriptsize 6} & \cellcolor{red!14!green!35}42.596 $\pm$ {\tiny 58.12} \textbf{\scriptsize 2} & \cellcolor{red!86!green!35}0.296 $\pm$ {\tiny 0.14} \textbf{\scriptsize 7} & \cellcolor{red!71!green!35}0.387 $\pm$ {\tiny 0.18} \textbf{\scriptsize 6} & \cellcolor{red!71!green!35}0.782 $\pm$ {\tiny 0.10} \textbf{\scriptsize 6} & \cellcolor{red!14!green!35}0.027 $\pm$ {\tiny 0.05} \textbf{\scriptsize 2} & \cellcolor{red!21!green!35}0.026 $\pm$ {\tiny 0.05} \textbf{\scriptsize 2} & \cellcolor{red!29!green!35}0.111 $\pm$ {\tiny 0.18} \textbf{\scriptsize 3} \\
      Difix3D & \cellcolor{red!71!green!35}6 & \cellcolor{red!43!green!35}0.024 $\pm$ {\tiny 0.03} \textbf{\scriptsize 4} & \cellcolor{red!93!green!35}0.283 $\pm$ {\tiny 0.16} \textbf{\scriptsize 8} & \cellcolor{red!43!green!35}32.784 $\pm$ {\tiny 35.54} \textbf{\scriptsize 4} & \cellcolor{red!71!green!35}0.300 $\pm$ {\tiny 0.14} \textbf{\scriptsize 6} & \cellcolor{red!86!green!35}0.385 $\pm$ {\tiny 0.17} \textbf{\scriptsize 7} & \cellcolor{red!57!green!35}0.785 $\pm$ {\tiny 0.11} \textbf{\scriptsize 5} & \cellcolor{red!100!green!35}0.010 $\pm$ {\tiny 0.01} \textbf{\scriptsize 8} & \cellcolor{red!100!green!35}0.010 $\pm$ {\tiny 0.01} \textbf{\scriptsize 8} & \cellcolor{red!86!green!35}0.061 $\pm$ {\tiny 0.04} \textbf{\scriptsize 7} \\
      NVS-Solver & \cellcolor{red!86!green!35}7 & \cellcolor{red!93!green!35}0.013 $\pm$ {\tiny 0.01} \textbf{\scriptsize 7} & \cellcolor{red!29!green!35}0.429 $\pm$ {\tiny 0.24} \textbf{\scriptsize 3} & \cellcolor{red!86!green!35}14.734 $\pm$ {\tiny 19.32} \textbf{\scriptsize 7} & \cellcolor{red!29!green!35}0.431 $\pm$ {\tiny 0.24} \textbf{\scriptsize 3} & \cellcolor{red!29!green!35}0.518 $\pm$ {\tiny 0.24} \textbf{\scriptsize 3} & \cellcolor{red!29!green!35}0.809 $\pm$ {\tiny 0.18} \textbf{\scriptsize 3} & \cellcolor{red!29!green!35}0.025 $\pm$ {\tiny 0.02} \textbf{\scriptsize 3} & \cellcolor{red!21!green!35}0.026 $\pm$ {\tiny 0.02} \textbf{\scriptsize 3} & \cellcolor{red!14!green!35}0.144 $\pm$ {\tiny 0.12} \textbf{\scriptsize 2} \\
      MVSplat360 & \cellcolor{red!100!green!35}8 & \cellcolor{red!93!green!35}0.013 $\pm$ {\tiny 0.01} \textbf{\scriptsize 8} & \cellcolor{red!14!green!35}0.445 $\pm$ {\tiny 0.20} \textbf{\scriptsize 2} & \cellcolor{red!100!green!35}10.161 $\pm$ {\tiny 9.57} \textbf{\scriptsize 8} & \cellcolor{red!14!green!35}0.466 $\pm$ {\tiny 0.16} \textbf{\scriptsize 2} & \cellcolor{red!14!green!35}0.605 $\pm$ {\tiny 0.15} \textbf{\scriptsize 2} & \cellcolor{red!86!green!35}0.760 $\pm$ {\tiny 0.11} \textbf{\scriptsize 7} & \cellcolor{red!71!green!35}0.017 $\pm$ {\tiny 0.02} \textbf{\scriptsize 6} & \cellcolor{red!71!green!35}0.017 $\pm$ {\tiny 0.02} \textbf{\scriptsize 6} & \cellcolor{red!100!green!35}0.055 $\pm$ {\tiny 0.04} \textbf{\scriptsize 8} \\
      \midrule
      Rank Corr.\ ($\rho$)$\uparrow$ & -- & 0.667 & 0.000 & \textbf{0.690} & -0.024 & -0.048 & 0.548 & 0.405 & 0.357 & 0.476 \\
      \bottomrule
    \end{tabular}%
  }
\end{table}

\begin{table}
  \centering
  \caption{COLMAP reconstruction quality vs.\ human rankings of 8 NVS methods on MipNeRF360, full-metric version for the $K{=}3$ view split. Values are mean $\pm$ std over scenes. Colors encode ranks 1 (best, \colorbox{green!35}{green}) to 8 (worst, \colorbox{red!35}{red}); rows sorted by human rank (\textbf{H}). Bottom row: Spearman rank correlation ($\rho$) between human and metric-induced rankings.}
  \label{tab:colmap-human-rank-mip-all-k3}
  \resizebox{\columnwidth}{!}{%
    \begin{tabular}{lcccccccccc}
      \toprule
      Method & H$\downarrow$ & W-GPC$\uparrow$ & ICM$\uparrow$ & Ang.\ Cov.$\uparrow$ & GPC$\uparrow$ & Avg.\ Den.$\uparrow$ & Avg.\ Cons.$\uparrow$ & GPC All$\uparrow$ & ICM All$\uparrow$ & Reg.\ Rate$\uparrow$ \\
      \midrule
      DepthSplat & \cellcolor{red!0!green!35}1 & \cellcolor{red!0!green!35}0.266 $\pm$ {\tiny 0.35} \textbf{\scriptsize 1} & \cellcolor{red!0!green!35}0.623 $\pm$ {\tiny 0.28} \textbf{\scriptsize 1} & \cellcolor{red!0!green!35}139.076 $\pm$ {\tiny 155.77} \textbf{\scriptsize 1} & \cellcolor{red!0!green!35}0.625 $\pm$ {\tiny 0.27} \textbf{\scriptsize 1} & \cellcolor{red!0!green!35}0.692 $\pm$ {\tiny 0.30} \textbf{\scriptsize 1} & \cellcolor{red!0!green!35}0.907 $\pm$ {\tiny 0.06} \textbf{\scriptsize 1} & \cellcolor{red!0!green!35}0.241 $\pm$ {\tiny 0.35} \textbf{\scriptsize 1} & \cellcolor{red!0!green!35}0.241 $\pm$ {\tiny 0.35} \textbf{\scriptsize 1} & \cellcolor{red!0!green!35}0.344 $\pm$ {\tiny 0.46} \textbf{\scriptsize 1} \\
      S-V-Camera & \cellcolor{red!14!green!35}2 & \cellcolor{red!14!green!35}0.060 $\pm$ {\tiny 0.11} \textbf{\scriptsize 2} & \cellcolor{red!29!green!35}0.424 $\pm$ {\tiny 0.11} \textbf{\scriptsize 3} & \cellcolor{red!14!green!35}47.626 $\pm$ {\tiny 78.69} \textbf{\scriptsize 2} & \cellcolor{red!29!green!35}0.425 $\pm$ {\tiny 0.12} \textbf{\scriptsize 3} & \cellcolor{red!57!green!35}0.490 $\pm$ {\tiny 0.13} \textbf{\scriptsize 5} & \cellcolor{red!14!green!35}0.874 $\pm$ {\tiny 0.06} \textbf{\scriptsize 2} & \cellcolor{red!14!green!35}0.061 $\pm$ {\tiny 0.08} \textbf{\scriptsize 2} & \cellcolor{red!14!green!35}0.061 $\pm$ {\tiny 0.08} \textbf{\scriptsize 2} & \cellcolor{red!14!green!35}0.161 $\pm$ {\tiny 0.19} \textbf{\scriptsize 2} \\
      ViewCrafter & \cellcolor{red!29!green!35}3 & \cellcolor{red!29!green!35}0.025 $\pm$ {\tiny 0.04} \textbf{\scriptsize 3} & \cellcolor{red!57!green!35}0.371 $\pm$ {\tiny 0.13} \textbf{\scriptsize 5} & \cellcolor{red!29!green!35}29.193 $\pm$ {\tiny 43.63} \textbf{\scriptsize 3} & \cellcolor{red!57!green!35}0.374 $\pm$ {\tiny 0.13} \textbf{\scriptsize 5} & \cellcolor{red!71!green!35}0.474 $\pm$ {\tiny 0.17} \textbf{\scriptsize 6} & \cellcolor{red!29!green!35}0.797 $\pm$ {\tiny 0.08} \textbf{\scriptsize 3} & \cellcolor{red!43!green!35}0.021 $\pm$ {\tiny 0.04} \textbf{\scriptsize 4} & \cellcolor{red!57!green!35}0.020 $\pm$ {\tiny 0.04} \textbf{\scriptsize 5} & \cellcolor{red!57!green!35}0.092 $\pm$ {\tiny 0.14} \textbf{\scriptsize 5} \\
      Long-LRM & \cellcolor{red!43!green!35}4 & \cellcolor{red!100!green!35}0.006 $\pm$ {\tiny 0.01} \textbf{\scriptsize 8} & \cellcolor{red!100!green!35}0.182 $\pm$ {\tiny 0.13} \textbf{\scriptsize 8} & \cellcolor{red!100!green!35}10.153 $\pm$ {\tiny 5.64} \textbf{\scriptsize 8} & \cellcolor{red!100!green!35}0.201 $\pm$ {\tiny 0.12} \textbf{\scriptsize 8} & \cellcolor{red!100!green!35}0.310 $\pm$ {\tiny 0.19} \textbf{\scriptsize 8} & \cellcolor{red!86!green!35}0.689 $\pm$ {\tiny 0.11} \textbf{\scriptsize 7} & \cellcolor{red!100!green!35}0.002 $\pm$ {\tiny 0.00} \textbf{\scriptsize 8} & \cellcolor{red!100!green!35}0.002 $\pm$ {\tiny 0.00} \textbf{\scriptsize 8} & \cellcolor{red!71!green!35}0.048 $\pm$ {\tiny 0.02} \textbf{\scriptsize 6} \\
      MVSplat360 & \cellcolor{red!57!green!35}5 & \cellcolor{red!64!green!35}0.011 $\pm$ {\tiny 0.01} \textbf{\scriptsize 5} & \cellcolor{red!43!green!35}0.391 $\pm$ {\tiny 0.15} \textbf{\scriptsize 4} & \cellcolor{red!86!green!35}10.589 $\pm$ {\tiny 7.98} \textbf{\scriptsize 7} & \cellcolor{red!43!green!35}0.404 $\pm$ {\tiny 0.15} \textbf{\scriptsize 4} & \cellcolor{red!14!green!35}0.571 $\pm$ {\tiny 0.16} \textbf{\scriptsize 2} & \cellcolor{red!71!green!35}0.695 $\pm$ {\tiny 0.11} \textbf{\scriptsize 6} & \cellcolor{red!71!green!35}0.009 $\pm$ {\tiny 0.00} \textbf{\scriptsize 6} & \cellcolor{red!71!green!35}0.009 $\pm$ {\tiny 0.00} \textbf{\scriptsize 6} & \cellcolor{red!86!green!35}0.043 $\pm$ {\tiny 0.01} \textbf{\scriptsize 7} \\
      MVGenMaster & \cellcolor{red!71!green!35}6 & \cellcolor{red!64!green!35}0.011 $\pm$ {\tiny 0.01} \textbf{\scriptsize 6} & \cellcolor{red!79!green!35}0.319 $\pm$ {\tiny 0.07} \textbf{\scriptsize 6} & \cellcolor{red!71!green!35}12.380 $\pm$ {\tiny 5.64} \textbf{\scriptsize 6} & \cellcolor{red!71!green!35}0.323 $\pm$ {\tiny 0.07} \textbf{\scriptsize 6} & \cellcolor{red!43!green!35}0.500 $\pm$ {\tiny 0.06} \textbf{\scriptsize 4} & \cellcolor{red!100!green!35}0.642 $\pm$ {\tiny 0.07} \textbf{\scriptsize 8} & \cellcolor{red!86!green!35}0.006 $\pm$ {\tiny 0.00} \textbf{\scriptsize 7} & \cellcolor{red!86!green!35}0.006 $\pm$ {\tiny 0.00} \textbf{\scriptsize 7} & \cellcolor{red!100!green!35}0.038 $\pm$ {\tiny 0.00} \textbf{\scriptsize 8} \\
      NVS-Solver & \cellcolor{red!86!green!35}7 & \cellcolor{red!86!green!35}0.009 $\pm$ {\tiny 0.01} \textbf{\scriptsize 7} & \cellcolor{red!79!green!35}0.319 $\pm$ {\tiny 0.22} \textbf{\scriptsize 7} & \cellcolor{red!43!green!35}17.001 $\pm$ {\tiny 16.36} \textbf{\scriptsize 4} & \cellcolor{red!86!green!35}0.321 $\pm$ {\tiny 0.22} \textbf{\scriptsize 7} & \cellcolor{red!86!green!35}0.393 $\pm$ {\tiny 0.23} \textbf{\scriptsize 7} & \cellcolor{red!57!green!35}0.754 $\pm$ {\tiny 0.17} \textbf{\scriptsize 5} & \cellcolor{red!57!green!35}0.020 $\pm$ {\tiny 0.02} \textbf{\scriptsize 5} & \cellcolor{red!43!green!35}0.021 $\pm$ {\tiny 0.02} \textbf{\scriptsize 4} & \cellcolor{red!29!green!35}0.155 $\pm$ {\tiny 0.07} \textbf{\scriptsize 3} \\
      Difix3D & \cellcolor{red!100!green!35}8 & \cellcolor{red!43!green!35}0.021 $\pm$ {\tiny 0.02} \textbf{\scriptsize 4} & \cellcolor{red!14!green!35}0.453 $\pm$ {\tiny 0.15} \textbf{\scriptsize 2} & \cellcolor{red!57!green!35}15.476 $\pm$ {\tiny 11.60} \textbf{\scriptsize 5} & \cellcolor{red!14!green!35}0.450 $\pm$ {\tiny 0.15} \textbf{\scriptsize 2} & \cellcolor{red!29!green!35}0.567 $\pm$ {\tiny 0.10} \textbf{\scriptsize 3} & \cellcolor{red!43!green!35}0.776 $\pm$ {\tiny 0.14} \textbf{\scriptsize 4} & \cellcolor{red!29!green!35}0.042 $\pm$ {\tiny 0.09} \textbf{\scriptsize 3} & \cellcolor{red!29!green!35}0.043 $\pm$ {\tiny 0.09} \textbf{\scriptsize 3} & \cellcolor{red!43!green!35}0.107 $\pm$ {\tiny 0.17} \textbf{\scriptsize 4} \\
      \midrule
      Rank Corr.\ ($\rho$)$\uparrow$ & -- & \textbf{0.595} & 0.310 & 0.548 & 0.310 & 0.143 & \textbf{0.595} & 0.429 & 0.333 & 0.429 \\
      \bottomrule
    \end{tabular}%
  }
\end{table}

\begin{table}[htb]
  \centering
  \caption{Category-level behavior of COLMAP-based metrics on SysCON3D at $K{=}21$ under our custom comparison protocol. Gaussian Noise and Identical Images are assigned zero-valued metrics by convention since they do not yield meaningful COLMAP reconstructions. Rows report mean $\pm$ std across scenes. The final row reports category-level Spearman rank correlation ($\rho$) with the ideal inconsistency ordering:
\textit{Consistent} $>$ \textit{Mixed-1-Outlier} $>$ \textit{Mixed-Controlled} $>$ \textit{Gaussian Noise} $=$ \textit{Identical Images}. As can be seen, our COLMAP metrics are able to robustly rank the SysCON3D variants.}
  \label{tab:syscon3d_k21_custom_spearman}
  \begin{tabular}{lcccc}
    \toprule
    Variant & N & W-GPC$\uparrow$ & ICM$\uparrow$ & Coverage$\uparrow$ \\
    \midrule
    Consistent       & 9 & 0.011685 $\pm$ 0.008193 & 0.286463 $\pm$ 0.088432 & 19.555 $\pm$ 16.216 \\
    Mixed-1-Outlier  & 8 & 0.010812 $\pm$ 0.000000 & 0.191046 $\pm$ 0.000000 & 20.330 $\pm$ 0.000 \\
    Mixed-Controlled & 8 & 0.004896 $\pm$ 0.003891 & 0.210709 $\pm$ 0.126232 & 18.612 $\pm$ 17.540 \\
    Gaussian Noise   & 0 & 0.000000 $\pm$ 0.000000 & 0.000000 $\pm$ 0.000000 & 0.000 $\pm$ 0.000 \\
    Identical Images & 0 & 0.000000 $\pm$ 0.000000 & 0.000000 $\pm$ 0.000000 & 0.000 $\pm$ 0.000 \\
    \midrule
    Spearman $\rho$  & -- & \textbf{1.000} & 0.935 & 0.935 \\
    \bottomrule
  \end{tabular}
\end{table}

\begin{figure}
    \centering
    \includegraphics[width=\linewidth]{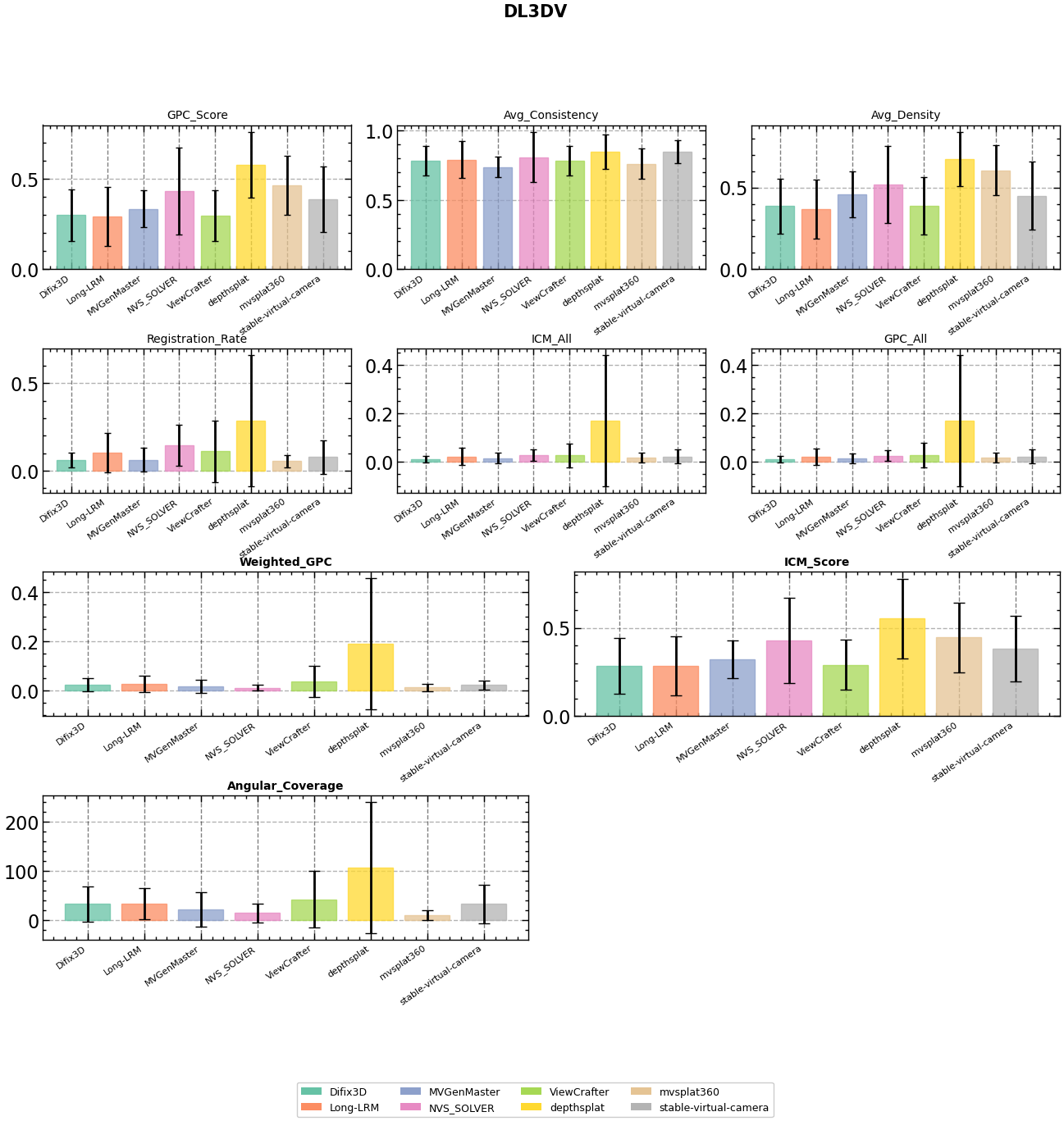}
    \caption{COLMAP based metric on DL3DV for K=3 view split.}
    \label{fig:dl3dvk3}
\end{figure}
\begin{figure}
    \centering
    \includegraphics[width=\linewidth]{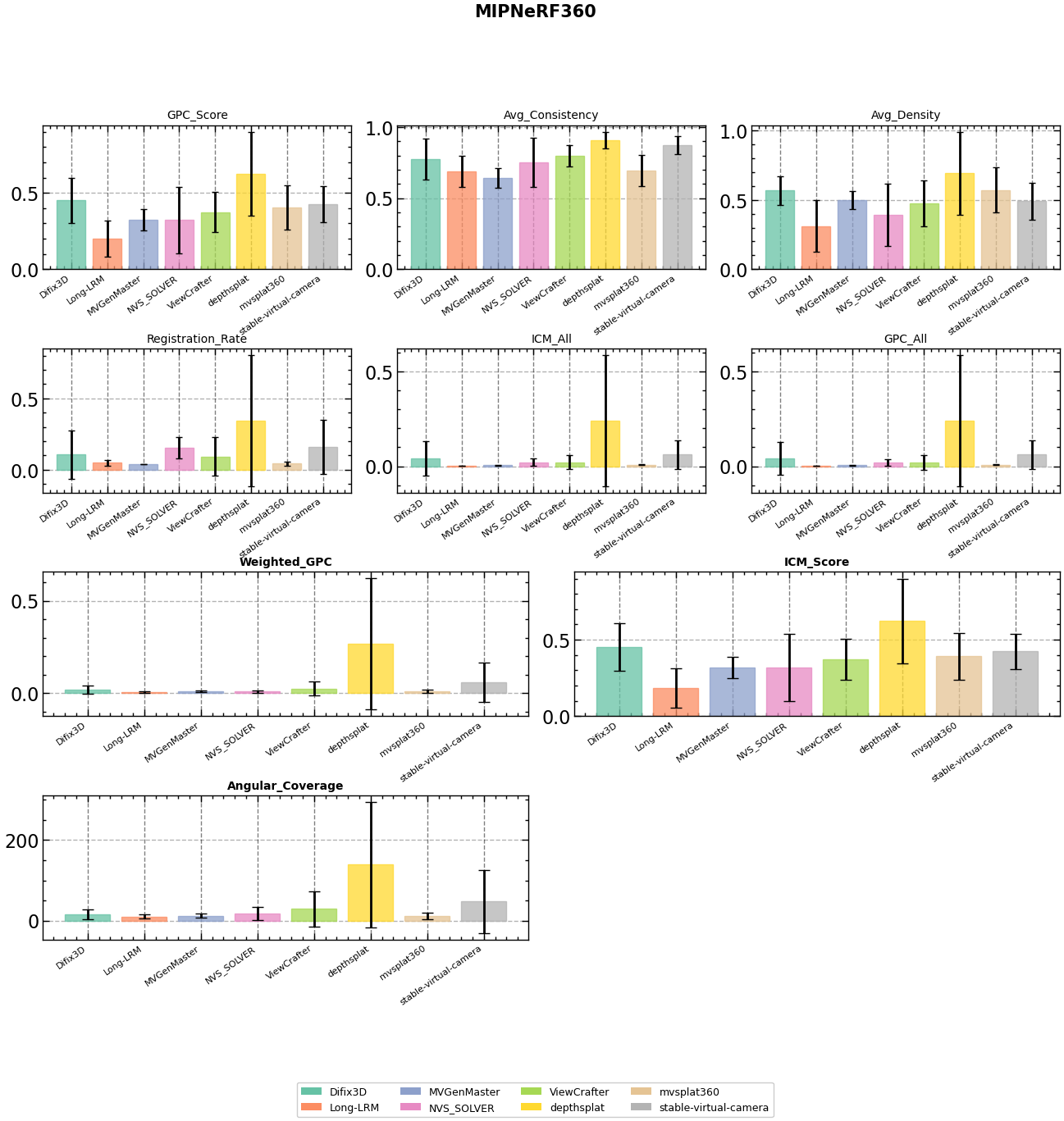}
    \caption{COLMAP based metric on MipNeRF360 for K=3 view split.}
    \label{fig:mip}
\end{figure}

\end{document}